\documentclass[acmsmall, numeric, screen, review=false, natbib=false]{acmart}
\AtBeginDocument{%
  }

\RequirePackage[
    datamodel=acmdatamodel,
    style=acmnumeric, 
    sorting=none
]{biblatex}
\setcopyright{acmlicensed}
\copyrightyear{2026}
\acmYear{2026}
\acmDOI{XXXXXXX.XXXXXXX}
\acmJournal{TOCHI}
\acmISBN{978-1-4503-XXXX-X/2025/10}
\usepackage{tabularx}
\usepackage{algorithm}
\usepackage{algpseudocode}
\usepackage{float}
\usepackage{threeparttable, booktabs, graphicx}
\usepackage{hyperref}
\usepackage{url}
\newcolumntype{Y}{>{\raggedleft\arraybackslash}X}
\newcolumntype{Z}{>{\raggedright\arraybackslash}X}

\newcolumntype{L}[1]{>{\raggedright\arraybackslash}p{#1}} 
\newcolumntype{C}[1]{>{\centering\arraybackslash}p{#1}}  
\usepackage{siunitx}
\usepackage[section]{placeins}
\newcommand{\Ind}{\mathbf{1}}

\let\citeyear\relax
\let\citeauthor\relax

\begin{document}

\title{Precision Proactivity: Measuring Cognitive Load in Real-World AI-Assisted Work}

\author{Brandon Lepine }
\email{blepine@ucsb.edu}
\affiliation{%
  \institution{College of Engineering, UC Santa Barbara}
  \city{Santa Barbara}
  \state{California}
  \country{USA}
}

\author{Juho Kim}
\affiliation{%
\institution{School of Computing, KAIST}
\country{Republic of Korea}
}

\author{Pamela Mishkin}
\affiliation{%
\institution{Independent}
\country{USA}
}

\author{Matthew Beane}
\affiliation{%
  \institution{College of Engineering, UC Santa Barbara}
  \city{Santa Barbara}
  \state{California}
  \country{USA}
}
\renewcommand{\shortauthors}{Lepine et al.,}

\begin{abstract}
Systems like ChatGPT and Claude assist billions through proactive dialogue—offering unsolicited, task-relevant information. Drawing on Cognitive Load Theory, we study how cognitive load shapes performance in AI-assisted knowledge work. We recruited 34 financial professionals to complete a complex valuation task using GPT-4o and developed a transcript-based framework estimating intrinsic and extraneous load from computational indicators anchored in a task decomposition and knowledge graph. Across 1,178 participant–subtask observations, AI-generated content usage is positively associated with quality, while extraneous load shows the largest negative association—roughly three times that of intrinsic load. Mediation reveals a compensatory pathway partially offsetting but not eliminating load-related deficits. Extraneous load persists within speakers and spills asymmetrically to model responses. Model-initiated task switching is the strongest predictor of decline. Expertise moderates these dynamics: less experienced professionals face larger penalties and derive greater marginal gains from AI-generated content, yet are not those who most increase uptake under load.
\end{abstract}



\received{5 March 2026}

\maketitle

\section{Introduction}

The productivity promise of AI rests on the assumption that systems can proactively help workers navigate complex problems and compensate for variability in human expertise \cite{gibson2024AI}. Proactivity--offering unsolicited, task-relevant information--has been linked to engagement \cite{xiao2020tell} and efficiency \cite{kuang2024enhancing}, yet maintaining coherence in extended task-oriented dialogue remains difficult \cite{wang2023dialogue}. In high-stakes knowledge work, even ostensibly helpful assistance increases coordination demands: users must select what to trust, integrate it with prior context, and maintain a coherent task representation across turns. These added processing costs can undermine performance when interaction structure becomes cluttered or misaligned \cite{lee2025impact}.

We introduce a portable measurement approach grounded in Cognitive Load Theory~\cite{sweller1994cognitive} that estimates intrinsic and extraneous cognitive load from conversational transcripts using theory-grounded computational indicators. While cognitive load is a foundational concept in HCI, its measurement in naturalistic human-AI interaction has been limited to coarse-grained methods like self-report \cite{kosch2023cognitive}, making it difficult to understand the moment-to-moment cognitive dynamics that determine the success of AI interaction. Our approach captures cognitive burden at the utterance and conversation level through domain-specific coherence metrics, task decomposition analysis, and conversational dynamics, enabling analysis not just of what users accomplish but of \textit{how} they are cognitively affected during interaction. We instantiate this approach in the domain of financial valuation and demonstrate it in a study of 34 finance professionals performing a complex financial valuation task with GPT-4o assistance; porting to other domains requires rebuilding domain-specific components such as the task decomposition and knowledge graph representation. 

\begin{figure}[t]
    \centering
    \includegraphics[width=\linewidth]{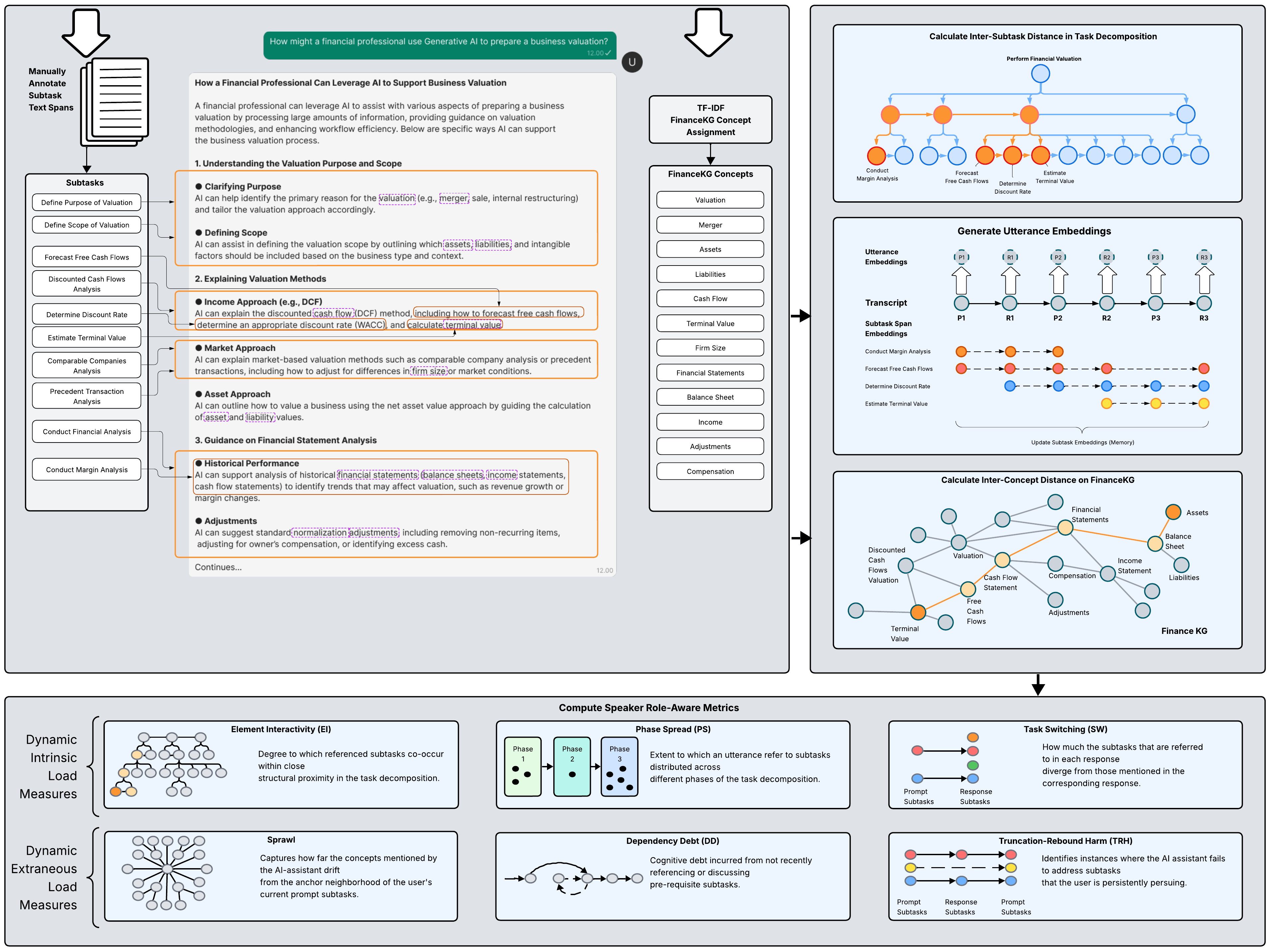}
    \caption{
         Example turn-level annotation of a user--LLM conversation using our computational framework.
        For each prompt--response pair, we estimate intrinsic and extraneous cognitive load from the transcript and compute interaction-level indicators of structural misalignment (e.g., task switching, information sprawl, phase spread).
        These measures are derived from task-decomposition structure and semantic coherence and enable fine-grained analysis of how cognitive burden and conversational organization evolve within and across speakers.
    }
    \label{fig:measurement}
\end{figure}

Our analysis yields three key findings. First, at the participant-subtask level, AI-generated content usage (AIGCU) is associated with improved output quality, while both intrinsic and extraneous load are independently negatively associated with quality. Across core mixed-effects specifications, greater usage of AI-generated content is associated with higher output quality (e.g., $\beta \approx 0.32$--$0.35$, $p<0.01$), whereas extraneous load exhibits the largest negative association (e.g., $\beta \approx -0.47$ to $-0.49$, $p<0.01$), roughly three times the magnitude of intrinsic load (e.g., $\beta \approx -0.15$, $p<0.01$). Moreover, we find no reliable evidence that the marginal association between AIGCU and quality varies with either extraneous load or intrinsic load: AIGCU $\times$ EL and AIGCU $\times$ ICL interactions are consistently null. The dominant pattern is additive: AI content usage and cognitive load contribute independently and in opposing directions to performance. 

Second, these associations are not uniform across workers. Experience moderates how extraneous load relates to both AIGCU uptake and output quality. Less experienced professionals exhibit a stronger positive association between AIGCU and quality, yet also show a substantially larger negative association between extraneous load and quality. In a moderated mediation model, more experienced professionals display a steeper positive association between extraneous load and subsequent AIGCU uptake, indicating that under higher-load conditions they incorporate more AI-generated content--potentially because their domain expertise may support more selective integration of model outputs. However, the marginal association between AIGCU and quality is smaller for experienced professionals, whereas less experienced professionals show larger quality gains per unit of AIGCU while also remaining more vulnerable to extraneous load. 

Third, at the utterance level, extraneous load is temporally persistent and largely governed by within-speaker momentum rather than reciprocal propagation between speakers. While unconditional comparisons suggest generally higher extraneous load in model responses than user prompts, this apparent role asymmetry collapses after conditioning on autoregressive dynamics and conversational controls. We observe modest, unidirectional spillover from prompt load to subsequent response load, but no reliable response-to-prompt spillover. Both speakers exhibit substantial within-speaker persistence over successive turns. Taken together, these dynamics suggest that extraneous load in human--LLM dialogue is path-dependent: once either speaker enters a high-load conversational pattern, that pattern tends to persist, and the interaction provides weak endogenous mechanisms for restoring coherence.

These findings extend Cognitive Load Theory in several ways: they show that extraneous load in human--AI dialogue is not a static property of information presentation but a path-dependent, jointly constructed feature of the interaction; that proactivity operates within path-dependent cognitive dynamics and its downstream consequences for performance depend not on whether it moderates the marginal load-quality association (which is consistently additive), but on whether it reinforces or disrupts the self-sustaining load trajectories that independently erode quality over successive turns.

These findings contribute methodologically and theoretically to HCI research on proactive dialogue systems. Methodologically, we provide a framework for measuring cognitive load from transcripts at the resolution needed to study interaction dynamics. Theoretically, we show that the benefits of AI-generated content use coexist with large, independent performance penalties associated with extraneous load, and that professional expertise shapes both the severity of these penalties and the effectiveness of the compensatory channel through which AI content offsets them. We conclude by proposing design strategies for ``precision proactivity'': systems that can detect entrenched high-load trajectories, constrain scope expansion, and actively reimpose structure to dampen rather than perpetuate cognitive burden---strategies that may also help attenuate the multi-turn degradation dynamics documented by \cite{laban2025llms}.

\section{Related Work}

\subsection{Proactivity in Dialogue Systems}\label{subsec: proactivity}

Dialogue systems aim to provide social support or functional services through natural language interactions \cite{deng2023survey}, with task-oriented dialogue (TOD) systems—those that support users in performing specific tasks \cite{hosseini2020simple}—being of particular interest given the rapid adoption of conversational AI in knowledge work \cite[e.g.,][]{brynjolfsson2023generative, dell2023navigating, eloundou2023gpts, teutloff2025winners}. Deng and colleagues \cite{deng2023rethinking} define proactivity as the capability to create or control conversations by taking initiative and anticipating impacts on users, rather than only passively responding. In TOD settings, proactivity involves automatically providing additional information that is useful but not explicitly requested, with the goal of improving conversation efficiency and quality \cite{balaraman2020proactive}.

\subsubsection{Benefits of Proactivity: Efficiency, Engagement, and Trust}

Proactivity in task-oriented dialogue (TOD) systems offers several potential benefits. Anticipating user needs can reduce the need for more conversational turns and improve efficiency during task performance and problem solving \cite{dong2025protod, hu2023enhancing}. Tailoring suggestions to user profiles can increase engagement and perceived responsiveness \cite{wang2023dialogue, brenna2025toward, lim2025proactivity}. Proactive disclosures--such as surfacing confidence estimates or providing explanations alongside recommendations--has also been associated with increased user trust and calibrated reliance \cite{buccinca2021trust, vasconcelos2023explanations}. 

However, the effectiveness of proactivity depends on its frequency and intensity. Moderate levels of proactive assistance are often preferred to purely reactive systems, yet overly frequent or intrusive interventions can reduce perceived autonomy, increase cognitive burden, and ultimately diminish trust and satisfaction \cite{kraus2020diy, kraus2020effects, kraus2021trust}. Rather than being uniformly beneficial, proactivity appears to operate within a narrow "sweet spot", beyond which additional interventions may undermine the interaction. 

\subsubsection{Cognitive Challenges of Proactive Systems}
Despite the numerous benefits of proactivity, these systems face significant barriers that can impose cognitive burdens on users. Recent work by Pu and colleagues \cite{pu2025assistance} demonstrates this tension well: while proactive programming assistants reduced users' interpretation time, they also simultaneously increased workflow disruptions--particularly when actions were unclear or poorly timed. 

\paragraph{Timing and Interruption}

Appropriate timing for proactive interventions represents a fundamental challenge. Poor timing can result in substantial user frustration \cite{berube2024proactive} and diminish trust \cite{ahire2025designing, kraus2020effects}, partly because proactive interventions can function as \textit{interruptions} that ``break the coherence of an ongoing task and block its further flow'' \cite{pu2025assistance}. Research has consistently shown that interruptions impair memory, emotional well-being, and task execution \cite{wiberg2021time, boehm-davis2009interruption}. 

Time-based triggers, such as user idleness, are unreliable indicators of readiness for assistance because they often misinterpret user intent; pauses may reflect planning, reading, or external coordination rather than confusion or need \cite{kraus2020diy, pu2025assistance}. Interventions aligned with meaningful task transitions--such as subtask completion or boundary shifts--are more consistently effective. When assistance is delivered at appropriate moments in the task flow, users report higher perceptions of system competence. Poorly timed interruptions, however, can disrupt workflow and increase the likelihood that users ignore or reject the system's suggestions \cite{kraus2023improving}. 

\paragraph{Information Selection and Processing}
Determining what information to proactively provide also presents cognitive challenges. Traditional dialogue state tracking (DST) approaches rely on fixed, domain-specific ontologies to determine which information to surface given the current dialogue state \cite{balaraman2021dialogue, feng2023dialogue}. While effective in constrained settings, such ontologies can be brittle in open-ended or evolving domains \cite{xu2024chain}. As task scope expands across turns, systems may surface information that is locally coherent but globally misaligned with the user's evolving task representation-requiring users to triage responses for relevance, reconcile inconsistencies, and maintain coherence. 

These challenges are amplified in longitudinal interaction. In multi-turn settings, each proactive intervention does not operate in isolation, it becomes part of an accumulating conversational trajectory \cite[e.g.,][]{laban2025llms}. Suggestions that are individually reasonable may, over time, expand task scope, increase topic dispersion, or introduce coordination demands that compound across turns. Proactivity therefore, presents a structural tradeoff; while it can reduce effort by anticipating needs, it may simultaneously increase cognitive coordination costs as interactions unfold. 

Despite the rich tradition of HCI research focused on attention, interruption, and cognitive workload \cite[e.g.,][]{iqbal2007disruption, mark2008cost, roda2006attention}, relatively little work has systematically examined how proactive dialogue systems shape users' cognitive burden over extended interaction and how these dynamics translate into downstream performance outcomes.

\subsection{Cognitive Load Theory}\label{subsec: CLT}
Beyond considering how the timing and content of proactive interventions influence efficiency and user engagement, we turn our attention to how proactivity affects the cognitive demands placed on users during task completion. Cognitive Load Theory (CLT) provides a framework for understanding these demands \cite{sweller1994some, sweller1994cognitive}. CLT's central premise is that working memory—the limited pool of cognitive resources available for processing information during task performance \cite{cowan2001magical}—can be occupied by different types of load. When cognitive load is manageable, resources remain available for schema construction and deeper processing thereby enabling learning; when it exceeds capacity, performance declines \cite{sweller1998cognitive, edmunds2000information}. CLT identifies two primary forms of cognitive load: \textit{intrinsic} and \textit{extraneous}. 

\textit{Intrinsic} Cognitive Load (ICL) reflects the complexity inherent to the task itself, arising from \textit{element interactivity}—the extent to which multiple task-relevant elements must be processed in relation to one another \cite{sweller1994cognitive, sweller2010element}. Intrinsic load is not fixed across individuals; domain schemas stored in long-term memory allow experts to treat interacting elements as integrated units, substantially reducing effective load relative to novices \cite{mcvee2005schema, marshall1995schemas, anderson_1983}. \textit{Extraneous} Cognitive Load (ECL) arises from the presentation of information rather than the task itself, and can be altered through information design \cite{sweller1994some, sweller1998cognitive}. ECL impairs performance by consuming working memory resources that could otherwise support learning and problem-solving—forcing users to integrate dispersed information or filter redundant content \cite{sweller1994cognitive, kalyuga1999attention, kalyuga2000learner}. In digital environments, poorly structured content can increase perceived cognitive burden \cite{just198reading}, while information that is irrelevant but perceived as interesting diverts resources from truly pertinent content \cite{kienitz2023seductive}. A small body of work suggests that moderate ECL can sometimes enhance engagement on simple tasks \cite{klepsch2020understanding, lespiau2024reasoning}, but these benefits do not appear to extend to the complex, high-stakes tasks examined here.


\section{Research Questions}\label{sec: research_questions}
The literature reveals a fundamental tension in proactive dialogue systems: while proactivity can enhance efficiency, engagement, and trust, it may simultaneously impose cognitive costs that can undermine these benefits. Existing research has identified timing and content selection challenges in proactive systems, yet has not systematically examined these challenges through the lens of cognitive load theory. 
This gap is consequential given the widespread reliance on LLM-based conversational agents for economically valuable work across both individual and organizational contexts \cite{anthroindex2026}. Prior work indicates that these proactive tendencies might be imposing additional processing costs that undermine task performance-directly contradicting the intended benefits of proactive assistance at great scale.

Drawing from this synthesis of proactive dialogue and cognitive load research, we pose three primary research questions: 

\begin{itemize}
    \item [\textbf{RQ1:}] \textit{How do extraneous cognitive load dynamics from LLM interactions affect users' ability to benefit from AI generated content?}
\end{itemize}

While proactive LLMs aim to improve task performance by providing relevant information without explicit user requests, cognitive load theory suggests that incoherent or irrelevant information can consume working memory resources and impair performance.

\begin{itemize}
    \item [\textbf{RQ2:}] \textit{How do cognitive load dynamics propagate between users and proactive LLMs during extended task-oriented conversations?}
\end{itemize}

Prior work has identified that proactive interventions can function as interruptions. Moreover, LLMs tend to become incoherent over the course of multi-turn conversations—in part because of their autoregressive nature \cite{laban2025llms}. Yet little is known about how these cognitive demands evolve dynamically as conversations progress, or how users and systems mutually shape these dynamics.

\begin{itemize}
    \item [\textbf{RQ3:}] \textit{What specific LLM behaviors and dynamics contribute most to the effects of extraneous cognitive load during task completion?}
\end{itemize}

Cognitive load theory suggests that factors such as information coherence, task switching, and interruptions significantly impact extraneous load. Identifying the specific behaviors that drive cognitive costs can inform more cognitively aligned design approaches.

\section{Measuring Cognitive Load from Conversational Traces}
\begin{figure}[t]
    \centering
    \includegraphics[width=\linewidth]{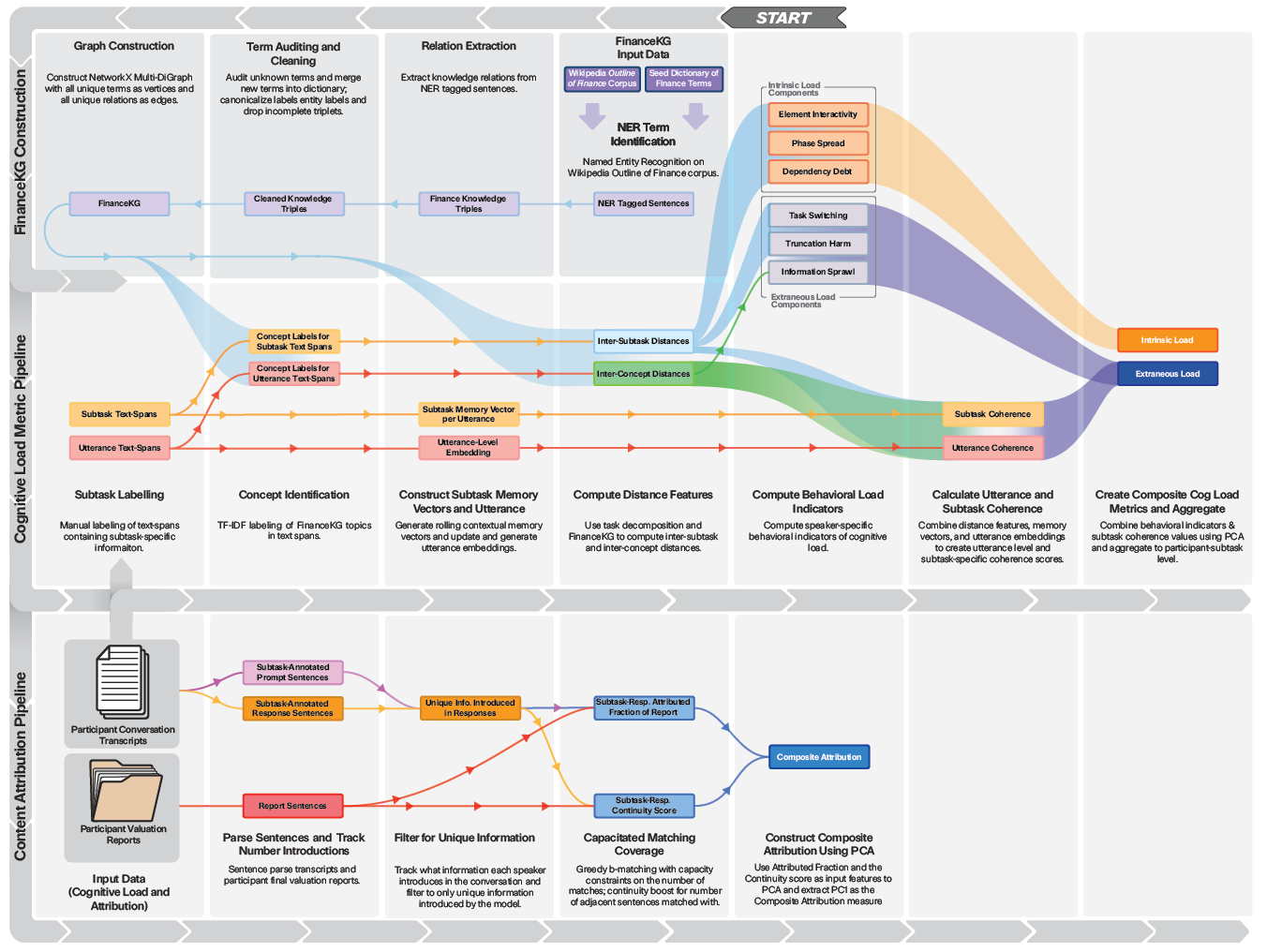}
    \caption{
    \textbf{Overview of Computational Pipeline} \\
    \textbf{Top:} Construction of the Finance Knowledge Graph (FinanceKG), including document collection, concept extraction, embedding, and graph formation. The resulting graph provides structured semantic distances and concept anchors used in extraneous load calculations (see SI \S S6). \\
    \textbf{Middle:} Participant–subtask pipeline for computing utterance-level Intrinsic Load (IL), Extraneous Load (EL), and behavioral features. This stage integrates subtask memory vectors, FinanceKG-based concept distances, autoregressive memory blending, and coherence penalties. All load measures are standardized at the utterance level prior to panel modeling (see SI \S S4). \\
    \textbf{Bottom:} Attribution pipeline for estimating AI-generated content usage (AIGCU) in the final valuation report. Transcript sentences are matched to report sentences under capacitated, thresholded similarity constraints. Content usage metrics are computed independently of the load measures (see SI \S S5).
    }
    \label{fig:pipeline-overview}
\end{figure}

We develop computational measures to capture how cognitive demands evolve during human-AI conversations. While traditional CLT assessments rely on self-reports \cite{hart1988development, babaei2025should} or dual-task paradigms \cite{esmaeili2021current, brunken2002cognitive}, we estimate cognitive load from conversation traces using theory-grounded computational indicators, enabling analysis at scale. 

A terminological note is warranted. Throughout this paper, we use the shorthand ``cognitive load'' to refer to these computational estimates derived from conversational traces, rather than to physiological or self-reported psychological states. Our indicators are grounded in CLT's theoretical constructs and operationalized through domain-specific task structure, but they should be understood as theory-grounded proxies---not direct measurements of neural or subjective load. We make this distinction explicit here so that subsequent references to ``intrinsic load'' and ``extraneous load'' are read accordingly; construct validity is further addressed in \S\ref{res:measurement_validation} and Appendix \S \ref{app:construct_validation}.

This approach is not merely a methodological workaround but a deliberate design choice suited to the demands of naturalistic, long-form knowledge work. Traditional instruments such as NASA-TLX \cite{hart1988development} cannot differentiate load by speaker, cannot track moment-to-moment evolution within a conversation, and---critically---cannot be administered without ecological intrusion: repeatedly interrupting professionals with survey instruments during a complex, multi-hour financial valuation task would fundamentally disrupt the workflow under study and compromise the very phenomena we aim to observe. Our transcript-based framework addresses all three limitations, yielding turn-by-turn, speaker-differentiated load estimates across extended naturalistic interaction without altering participants' task behavior.

\subsection{Decomposing the Task }\label{subsec: decomposing}
Analysis of turn-by-turn\footnote{By "turn" we are referring to a conversational exchange consisting of a user prompt paired with the corresponding response from the AI system. We also collectively refer to a prompt or response using the term "utterance" in a generic fashion.} work interactions requires a similarly fine-grained measurement apparatus. In the context of performing any given task, a user may query the AI model for information or assistance on a set of sub-elements and this set can be of arbitrary size–making it difficult to quantify how much distinct information is being transacted within a given utterance. Estimating the cognitive load that a user is likely encountering at any given time however, requires that we can both quantify the number of sub-elements that are referred to in each utterance and appraise their \textit{internal coherence} (i.e., their coherence within the context of that utterance) and their \textit{external coherence} (i.e., their coherence with respect to the foregoing conversation). 

To achieve this here, we decompose the financial valuation task that is the subject of our empirical study. Task decomposition has often been used in various domains for similar purposes. Early proponents of management science leveraged task decomposition as a means of standardizing work in order to boost productivity by creating more specialized roles in the divisions of labor \cite{taylor1911scientific}. Cognitive scientists have found that task decomposition is a fundamental cognitive process undertaken by humans in both the planning and execution of tasks and, consequently, have used decompositions as a methodological ploy in their empirical work to better understand these processes \cite{correa2023humans, coffey2013task, knisely2020cognitive}. More recently, AI researchers have employed task decomposition as a means to improve the performance of AI systems when undertaking complex tasks–particularly in more agentic settings \cite{kazemitabaar2024improving, khot2022decomposed, wang2025tdag}. 

The key utility of task decomposition for our purposes is that it provides a measurement apparatus for tracking what subtask-specific information is transacted by a given speaker and at a given time. Using chat transcripts, this enables computational analysis of task performance and cognitive load dynamics while simultaneously providing a structure for experts to grade participants' work outputs. The first author has finance experience and derived an initial, naive decomposition using supplemental information such as how-to guides from accepted industry standard sites (e.g., Investopedia). We recruited finance professionals\footnote{These professionals were each compensated \$20 to ensure the quality of their work.} from Upwork ($N=87$, $\mu=10.8$ years of experience) to review and correct the task decomposition, ceasing this data collection when saturation was reached. 

\subsection{Measuring Intrinsic Cognitive Load}\label{subsec:measuring-intrinsic-load}
We operationalize intrinsic cognitive load (ICL) in terms of task-level element interactivity—the degree to which multiple information elements must be processed and related simultaneously in working memory for successful task performance \cite{sweller1994cognitive, sweller2010element}. Because element interactivity is a property of the task material rather than the individual learner, our ICL measure captures objective task complexity - the coordination demands imposed by the subtask structure itself. Participant-specific variation in experienced intrinsic load, which CLT attributes to expertise-driven schema automation \cite{mcvee2005schema, marshall1995schemas, anderson_1983}, is addressed through the experience moderation analyses in \S~\ref{par:expert-heterogeneity}. Rather than treating this task complexity as a monolithic property, we model it as arising from a small set of complementary task characteristics that capture different sources of interdependence among task elements.

Specifically, we characterize each subtask along seven dimensions that have been linked in prior work to cognitive complexity and coordination demands: cognitive intensity, evaluative reasoning, argument construction, openness of solution space (closed–open), degree of structuring (unstructured–structured), problem density, and reciprocal interdependence.\footnote{See \S~S2.1 in the Supplemental Information for full rationale and formulation.} Each dimension reflects a distinct way in which task elements must be jointly represented, compared, or coordinated, thereby increasing element interactivity and intrinsic load. For example, evaluative and argumentative tasks require holding multiple criteria or premises in mind simultaneously, while high problem density and reciprocal interdependence increase the number and coupling of elements that must be tracked at once.

To obtain coverage across the full financial valuation task decomposition, we combine expert ratings on a subset of subtasks with AI-assisted scaling to estimate these characteristics for the remaining subtasks (details in Supplementary Appendix \S S2.1). We then compute a subtask's intrinsic load as the sum across the seven dimensions, capturing intrinsic load as the cumulative demands imposed by multiple interacting sources of complexity, consistent with Cognitive Load Theory's treatment of intrinsic load as task-dependent and invariant across individuals with comparable expertise \cite{sweller2010element, sweller1994some}. When multiple subtasks appear in an utterance, we average their scores to obtain an utterance-level intrinsic load estimate.

\subsubsection{Dynamic Features of Intrinsic Cognitive Load}\label{subsubsec: dynamic-features-of-icl}
In addition to aggregate scores, we capture three dynamic aspects of intrinsic cognitive load that vary with conversational context. These features reflect how individuals construct and maintain cognitive maps of a task space \cite{peer2021structuring} where effective performance depends on tracking relationships among subtasks; tracking transitions across different phases of the task; and tracking prerequisite dependencies that offer critical contextual information. Specifically, we track: \textit{(i)} when users must coordinate multiple subtasks simultaneously (\textit{element interactivity}), \textit{(ii)} when participants shift across dispersed phases of the workflow (\textit{phase spread)}, and when tasks are referenced without having recently engaged with their prerequisites (\textit{dependency debt}). Each measure operationalizes a different way in which conversational dynamics at the time of task performance can strain working memory. Also see Supplemental Information \S~S2.2 for more details. 

\paragraph{Element Interactivity}
When multiple closely related subtasks are mentioned (e.g., "calculate Net Present Value" and "Determining a Discount Rate"), users must coordinate these elements simultaneously. Multiple Resource Theory \cite{wickens2002multiple, wickens2008multiple} suggests that such coordination is especially taxing when elements draw on overlapping pools of cognitive resources. In this model, cognitive resources are considered to be a limited commodity–a notion supported by neuro-ergonomics studies finding that oxygen consumption within the brain is both limited and supports cognitive performance \cite{wickens2024multiple, ayaz2018neuroergonomics, hirshfield2024toward}. These \textit{resource demands} \cite{norman1975data} can outstrip the pool of available resources and impede performance \cite{wickens1976effects}. 
We capture this resource constrained nature of human cognition by measuring the fraction of mentioned subtask pairs situated in close proximity to one another within the task decomposition. The axiom here is that subtasks that are structurally proximate to one another will tend to call on similar pools of cognitive resources because they involve similar concepts, constructs, and objectives, increasing coordination demands on working memory.

\paragraph{Phase Spread} Shifting between dispersed phases\footnote{We use the term "phase" in reference to the the subtasks that comprise the first level down in the task decomposition hierarchy, as well as their respective subtasks. That is, a phase is a \textit{sub-tree} of the task decomposition hierarchy. For example, in the context of the financial valuation task a phase would be "Perform Data Collection", "Conduct Financial Analysis", or "Perform Industry Analysis".} of the task (e.g., moving from data collection to final reporting) requires maintaining a coherent cognitive map of the workflow \cite{peer2021structuring, garvert2023hippocampal, bokeria2021map}. Each phase is assigned an index based on their sequential ordering in the task decomposition–with all subtasks that constitute that phase receiving the same index number. The intuition is that a set of subtasks concentrated in the same phase or adjacent phases are more stable because they are more temporally cohesive. In other words, they comprise a smaller subset of the cognitive map held by the task performer. Because maintaining the workflow's cognitive map requires online computation and cognition \cite{sweller1994cognitive, cowan2001magical, marshall1995schemas}, utterances which involve a smaller range of phases prevent the task performer from having to render the entirety of the cognitive map–instead, they can load a smaller subset which imposes less cognitive burden. 

\paragraph{Dependency Debt} Dependency debt arises when a task is visited or revisited without recently engaging its conceptual or procedural prerequisites. In such cases the performer cannot rely on an already active task representation and must instead retrieve and reconstruct prerequisite elements from long-term memory. CLT suggests that this reconstruction process imposes additional demands on working memory, as multiple interacting elements must be simultaneously reactivated and coordinated \cite{sweller1994cognitive, sweller1994some}. 

The reactivation of prerequisite elements requires the performer to recall prior assumptions, intermediate results, and structural relationships before proceeding \cite{boehm-davis2009interruption, brazzolotto2022interruptions}. This retrieval and reintegration process effectively expands the number of elements that must be held and coordinated in working memory at once. As a result, even if the task itself is not inherently more complex, its effective intrinsic load increases due to the need to rebuild coherence across phases. We therefore conceptualize dependency debt as a dynamic structural property of interaction, capturing the cognitive burden imposed not by task difficulty alone, but by temporal gaps in the activation of prerequisite tasks.

\subsection{Measuring Extraneous Cognitive Load}\label{subsec:measuring-extranoeus}
While intrinsic cognitive load reflects the inherent complexity of a task, extraneous cognitive load (ECL) arises from the \textit{conditions} under which task-relevant information is encountered \cite{sweller1994some, sweller1998cognitive}. In CLT, ECL is understood as avoidable cognitive burden imposed by information that is poorly structured, mistimed, redundant, or misaligned with the learner's goals. In other words it is a burden that consumes working-memory resources without contributing to task progress or schema construction. 

Across the CLT and HCI literature, a recurring theme is that ECL emerges when individuals must expend cognitive effort to \textit{establish}, \textit{maintain}, or \textit{restore} coherence in their mental representations of the task \cite{clark1996using, pickering2004toward, sweller2010element}. When information is presented in a manner that fails to support a coherent task model–whether due to poor formatting, inappropriate timing, or irrelevant content–working memory is diverted away from productive processing toward reconciling inconsistencies, filtering noise, and re-aligning goals. 

We therefore conceptualize extraneous cognitive load as the working-memory cost of compensating for incoherence in the information environment. This framing allows us to integrate a wide range of established ECL mechanisms under a common theoretical lens and provides a foundation for computational operationalization in conversational settings. 

To structure this discussion, we distinguish three interrelated dimensions along which coherence may break down: \textit{(i)} how information is presented, \textit{(ii)} when it is presented, and \textit{(iii)} what information is presented. These dimensions are analytically separable but often interact in practice. 

\subsubsection{How Information is Presented: Local Coherence}
The manner in which information is presented directly affects users' ability to integrate it into a coherent mental representation. CLT research has long demonstrated that poorly structured presentations impose unnecessary load by forcing learners to hold partial representations in working memory while searching for missing relations \cite{chandler1992split, kalyuga1999attention}. Long or syntactically complex sentences, dispersed visual layouts, and split attention designs all increase the need for temporary storage and integration, thereby elevating extraneous load \cite{sigurd2004word, mikk2008sentence, luna2018cognitive}.

In conversational systems, similar effects arise when responses are verbose, loosely structured, or internally inconsistent. Even when information is task-relevant, presentation formats that obscure relationships between elements reduce local coherence and increase the cognitive effort required to interpret and apply the content. These costs are avoidable in principle and therefore fall squarely within the definition of extraneous load. 

\subsubsection{When Information is Presented: Temporal Coherence}
Extraneous load is also shaped by the \textit{timing} of information relative to the user's ongoing task state. Interruptions, mid-task notifications, and unsolicited interventions can disrupt attentional focus and degrade working-memory representations of task goals \cite{boehm-davis2009interruption, wiberg2021time}. When such disruptions occur, users must allocate cognitive resources to suspending the primary task, encoding the interruption, and later reconstructing the original goal structure–a process known to increase error rates, completion times, and subjective workload \cite{altmann2004task, monsell2003task, altmann2002memory, brazzolotto2022interruptions}. 
Importantly, the cognitive cost of interruptions is not solely a function of their occurrence but of their alignment with task progress. Even predictable or brief interruptions can impose extraneous loads if they arrive during cognitively demanding phases or require users to shift task sets (i.e., "context switching") \cite{kiesel2010control, rogers1995costs}. From a coherence perspective, poorly timed information fragments the temporal continuity of the task model, forcing users to repeatedly rebuild context in working memory, increasing cognitive load. 

\subsubsection{What Information is Presented: Semantic and Goal Coherence}

Finally, extraneous load is influenced by the \textit{content} of the information itself. Information that is irrelevant, redundant, or misaligned with user expectations increases cognitive demands by requiring users to evaluate its relevance, suppress distractions, or reconcile inconsistencies \cite{wilson2002truthfulness, sperber1995relevance}. CLT studies consistently show that such content consumes working-memory resources without contributing to learning or task completion \cite{park2015cognitive, trypke2023redundancy, hazan2024influence, masotina2024relevance}. 

In task-oriented dialogue, this problem is particularly acute. Responses that introduce unexpected subtasks, omit critical information, or drift toward tangential topics force users to perform additional sense-making work–to decide whether to follow, ignore, or control the system \cite{boehm-davis2009interruption}. These reconciliation processes directly compete with task-relevant cognition and thus constitute extraneous load. Empirical HCI work shows that such misalignment increases frustration, anxiety, and error rates, even when the information is superficially related to the task domain \cite{bailey2006attention, park2015cognitive}. 

\subsubsection{Operationalizing Incoherence}
Our theoretical framing treats extraneous load as the working-memory cost of restoring coherence when the information environment fails to support a stable task model. Operationally, this requires a coherence signal that is (i) sensitive to domain-specific task dependencies, (ii) defined over temporally unfolding dialogue, and (iii) measurable both as localized, turn-level deviations and as sustained drift across a conversation. We therefore operationalize incoherence using a task-grounded coherence model augmented with a finance knowledge graph and rolling contextual memory.

\paragraph{Why Generic Dialogue Coherence is Insufficient}
Generic coherence metrics typically evaluate whether a response is topically consistent with conversational context. For example, GRADE \cite{huang2020grade} represents dialogues as topic graphs aligned with commonsense knowledge graphs such as ConceptNet \cite{speer2017conceptnet}, and DynaEval \cite{zhang2021dynaeval} models dialogue-level dynamics using graph neural networks over utterance dependencies. More recent approaches reframe coherence as a problem of entailment (e.g., DialogueEntailment \cite{dziri2019evaluating}), or attempt to produce quantifiable scores aligned with human ratings (e.g., QuantiDCE \cite{ye2021towards}). While effective in open-domain conversations, these methods remain limited in task-oriented settings because they capture \textit{surface topicality} rather than domain-specific task-dependencies and knowledge that determine whether a response supports the user's mental model of the task. 

\paragraph{Domain-Grounded Coherence via a FinanceKG}
In task-oriented domains such as financial valuation, surface topical similarity can be misleading (e.g., a response elaborating on comparable company selection criteria when the user is asking about projecting free cash flows-both fall within financial valuation but address different phases of the task.). Prior work shows that domain-specific knowledge graphs capture the semantic and dependency structure needed for more discriminative alignment judgments \cite{abu2021domain, ding2025enhancing}. We therefore augment our coherence model with a Finance Knowledge Graph (FinanceKG) and use shortest-path distance in the graph as a penalty for conceptual drift-allowing the model to distinguish responses that are \textit{topically related} but \textit{structurally misaligned} from those that address the user's active subtask. Construction details are provided in the Supplementary Information \S~S6

\subsubsection{Rolling Contextual Memory}
Because dialogue unfolds over time, coherence cannot be measured solely by comparing adjacent turns. We therefore maintain evolving memory states for each subtask, continuously updating these representations with embeddings from new utterances. These rolling representations are then pooled with FinanceKG topic embeddings to construct utterance-level vectors that reflect both immediate and cumulative conversational context. Similar to dialogue coherence models that emphasize whole-dialogue dynamics rather than isolated turns \cite{zhang2021dynaeval, dziri2019evaluating}, this design aligns with theories of conversational grounding and interactive alignment, which view dialogue as the maintenance of a shared and evolving situation model \cite{clark1996using, pickering2004toward}. When an LLM response disregards or contradicts the established context, users must reconcile these inconsistencies, directly imposing extraneous cognitive load \cite{sweller1994cognitive, altmann2002memory, altmann2014momentary}. Also see Section S1.0.3 in the Supplemental Materials.

\subsubsection{Two-Level Measurement Framework}
To reflect both localized coherence failures and sustained drift, we evaluate coherence–and thus extraneous load–at two levels. At the \textit{utterance level}, we assess alignment between each user prompt and the corresponding model response, capturing momentary spikes in incoherence. At the \textit{participant-subtask level}, we track how consistently a given subtask is represented across all of a participant's conversational turns, capturing whether the system maintained alignment with the user's evolving task model over the course of the dialogue. 

\subsubsection{Anchored Coherence as a Measure of Extraneous Load}\label{subsubsec:anchored-coherence}
Our primary measure of extraneous cognitive load captures the extent to which LLM responses remain aligned with the user's evolving task model over the course of the interaction. Drawing on Cognitive Load Theory, we treat misalignment in task-relevant communication as a source of avoidable working-memory demand: when responses fail to connect to the subtasks users are actively reasoning about, users must expend cognitive effort to reconcile, reinterpret, or re-establish coherence \cite{sweller1994cognitive, clark1996using, altmann2002memory}. 

We operationalize this construct through \emph{prompt-anchored subtask coherence} (PSC). Conceptually, PSC assesses whether the subtasks invoked by the user are meaningfully addressed by the system, and conversely, whether proactive content introduced by the systems connects back to the user's stated goals. This bidirectional anchoring reflects theories of conversational grounding and interactive alignment, which emphasize that dialogue coherence is jointly maintained by speakers over time \cite{clark1996using, pickering2004toward}. Rather than relying on surface-level topical similarity, PSC evaluates alignment relative to a structured task decomposition and a domain-specific knowledge graph. This ensures that coherence is defined with respect to dependencies and representations that actually matter for task execution, rather than generic semantic relatedness. When alignment is high, users can integrate responses with minimal additional cognitive effort; when alignment is low, users must engage in compensatory processing, directly contributing to extraneous cognitive load. To facilitate interpretation, we report extraneous load as the complement of coherence (EL-PSC), such that higher values indicate greater avoidable cognitive burden arising from conversational incoherence. The formal definitions are provided in \S S4 of the Supplemental Materials.

A clarification on analytic scope is warranted. At the participant-subtask level, our interest centers on response-side extraneous load-the burden imposed on users when model outputs fail to cohere with their active cognitive model of the task. Prompt-side coherence is less consequential here because users are producing, rather than receiving content. At the utterance level, our analytic purpose shifts from estimating load experienced by the user to characterizing coherence dynamics across the conversation as a whole. Prompt extraneous load accordingly indexes the user's own organizational state which is informative because it conditions the model's response and subsequent interaction trajectory. Thus, at the utterance level, response EL reflects the processing burden imposed on the user, while prompt EL captures the organizational conditions the user brings to the exchange. 

\paragraph{Avoiding Mechanical Autoregressive Persistence in Utterance-Level Analysis}
While the participant-subtask formulation incorporates memory blending to approximate evolving subtask-specific memory, the utterance-level extraneous load metric used for analyzing temporal dynamics does not incorporate rolling contextual-memory representations from prior turns. It combines two components: \textit{(i)} clutter, computed entirely within the focal utterance from subtask-span and concept embedding alignment (SI \S 4.2), and \textit{(ii)} information sprawl, which evaluates conceptual divergence relative to the other speaker's subtask anchors (SI \S3.2). For responses, sprawl is assessed within the same conversational turn (response topics against prompt anchors); for prompts,, sprawl is assessed against the preceding response's subtask anchors.The key design constraint is that neither component carries forward accumulated discourse state across multiple turns, ensuring that serial correlation in the dynamic models reflect empirical temporal dependence rather than mechanical inheritance from the measurement procedure. We note that the cross-turn anchoring of prompt sprawl introduces a single-lag dependency on the prior response's subtask content; however, this dependency would bias towards detecting response-to-prompt spillover, making the null spillover result in Table~\ref{tab:rq2_core_dynamics} conservative with respect to this design. Formal definitions and construction details are provided in SI \S4.2

\subsubsection{Individual Behavioral Features of Coherence}
Beyond aggregate coherence, we propose three recurring conversational patterns that give rise to extraneous cognitive load by disrupting coherence in distinct ways based on the CLT literature. Each reflects a different mechanism through which users are forced to allocate working memory-resources toward managing misalignment rather than progressing on the task. Full formulations are given in \S S3 of the Supplemental Materials. 

\paragraph{Task Switching}
Task switching captures \textit{partial} interruptions in which attention is redirected to a different subtask without fully abandoning the original goal\cite{altmann2004task}. Prior work has shown that such shifts impose cognitive costs by requiring reconfiguration of the active goal representation and temporary retention of displaced task elements in working memory \cite{monsell2003task, kiesel2010control}. In human-LLM dialogue, this occurs when the system proactively introduces subtasks that the user did not explicitly request, or when conversational focus oscillates across subtasks. These switches have the potential to fragment temporal coherence and increase extraneous load by forcing users to repeatedly re-establish their task focus. 

\paragraph{Information Sprawl}
Information sprawl reflects semantic divergence between the user's focal subtasks and the concepts introduced in the systems response. Research on attention and working memory demonstrates that conceptually distant or tangential information increases filtering demands and distracts from goal-relevant processing \cite{cowan2012models, oberauer2019working}. When responses expand beyond the conceptual neighborhood implied by the user's prompt, users must evaluate relevance, suppress distractions, and decide what to ignore–processes that consume working-memory resources without advancing task completion. 

\paragraph{Truncation}
Truncation captures cases where user-specified subtasks are omitted from an LLM response and subsequently reintroduced  by the user in the following turn–signifying that the model defied the user's request and the user is persisting in attempting to address that subtask. Work on task interruption and resumption shows that unresolved goals impose cognitive costs by requiring users to actively maintain them in working memory, monitor for completion, and re-articulate when ignored \cite{altmann2004task, altmann2007timecourse}. In conversational settings, truncation reflects a failure of goal continuity, forcing users to perform additional coordination work that directly contributes to extraneous cognitive load.

\subsection{Measuring AI Content Usage}\label{subsec:measuring_ai_content_usage}

Having measured cognitive load during conversations, we next ask whether and how participants incorporate AI-generated content into their final work products. Our objective is to measure \emph{content uptake}-the degree to which model-generated material actually appears in participants' final reports-irrespective of whether that content is objectively accurate with respect to the valuation task. What matters for our analysis is not the correctness of the model's outputs but whether participants incorporated it and how that incorporation relates to expert graders' assessments of output quality. 

\subsubsection{Participant--Subtask Level}

We define AI-Generated Content Usage (AIGCU) for each participant-subtask pair as the degree to which content introduced in model responses appears in the participant's final valuation report. This distinguishes AI contributions that genuinely shape work outputs from those that are ignored or discarded. 

To estimate reuse, we employ a capacitated semantic matching procedure between response sentences and report sentences. The procedure captures two complementary dimensions of content incorporation: \textit{(i)} \textit{coverage}, the fraction of report content attributable to model responses, and \textit{(ii)} \textit{structural continuity}, whether matched content appears in coherent runs rather than scattered fragments. These signals are standardized and combined via Principal Component Analysis (PCA) into a single composite AIGCU score. Full construction details appear in the Supplementary Information (\S S5.1). 

\subsubsection{Utterance Level}

To examine when and which responses are incorporated, we extend AIGCU to the utterance level by attributing report reuse to individual response utterances. This turn-indexed measure enables analyses of selective uptake and temporal dynamics across the conversation. 

As in the participant-subtask variant, attribution relies on capacitated semantic matching and continuity diagnostics. To prevent inflated usage estimates, attribution is restricted to unique content introduced by the response itself, excluding repetition or user-supplied material. Composite utterance-level usage scores are constructed via PCA from the attribution channels. Full methodological details appear in the SI (\S S5.2).

\section{Methods}

Here we apply our analytical framework to examine how cognitive load shapes the value of AI assistance in a complex financial valuation task. We use the cognitive load measures (\S~\ref{subsec:measuring-intrinsic-load} \& \S~\ref{subsec:measuring-extranoeus})to track mental burden throughout participant-ChatGPT conversations, while the content usage measures (\S~\ref{subsec:measuring_ai_content_usage}) quantify how much AI-generated material participants incorporated into their final reports. By analyzing the relationship between these cognitive costs and content benefits, we identify when and why proactive AI behaviors help versus hinder task performance.

\subsection{Procedure}
We performed a study where participants completed a complex financial valuation task under time pressure and with the assistance of ChatGPT (GPT 4o v2024-04-01). The study was quasi-experimental in that we observed natural variation in AI usage patterns and cognitive load rather than randomly assigning participants to different AI interaction conditions, allowing us to examine how spontaneous differences in human-AI collaboration affected outcomes. The study protocol was reviewed and deemed exempt by the local Institutional Review Board (IRB) and all participants provided informed consent. 

\subsection{Participants}
We recruited 38 finance and accounting professionals via Upwork  (median experience = 4 years). Four of our participants' data was excluded \textit{a priori} for incomplete transcripts, failing to follow instructions, or too few observations leaving $N=34$ participants for analysis. Participants were compensated at a rate of \$75 per hour for their time to motivate quality work. 

\subsection{Study Task}
Participants were provided with a short scenario about a fictional, struggling, tire-manufacturing firm undergoing merger talks to orient them to the study task. They were also provided with synthetic financial data for the fictional firm to ensure that no data appeared in ChatGPT's training data. Participants were instructed to take no more than 10 hours to perform the valuation. Once 10 hours had passed, participants were instructed to stop work–even if they felt that their valuation reports were incomplete. Conducting a financial valuation is known as a complex and demanding task and, in professional settings, is typically performed by teams of analysts over a series of days or even weeks, depending on the complexity and the reason the valuation is being conducted \cite{damodaran2012investment}. We selected a complex, time-consuming, and cognitively taxing real-world task and asked participants to complete it under time constraints, both to emulate real-world conditions and encourage the usage of ChatGPT. We instructed participants to use ChatGPT when it made most sense to them and as frequently or infrequently as they wished. At the completion of the study task, participants provided us with their final valuation report, other interim work outputs such as Excel files, and their ChatGPT interaction transcripts. Their interaction transcripts were downloaded, scrubbed of personally identifying information, assigned a participant identification, and stored on a password-protected hard drive only available to the research team. 

\subsection{Qualitative Coding Procedure}
We used qualitative coding software ATLAS.ti (v24.2.1) to analyze 2,183 pages of transcript text ($\mu=64.21$, $median=36.5$) that contained 1,176 conversational turns ($\mu=34.35$, $max=179$). The average length of prompts was 34 words, and the average length of responses was 404 words.

Prior to coding the substantive content of the transcripts, two coauthors first labeled the turn number for each transcript and identified the speaker for each utterance. Structured qualitative coding \cite{bingham2023data} involved the use of the financial valuation task decomposition as a coding schema to identify all text spans that corresponded to each subtask.  We made multiple passes over each transcript to ensure that all relevant subtask text spans were properly tagged, stopping when saturation was reached. In instances of ambiguity, tie-breaking meetings were held to determine the appropriate subtask code to assign. Moreover, we consistently compared the subtasks mentioned in a given prompt or response with those mentioned in the corresponding utterance within the dyadic turn to ensure that no mentioned subtasks were overlooked when labeling the text spans. We manually identified and labeled more than 8,500 discrete code instances across all 34 participant transcripts ($\mu=252.33$ instances per transcript).\footnote{Our qualitative labeling procedure was not strictly limited to subtasks in the financial valuation decomposition. Rather, we also assign codes for emerging phenomena of interest as is often the case in qualitative data analysis.}

\subsection{Output Grading}
After de-identifying participants' final work outputs, we recruited a pool of financial valuation experts from Upwork\footnote{Our finance experts were compensated for each report that they graded} (median of 18 years of experience). We provided our experts with a rubric for grading participants' work outputs that listed each of the 101 subtasks in the financial valuation decomposition. The rubric asked raters to evaluate the provided output on three criteria on a 5-point Likert scale: completeness, output quality, and room for improvement. We also posed a single binary question to indicate if the subtask represented a work output that would be considered satisfactory in a work context.

\section{Results}

\subsection{Data Overview}
Following our data processing pipeline, we created a dataset mapping each participant's engagement with subtasks across both their ChatGPT transcripts and final reports. Out dataset contains 3,434 participant-subtask observations (34 participants $\times$ 101 graded subtasks), of which expert graders deemed 3,394 (98.84\%) as attempted in participants' reports. We identified 1,396 observations (41.13\%) where subtasks appeared in participants' ChatGPT conversations and which were attempted in participants' final reports. These observations are the focus of our analysis, allowing us to trace how content shared with AI translated into final work outputs ($\mu=41.06$ subtasks per participant).

At the utterance-level of analysis, our dataset consists of 1,176 dyadic (User Prompt : Model Response) turns, amounting to 2,352 utterance observations ($\mu=69.15$; $\tilde{x}=44$ utterances per participant)\footnote{Herein, we use the term \textit{'utterance'} to refer to an entire prompt or an entire response, while a \textit{turn} constitutes the prompt-response pair.}. Of these observations 294 (12.51\%) make no reference to the subtasks in the Financial Valuation Task Decomposition Graph. In total, there are 6,232 subtask mentions across the 2,057 utterances ($\mu = 3.03$ mentions per utterance).

Table~\ref{tab:descriptives_all} (Appendix) reports descriptive statistics across analytic samples. At the participant-subtask level, Output Quality average 2.66 (SD=1.74) on the original 0-5 grading scale. AI-generated content usage exhibits substantial dispersion, consistent with heterogeneity in reliance on model-generated content. Professional experience averages approximately nine years. 

In the PSC-restricted sample, bidirectional PSC (raw coherence) has a mean of 0.33 (SD=0.62), indicating considerable variation in coherence across participant-subtask pairs. At the utterance level, intrinsic and extraneous load measures are centered near zero by construction but exhibit meaningful dispersion. Content usage at the utterance level is highly right-skewed, reflecting occasional segments of significant content usage within otherwise moderate interactions. 

\paragraph{Analytic Samples and Attrition}
Sample sizes vary across analytic levels due to construct availability and modeling requirements (see Appendix \S~\ref{apsec:data-and-sample-construction}). At the participant-subtask level, our primary analytic sample includes 1,396 observations corresponding to subtasks that were both attempted in participants' final reports and discussed in their interaction transcripts. This is further narrowed to 1,344 observations as we exclude observations that do not have behavioral indicators of intrinsic cognitive load. Models incorporating the extraneous load measure are estimated on 1,178 observations because the construction of the contextualized subtask representations requires sufficient repeated subtask mentions within a conversation for stability. 
At the utterance level, transcripts contain 2,351 usable utterances (1,176 turns). A single response utterance is missing where the LLM returned an error and is thus, omitted. Dynamic fixed-effects specifications further require complete data on current and lagged intrinsic and extraneous load measures; accordingly, these models are estimated on 1,940 utterance observations. 

\subsection{Participant-Subtask Level Analysis}

\subsubsection{Core Models (RQ1)}

\paragraph{Modeling Approach.} To estimate relationships among AI-generated content usage, cognitive load, and subtask output quality, we fit linear mixed-effects models with participant-level random intercepts, estimated via maximum likelihood \cite{meteyard2020best}. This accounts for repeated subtask observations nested within participants and supports model comparison across nested specifications. All continuous predictors are z-scored; fixed-effects inference uses Satterthwaite approximated degrees of freedom, which provide small-sample corrections for uncertainty in variance component estimation and are standard in mixed-effects modeling with unbalanced hierarchical data \cite{satterthwaite1946approximate}. 

\begin{table}[H]
\centering
\scriptsize
\setlength{\tabcolsep}{4pt}
\renewcommand{\arraystretch}{1.15}
\caption{Core mixed-effects models predicting output quality (RQ1).}
\label{tab:rq1_core_models}
\begin{tabular}{lcccc}
\toprule
 & \multicolumn{4}{c}{Dependent Variable: Output Quality} \\
\cmidrule(lr){2-5}
 & (1) AIGCU Only & (2) + Load & (3) + Controls & (4) AIGCU $\times$ EL \\
\midrule
\multicolumn{5}{l}{\textit{Main effects}} \\
AI-Generated Content Usage (AIGCU)
 & 0.274*** & 0.349*** & 0.350*** & 0.322*** \\
 & (0.057)  & (0.059)  & (0.059)  & (0.070)  \\
Intrinsic Cognitive Load (ICL)
 &        & $-$0.147*** & $-$0.149*** & $-$0.145*** \\
 &        & (0.043)     & (0.043)     & (0.043)     \\
Extraneous Load (EL--PSC)
 &        & $-$0.472*** & $-$0.485*** & $-$0.471*** \\
 &        & (0.078)     & (0.078)     & (0.080)     \\
\addlinespace
\multicolumn{5}{l}{\textit{Interaction}} \\
AIGCU $\times$ EL--PSC
 &        &        &        & 0.056 \\
 &        &        &        & (0.074) \\
\addlinespace
\multicolumn{5}{l}{\textit{Controls}} \\
Professional Experience (Years)
 &        &        & $-$0.524*** & $-$0.528*** \\
 &        &        & (0.184)    & (0.184)    \\
Prior Organizational Seniority
 &        &        & 0.353* & 0.355* \\
 &        &        & (0.187) & (0.187) \\
\addlinespace
Intercept
 & 2.875*** & 2.807*** & 2.814*** & 2.791*** \\
 & (0.183)  & (0.163)  & (0.146)  & (0.149)  \\
\midrule
Observations              & 1,344 & 1,178 & 1,178 & 1,178 \\
Participants              & 33    & 33    & 33    & 33    \\
Estimation                & ML    & ML    & ML    & ML    \\
\bottomrule
\end{tabular}
\vspace{0.6em}
\footnotesize \\
\textit{Notes.} Linear mixed-effects models estimated via maximum likelihood with participant-level random intercepts. All continuous predictors are standardized. EL–PSC denotes bidirectional prompt–response extraneous load. Standard errors in parentheses. * $p \leq 0.10$, ** $p \leq 0.05$, *** $p \leq 0.01$.
\end{table}

\paragraph{Results}
Table \ref{tab:rq1_core_models} reports the core models corresponding to our first research question. 

Model (1): AI-generated content usage. AI-generated content usage (AIGCU) is positively associated with output quality ($\beta=0.274$, $p<0.01$). 

Model (2): Adding intrinsic load (ICL) and extraneous load (EL-PSC) yields two negative associations with quality: EL-PSC ($\beta=-0.472$, $p<0.01$) and ICL ($\beta=-0.147$, $p<0.01$). AIGCU remains positive and significant ($\beta=0.349$, $p<0.01$) indicating that the usage of AI-generated content is associated with variance in output quality beyond cognitive load differences.  

Model (3): Background Controls. Adding professional experience and prior seniority does not materially change the main coefficients (AIGCU $\beta = 0.350$; ICL $\beta=-0.149$; EL-PSC $\beta=-0.485$, $p<0.01$). The coefficient for professional experience is negative $\beta=-0.524$, $p<0.01$), while seniority is positively associated at the 90\% confidence level ($\beta=0.353$, $p<0.10$).  

Model (4): Moderation by extraneous load (AIGCU $\times$ EL-PSC). The interaction between AIGCU and EL-PSC is positive but not statistically significant ($\beta=0.056$, $p=0.45$), providing no evidence that the marginal association between AIGCU and quality varies across levels of extraneous load in this specification. The main effects remain stable: EL-PSC ($\beta=-0.471$, $p<0.01$), ICL ($\beta=-0.145$, $p<0.01$) and AIGCU ($\beta=0.320$, $p<0.01$). Professional experience remains negative ($\beta=-0.528$, $p<0.01$) and seniority remains marginally positive ($\beta=0.355$, $p<0.1$)

Interpretation. Across models, higher AIGCU is consistently associated with higher quality, while extraneous load shows the largest negative association with quality. Extraneous load has a coefficient roughly three times the magnitude of intrinsic load ($\beta \approx -0.48$ vs.\ $\beta \approx -0.15$), strengthening the pattern that poorly structured AI dialogue is associated with substantially larger quality deficits than task-inherent complexity alone. This is consistent with CLT's prediction that avoidable processing demands are more detrimental than inherent task complexity, though our observational design does not permit causal attribution. The interaction between AIGCU and extraneous load does not reach significance.

\subsubsection{Cognitive Load and Content Usage Interactions}

\paragraph{Modeling Approach}
To test whether the relationship between AIGCU and output quality depends on cognitive load, we estimate interaction models between AIGCU and \textit{(i)} EL-PSC and \textit{(ii)} ICL, with and without controls (Table \ref{tab:quality_interactions_linear}). We also test nonlinear variants (Supplemental Materials \S~8.1).

\paragraph{Linear Interactions}

Across specifications, interaction terms between AIGCU and both forms of cognitive load are small in magnitude and statistically insignificant (see Table~\ref{tab:quality_interactions_linear} in Appendix \S \ref{app:linear-interaction-models}). Although extraneous load and intrinsic load are each negatively associated with output quality, and AIGCU is positively associated with quality, there is no evidence that the marginal association between AIGCU and performance systematically varies across levels of load. 

In joint specifications including both load types, this pattern persists. Extraneous load exhibits the largest negative association with quality, intrinsic load shows a smaller but still significant negative association, and AIGCU remains positively associated with quality. However, neither interaction term approaches conventional significance thresholds. Taken together, these findings suggest an additive rather than multiplicative relationship: AI-generated content use and cognitive load independently relate to performance with no evidence that AI usage mitigates or amplifies the performance implications of cognitive load.

\paragraph{Nonlinear Checks}
Table~S4 in the Supplemental Materials evaluates nonlinear variants. The EL-PSC$^2$ main effect remains null ($\beta \approx -0.035$ to $-0.036$, $n.s.$), confirming an approximately linear relationship between extraneous load and quality. Both the linear AIGCU $\times$ EL-PSC interaction (Models 1–2: $\beta \approx 0.08$–$0.09$, $n.s.$) and the quadratic AIGCU $\times$ EL-PSC$^2$ interaction ($\beta \approx -0.03$ to $-0.04$, $n.s.$) fail to reach significance, providing no evidence that the association between AIGCU and quality varies across the extraneous load range.

In the intrinsic-load nonlinear specification (Models 3–4), ICL$^2$ remains negative and statistically significant ($\beta \approx -0.069$, $p<0.01$), indicating a concave relationship between intrinsic load and quality. This suggests that performance declines accelerate at higher levels of intrinsic complexity rather than decreasing at a constant rate. However, neither the AIGCU $\times$ ICL interaction ($\beta \approx -0.085$, $n.s.$) nor the AIGCU $\times$ ICL$^2$ interaction ($\beta \approx -0.032$, $n.s.$) reaches significance, reinforcing that the relationship between AIGCU and quality does not vary with intrinsic load in either its linear or curvilinear form.

Across both specifications, the main associations remain stable and additive: extraneous load shows the largest negative association with quality ($\beta \approx -0.54$, $p<0.01$ in Model~4), followed by intrinsic load ($\beta \approx -0.25$, $p<0.01$), while AIGCU remains positively associated with quality throughout.

\subsubsection{Cognitive Load and Work Experience}

\begin{table}[t]
\centering
\footnotesize
\setlength{\tabcolsep}{3.5pt}
\renewcommand{\arraystretch}{1.15}
\caption{Expertise heterogeneity in associations between cognitive load, AI content usage, and output quality (core coefficients).}
\label{tab:expertise_heterogeneity_core}
\begin{tabular}{lccccc}
\toprule
 & \multicolumn{5}{c}{Dependent Variable: Output Quality} \\
\cmidrule(lr){2-6}
 & (1) Exp $\times$ EL
 & (2) Exp $\times$ AIGCU
 & (3) Exp $\times$ ICL
 & (4) Joint Two-Way
 & (5) Three-Way \\
\midrule
\multicolumn{6}{l}{\textit{Main effects}} \\
AIGCU
 & 0.332*** & 0.398*** & 0.350*** & 0.386*** & 0.375*** \\
 & (0.060)  & (0.062)  & (0.059)  & (0.062)  & (0.072)  \\
Intrinsic Load (ICL)
 & $-$0.150*** & $-$0.153*** & $-$0.149*** & $-$0.156*** & $-$0.154*** \\
 & (0.043)     & (0.043)     & (0.043)     & (0.042)     & (0.043)     \\
Extraneous Load (EL--PSC)
 & $-$0.467*** & $-$0.468*** & $-$0.485*** & $-$0.438*** & $-$0.442*** \\
 & (0.078)     & (0.078)     & (0.078)     & (0.078)     & (0.081)     \\
\addlinespace
\multicolumn{6}{l}{\textit{Cross-level interactions}} \\
Experience $\times$ EL--PSC
 & 0.145** &        &        & 0.208*** & 0.208*** \\
 & (0.074) &        &        & (0.077)  & (0.078)  \\
Experience $\times$ AIGCU
 &        & $-$0.121** &        & $-$0.159*** & $-$0.180*** \\
 &        & (0.050)    &        & (0.051)     & (0.062)     \\
Experience $\times$ ICL
 &        &        & $-$0.004 &        &        \\
 &        &        & (0.043)  &        &        \\
AIGCU $\times$ EL--PSC
 &        &        &        &        & 0.026 \\
 &        &        &        &        & (0.075) \\
Experience $\times$ AIGCU $\times$ EL--PSC
 &        &        &        &        & 0.048 \\
 &        &        &        &        & (0.076) \\
\midrule
Observations   & 1{,}178 & 1{,}178 & 1{,}178 & 1{,}178 & 1{,}178 \\
Participants   & 33    & 33    & 33    & 33    & 33    \\
\bottomrule
\end{tabular}
\vspace{0.5em}

\footnotesize
\textit{Notes.} Linear mixed-effects models with participant-level random intercepts. All continuous predictors are standardized. Columns~(1)--(3) introduce two-way experience interactions separately; Column~(4) includes both significant two-way interactions jointly; Column~(5) adds the AIGCU $\times$ EL--PSC interaction and three-way term. Controls (professional experience and organizational seniority), intercepts, and fit statistics are reported in Appendix Table~\ref{tab:expertise_heterogeneity_full} (\S\ref{app:expertise-hetero}). Standard errors in parentheses. * $p \leq 0.10$, ** $p \leq 0.05$, *** $p \leq 0.01$.
\end{table}

\paragraph{Expertise Heterogeneity}\label{par:expert-heterogeneity}
The preceding analyses estimate average associations across all participants. However, CLT predicts that the same objective load conditions should affect experts and novices differently, as domain-specific schemas compress effective intrinsic load and free working memory capacity to absorb extraneous demands. We test this by introducing cross-level interactions between professional experience (a between-participant moderator) and the within-participant load and AIGCU values. 

Table~\ref{tab:expertise_heterogeneity_core} reports five specifications. Column~(1) tests whether the extraneous load-quality association varies by years of professional experience. The positive interaction term ($\beta=0.145$, $p=0.05$) indicates that the negative association between extraneous load and quality is attenuated for more experienced professionals, though the relationship is marginal in this two way specification. Column~(2) tests whether the AIGCU-quality association varies by experience. The negative interaction ($\beta=-0.121$, $p=0.015$) indicates that less experienced professionals exhibit a stronger positive association between AI content usage and quality. Column~(3) confirms that the intrinsic load-quality association does not vary by experience ($\beta=-0.004$, $p=0.930$). Column~(4) estimates both significant experience moderations jointly, without higher-order terms. Both effects strengthen when estimated simultaneously: Experience $\times$ EL increases to $\beta=0.208$($p=0.007$) and Experience $\times$ AIGCU to $\beta=-0.159$ ($p=0.002$). This specification yields the best model fit among all five columns (AIC=$4270.0$). Column~(5) adds the AIGCU $\times$ EL interaction and a three-way Experience $\times$ AIGCU $\times$ EL term for completeness. Neither reaches significance (AIGCU $\times$ EL; $\beta=0.026$, $p=0.733$; three-way: $\beta=0.048$, $p=0.531$), and the additional terms do not improve fit relative to the joint two-way specification.

Table~\ref{tab:expertise_split} (Appendix \S~\ref{app:expertise-hetero}) provides a descriptive split-sample comparison that illustrates the pattern underlying these interactions. For less experienced professionals (below-median experience $N=657$; 20 participants), extraneous load is strongly negatively associated with quality ($\beta=-0.607$, $p<0.001$) roughly three times the magnitude observed for more experienced professionals ($N=521$; 13 participants), where the association does not reach significance ($\beta=-0.220$, $p=0.135$). The AIGCU $\times$ EL-PSC interaction is null in both subsample confirming that the expertise-dependent pattern operates through the direct load-quality association rather than through differential AIGCU buffering. The AIGCU main association is positive and significant for both groups ($\beta = 0.279$ and $0.359$, both $p<0.01$), indicating that AI content usage is associated with higher quality regardless of experience level.

\subsubsection{Behavioral and Structural Features Associated with Output Quality (RQ3)}

\paragraph{Modeling Approach}
We next examine behavioral disruptions and structural complexity features a predictors of output quality using mixed-effects models with participant-level random intercepts. Table~\ref{tab:rq3_core_features} reports the consolidated RQ3 model; disaggregated feature models are reported in Supplemental Materials \S~8.3. 

\begin{table}[H]
\centering
\scriptsize
\setlength{\tabcolsep}{4.2pt}
\renewcommand{\arraystretch}{1.15}
\caption{Behavioral features associated with output quality (RQ3).}
\label{tab:rq3_core_features}

\begin{tabular}{lccc}
\toprule
 & \textbf{(1) ICL Behavioral} & \textbf{(2) ECL Behavioral} & \textbf{(3) Both} \\
 & \textbf{$N=1{,}178$} & \textbf{$N=1{,}344$} & \textbf{$N=1{,}344$}\\
\midrule

\multicolumn{4}{l}{\textit{Behavioral Features of Extraneous Load}} \\
Task Switching      &   ---     & $-0.223^{***}$      & $-0.271^{***}$ \\

Information Sprawl  & ---       & $-0.040$             & $0.064$ \\

Truncation          & ---       & $-0.076$            & $-0.051$ \\

\addlinespace
\multicolumn{4}{l}{\textit{Behavioral Features of Intrinsic Load}} \\
Element Interactivity -- Prompt     & $0.166^{***}$     & --- & $0.208^{***}$\\

Element Interactivity -- Response   & $-0.025$          & --- & $-0.002$ \\

Dependency Debt -- Prompt           & $-0.185^{**}$     & --- & $-0.020$ \\

Dependency Debt -- Response&        $-0.046$            & --- &  $-0.087^{*}$\\

\addlinespace
\multicolumn{4}{l}{\textit{Composite Load Measures}} \\
Intrinsic Load  & ---            & $-0.179^{***}$   & --- \\

Extraneous Load & $-0.568^{***}$ & ---              & --- \\

\addlinespace
\multicolumn{4}{l}{\textit{Controls}} \\
AI-Generated Content Usage (AIGCU) & $0.342^{***}$  &  $0.345^{***}$    & $0.313^{***}$ \\
Professional Experience (Years) & $-0.445^{**}$     & $-0.475^{**}$     & $-0.414^{***}$ \\
Prior Organizational Seniority & $0.316$            & $0.317$           &  $0.304 $\\

\midrule
Participants & 33 & 33 & 33\\
Estimation & ML (random intercepts) & ML (random intercepts)& ML (random intercepts) \\
\bottomrule
\end{tabular}

\vspace{0.6em}
\footnotesize
\textit{Notes.} This table summarizes the mixed-effects specifications in SI \S~8.3. Column~(1) reports the individual behavioral features of intrinsic cognitive load with the composite bi-directional extraneous load measure and controls for AIGCU, Years of Professional Experience, and Prior Organizational Seniority. We exclude coefficients for Phase Spread as it did not reach significance in any of our specifications. Column~(2) reports the individual behavioral features of extraneous cognitive load with fixed effects for the composite Intrinsic load measure. Column~(3) then reports all individual behavioral features of both intrinsic and extraneous load. \textsuperscript{***}$p\leq0.01$, \textsuperscript{**}$p\leq0.05$, \textsuperscript{*}$p\leq0.1$.
\end{table}

\paragraph{Results}

Task switching. Task switching is negatively associated with quality ($\beta=-0.271$, $p<0.01$) and is the only role level behavioral disruption that remains significant in the joint specification. When entered individually, sprawl and truncation are also negatively associated with quality (sprawl $\beta=-0.162$. $p<0.05$; truncation $\beta=-0.083$, $p\leq0.10$; Table~S10 in the Supplemental Materials), but both are absorbed once switching cost is included, consistent with shared variance among extraneous load indicators. 

Information sprawl and truncation. In the consolidated model, information sprawl is positive but not significant ($\beta=0.064$, $n.s.$) and truncation is small and not significant ($\beta=-0.051$, $n.s.$). Neither behaves as a reliable independent predictor one switching and structural features are included. 

Prompt vs.\ response structural complexity. Prompt-side element interactivity is positively associated with quality ($\beta =0.166$, $p<0.01$), whereas response-side element interactivity is not ($\beta=-0.025$, $n.s.$). Prompt-side dependency debt is negatively associated with quality ($\beta=-0.185$, $p<0.05$), while response-side dependency debt is not significant ($\beta=-0.046$, $n.s.$). Phase spread is null across all specification and is omitted from the main table (see Tables S6-S8 in the Supplemental Materials).

\subsection{Mediation via AI-Generated Content Usage}

\begin{table}[H]
\centering
\caption{Mediation effects on output quality via AI-generated content usage.}
\label{tab:mediation_quality}
\footnotesize
\setlength{\tabcolsep}{4pt}
\renewcommand{\arraystretch}{1.1}
\begin{tabular}{p{0.48\linewidth} r r r p{0.26\linewidth}}
\toprule
Effect (Outcome: Output Quality) & Est. & SE & $p$ & 95\% CI \\
\midrule
Indirect: Intrinsic Load $\rightarrow$ AI Usage $\rightarrow$ Quality
 & 0.060*** & 0.011 & $<.001$ & [0.040,\ 0.084] \\
Indirect: Extraneous Load (EL--PSC) $\rightarrow$ AI Usage $\rightarrow$ Quality
 & 0.222*** & 0.033 & $<.001$ & [0.157,\ 0.286] \\
\addlinespace
Direct: Intrinsic Load $\rightarrow$ Quality (net of AI usage)
 & $-$0.164*** & 0.044 & $<.001$ & [$-$0.260,\ $-$0.083] \\
Direct: Extraneous Load (EL--PSC) $\rightarrow$ Quality (net of AI usage)
 & $-$0.762*** & 0.067 & $<.001$ & [$-$0.893,\ $-$0.619] \\
\addlinespace
Total: Intrinsic Load $\rightarrow$ Quality
 & $-$0.105** & 0.043 & 0.014 & [$-$0.195,\ $-$0.028] \\
Total: Extraneous Load (EL--PSC) $\rightarrow$ Quality
 & $-$0.540*** & 0.058 & $<.001$ & [$-$0.649,\ $-$0.426] \\
\bottomrule
\end{tabular}
\vspace{4pt}
\scriptsize
\textit{Note.} Mediation models estimated via structural equation modeling with maximum likelihood and percentile bootstrap standard errors (1{,}000 resamples), clustered by participant. The mediator is AI-generated content usage (AIGCU). Intrinsic and extraneous load are standardized measures as defined in the main text. Indirect effects reflect the product of load $\rightarrow$ AIGCU and AIGCU $\rightarrow$ quality paths; direct effects represent associations net of AIGCU; total effects equal the sum of direct and indirect components. All continuous variables are $z$-scored. \textsuperscript{***}$p\leq0.01$, \textsuperscript{**}$p\leq0.05$, \textsuperscript{*}$p\leq0.1$.
\end{table}
Because our design is observational, the mediation estimates should be interpreted as associational decompositions rather than identified causal pathways. Although the structural equation model specifies a directional sequence (Load $\rightarrow$ AIGCU $\rightarrow$ Quality), AI-generated content usage may itself reflect unobserved participant traits—such as baseline domain competence, verification diligence, or time-pressure strategies—that jointly influence both load and output quality. Thus, the indirect effects quantify how the total association between load and quality partitions across correlated pathways, not a causal mechanism in the experimental sense.

We estimate a mediation model using Structural Equation Modeling (SEM) \cite{ullman2012structural} to examine whether AI-generated content usage is associated with the relationship between cognitive load and output quality. Table~\ref{tab:mediation_quality} reports significant positive indirect associations for both intrinsic load (indirect $=0.060$, $p<0.001$) and extraneous load (indirect $ = 0.222$, $p<0.001$) via AIGCU, indicating that higher load conditions are associated with greater AI content usage, which in turn is positively associated with output quality. 

At the same time, both load types retain negative direct associations net of AIGCU (ICL direct = $-0.164$, EL direct = $-0.762$). Thus, AI usage does not moderate or buffer the marginal effect of load-consistent with the null interaction terms reported in Table~\ref{tab:quality_interactions_linear}. Instead, the results indicate parallel additive associations: load relates negatively to quality, while load also relates positively to AIGCU, which itself relates positively to quality. 

Evaluated arithmetically, the indirect pathway attenuates approximately 37\% of the direct intrinsic-load penalty and 29\% of the direct extraneous-load penalty. This pattern reflects a countervailing associative pathway rather than a multiplicative buffering effect.

\subsection{Mediation via AIGCU and Work Experience}

\begin{table}[t]
\centering
\small
\setlength{\tabcolsep}{6pt}
\renewcommand{\arraystretch}{1.15}
\caption{Conditional indirect and total effects on output quality at $\pm 1$ SD of professional experience.}
\label{tab:conditional_effects_core}
\begin{tabular}{lccc}
\toprule
 & Low Exp ($-1$ SD) & High Exp ($+1$ SD) & $\Delta$ \\
\midrule
\multicolumn{4}{l}{\textit{Indirect effects (Load $\to$ AIGCU $\to$ Quality)}} \\
ICL
 & 0.082*** & 0.023** & $-$0.060*** \\
EL--PSC
 & 0.190*** & 0.172*** & $-$0.018 \\
\addlinespace
\multicolumn{4}{l}{\textit{Direct effects (net of AIGCU)}} \\
ICL
 & $-$0.150*** & $-$0.192** &  \\
EL--PSC
 & $-$0.909*** & $-$0.424*** &  \\
\addlinespace
\multicolumn{4}{l}{\textit{Total effects (direct + indirect)}} \\
ICL
 & $-$0.067 & $-$0.169** &  \\
EL--PSC
 & $-$0.719*** & $-$0.252*** &  \\
\bottomrule
\end{tabular}

\vspace{0.6em}
\footnotesize
\textit{Notes.} Conditional effects computed from the moderated mediation model by evaluating professional experience at $\pm 1$ SD. Bootstrap percentile 95\% confidence intervals and full path coefficients are reported in Appendix Table~\ref{tab:conditional_effects}. \textsuperscript{***}$p\leq0.01$, \textsuperscript{**}$p\leq0.05$, \textsuperscript{*}$p\leq0.1$.
\end{table}

Moderated Mediation. To examine whether professional experience alters the pathways through which cognitive load relates to quality, we estimate a moderated mediation model in which experience moderates all three links in the Load~$\rightarrow$~AIGCU~$\rightarrow$~Quality chain: the \textit{a-paths} (load~$\rightarrow$~AIGCU uptake), the \textit{b-path} (AIGCU~$\rightarrow$~quality), and the direct \textit{c'-paths} (load~$\rightarrow$~quality; net of AIGCU). Estimation results appear in Tables~\ref{tab:moderated_mediation_paths} and~\ref{tab:conditional_effects_core}.

Three moderation associations reach significance. First, professional experience moderates the EL~$\rightarrow$~AIGCU a-path ($\beta_{mod}=0.294$, $p<0.001$). More experienced professionals show a substantially larger positive association between extraneous load and AI content uptake. At -1 SD of experience, the conditional a-path is modest ($\beta=0.308$, $p<0.001$) and at +1 SD, it is nearly three times larger ($\beta=0.896$, $p<0.001$). Thus, experienced professionals increase AI uptake much more sharply as extraneous load rises. This pattern is consistent with users' possession of stronger domain-specific schemas enabling selective extraction of useful material from cluttered interactions. Professional experience however, does not moderate the ICL~$\rightarrow$~AIGCU path ($\beta_{mod}=0.008$, $p=0.757$).

Second, professional experience moderates the AIGCU~$\rightarrow$~Quality b-path ($\beta_{mod}=-0.213$, $p<0.001$) indicating that less experienced professionals exhibit a stronger positive association between AI content usage and output quality. At -1 SD of experience, the conditional b-path is $\beta = 0.618$ ($p<0.001$); at +1 SD, it shrinks to $\beta=0.192$ ($p=0.005$). Less experienced professionals appear to extract greater quality lift per unit of AI content incorporated. 

Third, experience moderates the direct EL~$\rightarrow$~Quality path ($\beta_{mod}=0.243$, $p<0.001$) indicating that the direct association between extraneous load and quality, net of any AIGCU-mediated channel, is substantially attenuated for more experienced professionals. At -1 SD, the direct EL-quality association is $\beta=-0.909$ ($p<0.001$) and at +1 SD, it is $\beta=-0.424$ ($p=0.001$). Experience does not moderate the direct ICL~$\rightarrow$~Quality path ($\beta_{mod}=-0.021$, $p=0.704$). 

Table~\ref{tab:conditional_effects_core} displays the conditional total, direct, and indirect associations at $\pm1$ SD of experience. The total associations reveal a differentiated pattern. For less experienced professionals, extraneous load is strongly negatively associated with quality ($\beta=-0.719$, $p<0.001$), while intrinsic load is not significant ($\beta=-0.067$, $p=0.215$). For more experienced professionals, the total EL-quality association is substantial ($\beta=-0.252$, $p=0.007$), while the ICL-quality association reaches significance ($\beta=-0.169$, $p=0.038$). Thus, extraneous load remains negatively associated with quality at both experience levels, but the magnitude is dramatically smaller for experienced professionals. 
The indirect associations operating through the AIGCU channel are positive and significant for both experience groups across both load types (low experience ICL: $\beta=0.082$, $p<0.001$; EL: $\beta=0.190$, $p<0.001$; high experience ICL: $\beta=0.023$, $p=0.037$; EL: $\beta=0.172$, $p=0.009$). For the ICL pathway, the difference between groups is significant ($\Delta=-0.060$, $p=0.003$), indicating a stronger compensatory channel for less experienced professionals. The EL indirect difference does not reach significance. 

The moderated model fits substantially better than the baseline mediation ($\Delta$AIC=$-94.3$, $\Delta$BIC=$-53.7$), suggesting that experience-level heterogeneity is structurally important rather than a marginal refinement. 

Taken together, these results indicate that the associations documented in the preceding sections are not uniform across experience levels. Extraneous load is primarily associated with quality deficits among less experienced professionals, for whom the direct EL-quality penalty approaches one full standard deviation of quality per standard deviation of load ($\beta=-0.909$). More experienced professionals appear substantially more resilient to extraneous load: the direct association is attenuated by more than half ($\beta=-0.424$) and, while they incorporate more AI content under load, they derive less incremental quality benefit per unit incorporated.

\subsection{Utterance-Level Analysis (RQ2)}\label{sec:rq2}

\begin{table}[t]
\centering
\small
\setlength{\tabcolsep}{5pt}
\renewcommand{\arraystretch}{1.15}
\caption{Asymmetry and temporal persistence of extraneous cognitive load in human--LLM dialogue.}
\label{tab:rq2_core_dynamics}
\begin{tabular}{lcc}
\toprule
 & $\boldsymbol{\beta}$ & \textbf{$p$-value} \\
\midrule
\multicolumn{3}{l}{\textbf{Panel A: Contemporaneous Role Asymmetry}} \\
Prompt (vs.\ Response) $\rightarrow$ Extraneous Load 
 & $-0.041$ & $0.490$ \\
\addlinespace[3pt]
\multicolumn{3}{l}{\textbf{Panel B: Cross-Speaker Spillover (Lag 1)}} \\
Response $\rightarrow$ Next Utterance EL 
 & $0.015$ & $0.668$ \\
Prompt $\rightarrow$ Next Utterance EL 
 & $0.089^{***}$ & $<0.001$ \\
$\Delta$ Prompt vs.\ Response (Lag 1) 
 & $0.074$ & $0.134$ \\
\addlinespace[3pt]
\multicolumn{3}{l}{\textbf{Panel C: Within-Speaker Persistence (Lag 2)}} \\
Response $\rightarrow$ Response 
 & $0.170^{***}$ & $<0.001$ \\
Prompt $\rightarrow$ Prompt 
 & $0.268^{***}$ & $<0.001$ \\
$\Delta$ Prompt vs.\ Response (Lag 2) 
 & $0.098$ & $0.105$ \\
\bottomrule
\end{tabular}
\vspace{0.6em}
\footnotesize \\
\textit{Notes.} Estimates are drawn from dynamic fixed-effects panel models with Driscoll--Kraay standard errors. Extraneous load is standardized at the utterance level. Panel~A reports the within-participant contemporaneous difference between prompts and responses, controlling for AR(2) dynamics, utterance length, sequence position, and relative position. Panel~B captures cross-speaker spillover from the immediately preceding utterance ($t-1$). Panel~C captures within-speaker persistence across successive utterances by the same speaker ($t-2$). Full model specifications and robustness checks appear in Appendix \ref{app:temp-dynamics}; Table \ref{tab:rq2_full_models}. \textsuperscript{***}$p\leq0.01$, \textsuperscript{**}$p\leq0.05$, \textsuperscript{*}$p\leq0.1$.
\end{table}

\subsubsection{Speaker Differences in Cognitive Load}
Table~\ref{tab:rq2rq3_paired_ttests} in Appendix~\ref{app:t-test} compares participant's mean prompt vs. response load. Unconditionally, prompts exhibit significantly lower intrinsic load ($\Delta=-0.322$, $p<0.001$) and lower extraneous load ($\Delta=-0.540$, $p<0.001$) than responses. 

However, Panel~A of Table~\ref{tab:rq2_core_dynamics} reveals that this apparent role asymmetry does not survive conditioning on temporal dynamics. After controlling for AR(2) persistence, utterance length, sequence position, and participant fixed effects, the contemporaneous prompt--response difference in extraneous load is near zero and not significant ($\beta=-0.041$, $p=0.490$). Because responses occur later in the exchange, they inherit load through autoregressive persistence rather than generating it independently. Supplementary decomposition confirms this interpretation: controlling for AR(2) persistence along-without utterance length-eliminates the speaker difference entirely ($\beta=0.007$, $p=0.96$; Table~\ref{tab:el_gap_decomposition}), whereas controlling for length alone does not attenuate the gap but reverses it, indicating that the raw asymmetry reflect temporal accumulation rather than role-linked verbosity.

\subsubsection{Spillover and Persistence}
Panels~B and~C of Table~\ref{tab:rq2_core_dynamics} decompose the temporal structure of extraneous load into cross-speaker spillover (Lag~1) and within-speaker persistence (Lag~2). 

Cross-speaker spillover is asymmetric but modest in magnitude (Panel~B). Prompt extraneous load is positively associated with the immediately following response ($\beta=0.089$, $p<0.001$), indicating that higher-load prompts tend to be followed by higher-load responses. The reverse pathway--response-to-prompt spillover--is near zero and not significant($\beta=0.015$, $p=0.668$). The difference between the two spillover coefficients is suggestive but not does reach conventional significance ($\Delta=0.074$, $p=0.134$). 

Within-speaker persistence is strong for both speakers (Panel~C). Successive prompts are positively associated ($\beta=0.268$, $p<0.001$) as are successive responses ($\beta=0.170$, $p<0.001$). Prompts show somewhat stickier persistence than responses, though the difference is not significant at conventional levels ($\Delta=0.098$, $p=0.105$). These within-speaker associations are substantially larger than the cross-speaker spillover coefficients, indicating that extraneous load in human--LLM dialogue is primarily governed by each speakers' own trajectory rather than by transmission across the conversational boundary. 

Taken together, these results characterize extraneous load dynamics in human--LLM conversation as a process dominated by within-speaker momentum. The user's load at turn $t$ is most strongly associated with their own load at turn $t-2$, not with the model's intervening response. The one reliable cross-speaker pathway runs from prompts to responses--consistent with the model's output being shaped by the structure of the user's input--but even this association is modest relative to within-speaker persistence. The absence of significant response-to-prompt spillover suggests that users do not passively inherit extraneous load from model outputs; rather, their load trajectories appear to be self sustaining.

\subsubsection{Load and Content Usage Dynamics}

\begin{table}[t]
\centering
\small
\setlength{\tabcolsep}{4pt}
\renewcommand{\arraystretch}{1.12}
\begin{threeparttable}
\caption{Cognitive load and content usage dynamics across prompts and responses.}
\label{tab:content_usage_core}
\begin{tabular}{p{0.46\linewidth} p{0.46\linewidth}}
\toprule
\textbf{Panel A: What predicts response content usage (CCU)?}
&
\textbf{Panel B: Content usage dynamics chain} \\
\midrule

\textit{Contemporaneous response load}
&
\textit{Link A: Does response usage shape next prompt load?} \\

R($t$) IL spike $\rightarrow$ R($t$) AIGCU\hfill $0.214^{**}$ (0.019)
&
R($t$) AIGCU$\rightarrow$ P($t{+}1$) EL \hfill $0.032$ (0.287) \\

R($t$) EL spike $\rightarrow$ R($t$) AIGCU\hfill $0.188^{**}$ (0.028)
&
R($t$) EL $\rightarrow$ P($t{+}1$) EL  \hfill $0.090^{**}$ (0.042) \\

\addlinespace[6pt]
\textit{Preceding prompt load}
&
\textit{Link B: Does next prompt load shape next response usage?} \\

P($t$) IL $\rightarrow$ R($t$) AIGCU\hfill $0.106^{***}$ (0.003)
&
P($t{+}1$) EL $\rightarrow$ R($t{+}1$) AIGCU\hfill $0.079^{*}$ (0.060) \\

P($t$) EL $\rightarrow$ R($t$) AIGCU\hfill $0.042$ (0.244)
&
P($t{+}1$) IL $\rightarrow$ R($t{+}1$) AIGCU\hfill $0.102^{***}$ (0.001) \\

\addlinespace[6pt]
\textit{Cross-level interaction}
&
\textit{Reduced form: AIGCUpersistence} \\

P($t$) EL $\times$ R($t$) EL spike $\rightarrow$ R($t$) AIGCU\hfill $0.011$ (0.897)
&
R($t$) AIGCU$\rightarrow$ R($t{+}1$) AIGCU\hfill $0.115^{***}$ (0.004) \\

\midrule
\multicolumn{2}{l}{\textit{$N_{\text{Panel A}} = 971$ responses; $N_{\text{Panel B}} = 1{,}008$--$1{,}079$}} \\
\bottomrule
\end{tabular}

\begin{tablenotes}[flushleft]
\footnotesize
\item \textit{Notes.} $R(t)$ denotes the model response at conversational turn $t$, and $P(t)$ denotes the user prompt at turn $t$. Thus, $P(t{+}1)$ refers to the user's subsequent prompt following response $R(t)$.

Panel~A reports response-level fixed-effects models predicting content usage (CCU) in response $R(t)$ from contemporaneous response load spikes and preceding prompt load $P(t)$, including AR(2) controls for lagged AIGCUand load. Panel~B evaluates a dynamic chain linking response $t$ to the next prompt and subsequent response: Link~A regresses prompt extraneous load $P(t{+}1)$ on preceding response usage and load; Link~B regresses next-response usage $R(t{+}1)$ on preceding prompt load; the reduced form tests direct persistence of AIGCUacross responses. 

All models include participant fixed effects with Driscoll--Kraay standard errors and controls for utterance length, sequence position, and relative position. Full specifications appear in Appendix~\ref{app:cog-load-content-usage-utt}. \textsuperscript{***}$p\leq0.01$, \textsuperscript{**}$p\leq0.05$, \textsuperscript{*}$p\leq0.1$.
\end{tablenotes}
\end{threeparttable}
\end{table}

Table~\ref{tab:content_usage_core} reports response-only fixed-effects models predicting content usage and a dynamics chain testing whether content usage feeds forward across turns. 

\paragraph{Panel A: What is associated with content usage?}

Both response-level intrinsic load spikes ($\beta=0.214$, $p=0.019$) and extraneous load spikes ($\beta=0.188$, $p=0.028$) are positively associated with content usage, indicating that responses coinciding with elevated cognitive load also tend to reflect greater incorporation of AI-generated content. Preceding prompt intrinsic load is also positively associated with content usage ($\beta=0.106$, $p=0.003$), consistent with the pattern that more complex user inputs co-occur with higher content usage in the subsequent response. In contrast, preceding prompt extraneous load shows no significant association ($\beta=0.042$, $p=0.244$), and the interaction between prompt extraneous load and response extraneous load spikes is null ($\beta=0.011$, $n.s.$). The absence of an interaction indicates that the associations between load spikes and content usage are additive: response-level load conditions and prompt-level complexity each relate to content usage independently, without compounding. 

\paragraph{Panel B: Content usage dynamics chain}

Panel~B examines whether content usage in one response is associated with content usage in the subsequent response, and whether this association operates through the user's intervening prompt load. The evidence for an EL-mediated pathway is weak. Content usage in response $t$ is not significantly associated with the user's next prompt EL (Link~A: $\beta=0.032$, $n.s.$). However, response-level EL at time $t$ is positively associated with extraneous load in the subsequent user prompt (Link~A: $\beta=0.09$, $p=0.042$), though the cross-lag association does not condition on within-speaker persistence and therefore cannot be interpreted as net spillover in the sense estimated in Table~\ref{tab:rq2_core_dynamics}.\footnote{We note that our purpose here is not to replicate Table~\ref{tab:rq2_core_dynamics} spillover test but to test whether AI-generated content usage's dynamics chain operates through prompt Extraneous load as an intermediary. Thus, the $R(t)EL$ coefficient should be considered as a covariate in that chain and not the primary estimand. }

Despite the null indirect pathway through prompt EL, content usage exhibits robust direct persistence across consecutive responses (reduced form $\beta=0.115$, $p=0.004$). This pattern is consistent with content usage following its own temporal trajectory rather than operating through the extraneous load dynamics documented in Table~\ref{tab:rq2_core_dynamics}. Higher content usage in one response is associated with higher content usage in the next, independently of intervening load conditions.

\subsubsection{Measurement Validation and Robustness}\label{res:measurement_validation}
To confirm that our extraneous load composite captures structural processing costs rather than semantic diversity alone, we construct an alternative measure-\textit{concept spread}-that captures the diffuseness of the embedding space after residualizing on content volume. As detailed in Appendix~\ref{app:discrim_validity}, the extraneous load composite remains distinct from concept spread and related semantic-density measures, indicating that it reflects organizational and coordination demands rather than merely topic breadth or lexical diversity. 

\paragraph{Convergent and Construct Validity}
We additionally assess convergent validity by correlating the extraneous load composite with a comprehensive battery of established readability and text-complexity indices, as well as an LLM-based cognitive load estimation procedure (Appendix~\ref{app:construct_validation}). The composite exhibits moderate correlations in the expected direction-positive with grade-level and difficulty indices and negative with reading ease-while remaining meaningfully distinct from surface-level linguistic metrics. This pattern supports the interpretation that the composite captures processing burden associated with information organization and interaction dynamics, rather than simple sentence complexity. 

\paragraph{Pipeline Robustness}

Finally, because the extraneous and intrinsic load measures are constructed from a multi-stage computational pipeline, we evaluate their robustness to modeling assumptions using a specification curve analysis over 86 parameter combinations spanning subtask-memory dynamics, concept anchoring thresholds, coherence weights, and attribution hyperparameters (See Table S2 in Supplemental Materials). 

Across specifications, the estimated associations between extraneous load and output quality remain consistently negative-as do those for intrinsic load and quality-with no sign reversals observed. Likewise, the association between AI-generated content usage and quality remains positive across parameterizations. Although coefficient magnitudes vary modestly across configurations, statistical significance and directional consistency are preserved throughout the parameter space. These results indicate that the reported findings are not driven by a narrow set of modeling assumptions, but instead reflect stable patterns that persist across reasonable alternative operationalizations.

\section{Discussion}

Studies of human-AI interaction often default to a narrative that attributes impaired performance to model behaviors---without accounting for how users influence, and are influenced by, the dynamics of the conversation itself. While it has been shown that LLMs get lost in multi-turn conversations \cite{laban2025llms}, we too often fail to account for the fact that users follow them willfully---and may even inadvertently direct them---down the ``wrong path.''

Our process perspective directly interrogates this entanglement and yields a finding that challenges the prevailing narrative. The apparent speaker asymmetry in extraneous load---model responses exhibiting higher load than user prompts---collapses entirely once autoregressive dynamics are accounted for (Table~\ref{tab:rq2_core_dynamics}; Panel A).\footnote{ This analysis uses the utterance-level EL variant described in \S~\ref{subsubsec:anchored-coherence} that excludes rolling contextual memory to avoid mechanically inducing the persistence it measures. } The raw difference reflects temporal accumulation, not an inherent property of speaker role. Load trajectories are instead dominated by within-speaker momentum: once elevated, they tend to persist primarily through each speaker's own prior organizational structure, with only limited short-run adjustments to the other's behavior. This pattern points to an interactionally accumulated process rather than a unilaterally imposed one. In other words, the conversation's cognitive burden is not something the model \textit{inflicts} on the user rather, it is something both parties construct together, turn by turn. 

One plausible interpretation of this mutual persistence is that it reflects a misalignment between the objective function used to optimize LLM behavior and the broader goal of minimizing cognitive coordination costs in multi-turn interactions. Contemporary Reinforcement Learning from Human Feedback (RLHF) pipelines optimize models to maximize perceived helpfulness, relevance, and responsiveness to the current prompt, based on human preference rankings \cite{christiano2017deep, stiennon2020learning, ouyang2022training}. Here, the learned policy is rewarded for producing outputs that closely track the structure and intent expressed in the immediate input. 

Under such an objective, the optimal response is likely one that mirrors and elaborates on whatever conceptual organization the user provides, rather than one that restructures or simplifies it. This aligns with the prompt-to-response spillover we observe: response-level extraneous load co-varies with prompt-level load, consistent with a model that amplifies or propagates the user's informational structure. Importantly, spillover is unlikely to be the sole source of clutter and incoherence. Beyond immediate mirroring, responses exhibit persistence across turns, suggesting that once a conversational trajectory becomes complex, it may remain so. Rather than independently generating clutter de novo, the model may inherit, reinforce, and stabilize emergent conversational complexity over time.

Users, in turn, may adapt to such responsiveness. Prior work characterizes prompts as a "control surface" in task-oriented dialogue, through which users iteratively add constraints, examples, and contextual qualifiers to steer the model's behavior \cite{zamfirescu2023johnny, desmond2024exploring}. Under this view, increasing prompt specificity can become a strategy for exerting control when models reliably elaborate the structure provided. Such iterative elaboration departs from the typical human tendency toward underspecified communication, where shared context allows efficient compression \cite{piantadosi2012communicative, laban2025llms}.

Empirical evidence reinforces this broader pattern. Rewriting prompts to increase specificity improves response quality \cite{sarkar2025conversational}, yet large-scale analyses reveal that many real-world prompts remain underspecified \cite{kopf2023openassistant}, and models often respond directly rather than request clarification \cite{herlihy2024overcoming}. Taken together, these findings suggest an interaction dynamic in which specificity is rewarded, underspecification is tolerated, and clarification is under-incentivized. The result is a conversational regime that encourages increasingly elaborate prompts without systematically pruning irrelevant structure or attenuating accumulated complexity in subsequent responses. 

The spillover analysis reported above addresses a specific question: when the model produces a high-load response, do users reorganize their next prompt to to reduce clutter? The null response$\rightarrow$prompt spillover in utterance-level extraneous load suggests that such turn-by-turn response-conditioned pruning of clutter is not reliably observed along the dimensions we measure. We note that an unconditional cross-lag association between response EL and subsequent prompt EL is positive and significant in specifications that do not condition on within-speaker persistence (Table~\ref{tab:content_usage_core}, Panel B, Link A; $\beta=0.09$, $p=0.042$), but this association is absorbed once autoregressive prompt-to-prompt dynamics are accounted for (Table~\ref{tab:rq2_core_dynamics}, Panel B), indicating that the apparent transmission reflects the user's own entrenched load trajectory rather than a response-conditioned adjustment. This does not imply that users fail to adapt altogether. Rather, adaptation may operate at a broader strategic level-through cumulative increases in specificity or constraint-setting-without manifesting as consistent micro-adjustments to transient shifts in model-induced complexity.

This matters because ambiguity is adaptive in human communication. When context is informative, ambiguity allows reuse of efficient linguistic units and reduces cognitive effort without compromising comprehension \cite{piantadosi2012communicative}. Speakers tailor utterances to what is sufficient for grounding, relying on shared context rather than exhaustive specification \cite{clark1991grounding}, thereby avoiding information density that overloads comprehension \cite{jaeger2010redundancy}. The absence of response-to-prompt spillover in our data suggests that model responses do not function as effective grounding signals for users--users' organizational patterns are not responsive to what the model produces, only to their own prior behavior. In human dialogue, speakers continuously adjust based on listener feedback \cite{clark1991grounding}; in human-LLM dialogue, this adjustment channel appears to be largely absent, replaced by self-sustaining trajectories on both sides of the conversation. 

But when users communicate in these non-human-like ways–with high specificity and exhaustive enumeration–relevance-optimized models match that specificity in their responses. The within-speaker persistence and prompt-to-response spillover patterns we observe are consistent with a self-reinforcing cycle: as organizational complexity in prompts accumulates, the model accommodates rather than simplifies, and the user receives no signal that a different communication pattern may be more productive. Optimizing for human-like model behavior on one dimension (relevance) may ironically produce outputs incongruent with human cognitive processing precisely when users' own behavior has drifted furthest from natural communication norms.

Our expertise heterogeneity results sharpen this concern. Less experienced professionals are most vulnerable to extraneous load (Table~\ref{tab:conditional_effects_core}; $\beta_{total}=-0.719$) and derive the largest marginal quality gains per unit of AI generated content incorporated ($b=0.618$ vs $0.192$; Table~\ref{tab:conditional_effects}, Panel B), yet it is more experienced professionals who increase the uptake of AI content most sharply as load rises ($a=0.896$ vs $0.308$; Table~\ref{tab:conditional_effects}, Panel A). This dissociation suggests that the users who would benefit most from compensatory AI content are not those who most readily increase their uptake under load-underscoring the need for load-dampening interventions calibrated to user expertise where possible.

\subsection{Conditional Benefits of Using AI in Complex Work}

Our first research question asked how extraneous cognitive load arising from LLM interactions shapes users' ability to benefit from AI-generated content. Across specifications, AIGCU exhibits a robust positive association with output quality (Table~\ref{tab:rq1_core_models}), while extraneous load shows the largest negative association---roughly three times the magnitude of intrinsic load---and these patterns persist after controlling for participant background (Table~\ref{tab:rq1_core_models}, Model~3). This is consistent with CLT's prediction that extraneous load is the more consequential design target, as intrinsic load reflects irreducible task demands while extraneous load reflects processing costs that could, in principle, be mitigated through better information design \cite[e.g.,][]{ dell2023navigating, wiles2024genai, prasad2024towards}.

From a Cognitive Load Theory perspective, the magnitude of the gap between intrinsic and extraneous load is particularly consequential. While intrinsic load reflects the baseline demands of the valuation task, extraneous load arises from the coordination work required to interpret, evaluate, and integrate model-generated outputs across turns. In a conversational AI setting, these coordination demands are endogenous to the interaction itself. Our results therefore suggest that performance decrements stem less from the inherent complexity of the task and more from the structural and informational dynamics of the dialogue through which assistance is delivered.

At the same time, mediation analyses complicate a simple "load is harmful" interpretation. The indirect pathway through AIGCU attenuates approximately 37\% of the direct intrinsic-load penalty (Table~\ref{tab:mediation_quality}: $0.060/0.164$) and 29\% of the direct extraneous-load penalty (Table~\ref{tab:mediation_quality}: $0.222/0.762$) indicating partial but incomplete compensation. One plausible interpretation is that incorporating AI-generated content functions as a form of cognitive offloading: by externalizing portions of the reasoning to the model, users may attenuate some of the working-memory demands associated with organizing and producing complex output---an interpretation consistent with emerging evidence that reliance on LLMs can shift cognitive effort~\cite{kosmyna2025your}. But the direct penalty imposed by extraneous load is large enough that the net association remains strongly negative. 

The expertise heterogeneity results sharpen this pattern: less experienced professionals exhibit larger per-unit returns to AI-generated content use (Table~\ref{tab:conditional_effects} Panel B; $b = 0.618$ vs.\ $0.192$) and a substantially larger direct extraneous-load penalty ($\beta = -0.909$ vs.\ $-0.424$). However, the overall compensatory channel-the full indirect effect ($a \times b$ is similar across groups ($0.190$ vs. $0.172$). This reflects offsetting differences: experienced professionals increase AI uptake more strongly under load (larger $a$-path), whereas less experienced professionals derive greater marginal benefit per unit incorporated (larger $b$-path). Thus, the groups differ in how the indirect pathway operates, not in its overall magnitude.

Taken together, these results refine prevailing narratives about AI augmentation. Rather than a uniformly positive association with performance, AI assistance operates alongside---not in interaction with---the cognitive conditions of the conversation. This additive structure may help reconcile conflicting findings in prior HCI and CSCW research~\cite{kraus2020effects, kraus2020diy, kraus2021trust}, where the determining factor may not be the presence of assistance per se, but whether the parallel accumulation of extraneous processing demands outweighs the benefits that assistance provides. The finding underscores the importance of shifting analytical focus from \textit{whether} AI-generated content usage helps to \textit{when and under what conditions} it does so. However, these results treat cognitive load largely as a contextual condition surrounding AI use; they do not yet address how such conditions arise, persist, or evolve over the course of extended interaction.

\subsection{Asymmetric Cognitive Load Dynamics in Human-LLM Interaction}
Our second research question asked how cognitive load evolves between users and LLMs during extended task-oriented conversations. As reported above, the apparent speaker asymmetry in extraneous load collapses once autoregressive dynamics are accounted for (Table~\ref{tab:rq2_core_dynamics}, Panel A), revealing a temporal structure dominated by within-speaker persistence and modest, unidirectional cross-speaker spillover. Supplementary decomposition confirms that this attenuation is driven by autoregressive dynamics rather than response verbosity (Table~\ref{tab:el_gap_decomposition}): controlling for AR(2) persistence alone eliminates the speaker difference entirely ($\beta=0.007$, $p=0.96$), whereas controlling for utterance length without AR terms does not attenuate the gap but reverses it ($\beta=0.376$, $p=0.03$). Because longer utterances will naturally reference more subtasks and concepts, length partly reflects the informational density through which extraneous load operates rather than serving as an independent confound; the reversal indicates that once volume differences are removed, prompts are organizationally no simpler than responses. The raw speaker asymmetry thus reflects temporal accumulation--responses inherit load from prior conversational dynamics-rather than an inherent property of speaker role. The one reliable cross-speaker channel runs from prompts to responses ($\beta=0.089$, $p<0.001$); the reverse pathway is near zero and not significant. 

This finding extends Cognitive Load Theory by suggesting that extraneous load in human-LLM interactions is not solely a property of information presentation at any single moment, but a path-dependent feature of the interaction. The strong within-speaker persistence means that organizational patterns---whether productive or disruptive---tend to entrench over successive turns. The absence of reliable cross-speaker correction is particularly consequential. In human dialogue, speakers continuously adjust based on listener feedback~\cite{clark1991grounding}; in human-LLM dialogue, this bidirectional adjustment appears largely absent. The model does not reorganize a user's sprawling prompt into a more coherent response, and the user does not restructure their approach in response to model output. Each party continues along its own trajectory, and the conversation's cognitive demands reflect the accumulation of these independent paths.

These dynamics clarify why extraneous cognitive load exhibits such a strong and persistent association with performance deficits in RQ1. Load does not operate as an isolated, turn-level cost; it accumulates within speakers and, once elevated, tends to persist. This extends prior work on timing and interruptions in proactive systems by showing that misaligned interactions may reshape subsequent conversational trajectories, not merely disrupt a single moment. Designing effective proactive systems requires not only predicting when to intervene, but also identifying when load trajectories are becoming self-sustaining and actively restructuring the interaction to attenuate accumulated complexity.

\subsection{Behavioral Sources of Extraneous Load}

RQ3 asked which concrete LLM behaviors and interaction patterns contribute most to extraneous cognitive load and its downstream performance costs. The results point to a clear and consistent conclusion: not all proactive behavior is equally costly, and the most damaging forms of proactivity are those that disrupt users' mental model of the task rather than enrich it.

Among all behavioral indicators examined, unsolicited task switching by the model emerges as the strongest and most robust negative predictor of output quality. When the model introduces new subtasks that are not anchored in the user's immediately preceding focus, performance declines substantially---even after controlling for intrinsic task complexity, overall usage of AI-generated content, and participant background characteristics. This association persists in joint models alongside other behavioral indicators, suggesting that task switching captures a distinct and substantively important source of cognitive disruption rather than serving as a proxy for verbosity or information volume.

By contrast, information sprawl exhibits weaker and less stable effects once switching is accounted for. While sprawl appears harmful in isolation, its association is largely absorbed by correlated behaviors---particularly switching---when multiple features are modeled simultaneously. This distinction is important: it suggests that the problem is not necessarily the amount of information the model provides, but the organizational consequences of introducing that information in ways that expand or redirect task space. In some cases, additional information may even support higher-quality outputs---either by deepening coverage within an already relevant subtask, or because it represents a more ``complete'' analysis that users can readily take advantage of with minimal revisions. Similarly, truncation---instances where model responses fail to address the user's prompt---does not reliably predict output quality. While truncation may be frustrating or inefficient, it does not appear to impose the same sustained cognitive coordination costs as behaviors that actively reshape the task landscape.

The intrinsic load features-element interactivity and dependency debt, decomposed by speaker role-further refine this picture. Prompt-side element interactivity---reflecting users' mentioning of subtasks that are more closely related in the task decomposition---is positively associated with output quality, whereas its response-side counterpart is neutral or weakly negative. 

A similar asymmetry appears for dependency debt: when users reference subtasks without having recently engaged with their prerequisites, output quality tends to decline reliably, but response-side dependency debt does not exhibit a comparable association. Notably, this prompt-response asymmetry cuts across construct boundaries, appearing in intrinsic load features (element interactivity, dependency debt) as well as the extraneous load indicators where task switching dominates. The common pattern is not specific to either load type but rather to the locus of disruption: performance is most sensitive to the organizational state the user brings to exchange. When users' own cognitive maps are fragmented-whether through intrinsic complexity they have not adequately scaffolded, or through extraneous misalignment introduced by the system-quality suffers. Response-side complexity, by contrast, does not exhibit comparable independent associations with performance, suggesting that it is the interference with the user's active task representation, rather than the informational content of the response per se, that drives quality deficits.

From a theoretical perspective, these results extend Cognitive Load Theory by demonstrating that extraneous load in human-AI interactions is not only a function of presentation quality, but also of initiative misalignment. Proactivity becomes costly when the system takes initiative in ways that violate the user's current mental model of the task, introducing new elements that must be reconciled in working memory. Conversely, when users themselves introduce this complexity, that complexity is often integrated coherently and supports performance.

This distinction is important. It suggests that improving proactive dialogue systems is not simply a matter of reducing information density or limiting assistance, but of constraining when and how systems take initiative. Proactive behaviors that respect the user's current focus and cognitive map of the work---reinforcing, elaborating, or deepening an active subtask---are associated with far less cognitive disruption than those that redirect attention or expand scope. In light of the within-speaker persistence documented in RQ2, violations of task alignment need not be dramatic to be consequential. The strong autoregressive structure of extraneous load suggests that even modest increases tend to persist across subsequent turns rather than dissipating quickly. The organization of communications, once established, is sticky. Designing for effective proactivity therefore requires systems that are sensitive not only to task goals, but to the cognitive schema that users are invoking and maintaining during task performance.

\subsection{Implications for the Design of Proactive Systems}
Taken together, our findings suggest a path towards precision proactivity: dialogue systems that provide targeted assistance calibrated to users' cognitive states and task representations, rather than indiscriminately expanding information or initiative. Across RQ1-RQ3, the central design challenge is not how to increase the amount or sophistication of assistance, but how to regulate \textit{when} and to \textit{what extent} systems intervene so as to minimize extraneous cognitive load while preserving the compensatory benefits of AI-generated content. Below, we outline three implications for the design of proactive systems that follow from the cognitive and interactional patterns identified in our results.

\subsubsection{Detect and Respond to Elevated Cognitive Load}
Proactive systems should be designed to detect conversational signals of elevated cognitive load and respond by \textit{dampening}, rather than amplifying, cognitive demands. Our results show that extraneous load is strongly associated with degraded performance and that load levels exhibit strong within-speaker persistence--once a speaker's communication patterns skew towards higher extraneous load, they tend to remain elevated across subsequent turns. The modest prompt-to-response spillover (Table~\ref{tab:rq2_core_dynamics} Panel B; $\beta=0.089$, $p<0.001$) further indicates that the model's output tends to reflect the structure of the user's input rather than reorganizing it.

This pattern highlights a mismatch between current LLM behaviors and conversational norms. In human-human dialogue, turn-taking implicitly constrains the rate and volume of information exchange, reflecting shared assumptions about limited processing capacity \cite{clark1991grounding}. In contrast, LLMs are capable of responding exhaustively to long or complex prompts, and users appear to anticipate and encourage this behavior. While such responsiveness may be desirable from an information standpoint, our findings indicate that it is associated with substantive extraneous load as users are confronted with large quantities of information that they must integrate, filter, and manage. 

Designing for precision proactivity therefore requires systems to treat high-load prompts as signals for structural support rather than opportunities for maximal elaboration. When indicators of cognitive strain are present, systems should prioritize simplification, sequencing, and clarification--restricting the rate of information transmission and focusing on one or two focal issues at a time. This principle has direct antecedents in CLT-informed instructional design, where segmenting and sequencing complex material reduces extraneous load by allowing learners to process one element before integrating the next \cite{mayer2001learning}. By behaving more like a cognitively aware conversational partner--one that actively regulates information flow--systems may help interrupt the within-speaker persistence patterns that our results show are associated with degraded performance.

Our expertise heterogeneity findings further sharpen this recommendation. Less experienced professionals are most vulnerable to extraneous load (Table~\ref{tab:conditional_effects_core}; $\beta_{total} = -0.719$) and most reliant on the use of AI-generated content to compensate, suggesting that load-dampening interventions should be calibrated to user expertise where possible. However, this calibration must be dynamic rather than static. CLT research on the expertise reversal effect demonstrates that instructional supports that are beneficial for novices can become redundant and even counterproductive for more experienced performers, who already possess schemas that render explicit guidance extraneous \cite{kalyuga2003expertise}. A system that uniformly dampens output for all users risks imposing unnecessary simplification on experts--itself a form of extraneous load. Precision proactivity thus requires not only detecting elevated load, but modeling user expertise to determine the appropriate level and form of intervention.

This expertise sensitivity also carries implications for learning. CLT was developed as a theory of instructional design, and its core prediction is that extraneous load impairs not only immediate performance but schema acquisition--the process by which novices develop expert-like mental models through practice \cite{sweller1994cognitive, vanmerrienboer2005cognitive}. If the self-sustaining load patterns we document prevent novice users from constructing coherent task schemas during AI-assisted work, the long-term consequence may extend beyond diminished output quality to impaired skill development. Users who most need AI assistance to compensate for limited expertise may simultaneously be those for whom persistent extraneous load most impedes the learning that would eventually reduce their dependence on that assistance. This possibility warrants direct investigation: longitudinal studies tracking schema acquisition under varying load conditions during AI-assisted work could clarify whether precision proactivity serves not only immediate performance but durable skill formation.

\subsubsection{Constrain System-Initiated Scope Expansion Through Explicit Alignment}
Our results indicate that extraneous cognitive load is associated less with the sheer amount of information provided than with \textit{system-initiated expansion of task scope}. Among the behavioral features examined, unsolicited task switching-which is characteristic of proactivity-exhibits the strongest and most negative association with output quality, whereas information sprawl does not show an independent association once switching is accounted for. Importantly, this distinction suggests that the primary cognitive cost lies not in depth or completeness \textit{per se}, but in requiring users to reorient their attention and manage newly introduced dependencies. 

From a Cognitive Load Theory perspective, these scope expansions are associated with extraneous load because they require users to perform boundary management in addition to task execution \cite[i.e.,][]{brazzolotto2022interruptions, trafton2007task, wylie2000task}. While users often introduce complexity productively--by deepening analysis or integrating related subtasks--complexity introduced by the system is less likely to align with the user's current phase of work. As a result, even well-intentioned or "correct" expansions may be associated with lower performance by fragmenting attention and increasing coordination demands.

Based on this asymmetry, we propose that proactive systems should constrain system-initiated scope expansion by default, maintaining tight alignment with the user's active subtask unless explicitly invited to broaden. One architectural instantiation is a \textit{subtask focus stack}: the system maintains a running model of the user's active subtask set, inferred from the last $k$ user turns and any explicitly confirmed goals. At generation time, candidate response content is classified against subtasks in the focus stack; content falling outside this set is flagged as out-of-scope. When out-of-scope content exceeds a threshold, the system shifts from elaboration mode to \textit{boundary signaling mode} by providing a tight answer scoped to the active subtask, then surfaces deferred topics as optional expansions.

A complementary mechanism is \textit{explicit scope elicitation}--systems that actively query users to confirm or narrow the task scope before generating responses, rather than inferring scope from prompt structure alone. Our load dynamics results provide a strong rationale for this approach: we show that models optimized for relevance tend to mirror the organizational structure of user prompts rather than reorganizing them, and that users' own organizational patterns are unresponsive to model outputs (null response-to-prompt spillover). This means that neither party currently functions as a reliable corrective force when task scope drifts. Explicit scope confirmation introduces a structured checkpoint that can interrupt scope expansion \textit{before} it enters the response, rather than relying on users to filter it \textit{after}.

Prior work on mixed-initiative interaction has established that systems which negotiate initiative-alternating between leading and following based on task state-outperform those that adopt fixed interaction postures \cite{horvitz1999principles}. Explicit scope elicitation extends this principle by treating initiative negotiation not as a global mode but as a turn-level mechanism triggered by indicators of scope alignment. In increasingly agentic contexts, where models routinely ask clarifying questions before acting, this approach is already emerging as a practical pattern-our findings suggest a cognitive load basis for formalizing it as a default system behavior. 

At the same time, explicit elicitation is not costless. Clarifying questions may be experienced by the user as cumbersome interruptions that can temporarily increase cognitive load-particularly when they lack appropriate specificity \cite{zargham2022understanding, rahmani2024clarifying}. The design challenge, therefore, is not simply to increase clarification, but to time and structure it so that short-term disruption prevents longer-term entrenchment of complexity. 

\subsubsection{Enable Structured Disclosure Through User-Directed Navigation}
Finally, our findings suggest that proactive systems must carefully manage \textit{depth} without inadvertently triggering the scope shifts that are associated with extraneous load. While response-side complexity and truncations are not independently associated with poorer outcomes, users' incorporation of AI-generated content increases precisely during moments of heightened cognitive demand. This indicates that depth and detail remain valuable resources-particularly under strain-but only when they are delivered in ways that do not require users to reorganize their cognitive models of the task. 

Progressive disclosure offers a principled approach to resolving this tension \cite{nielsen2006progressive}. However, in conversational AI, progressive disclosure need not be limited to shorter initial responses followed by elaboration upon request. Decades of research on web navigation and information architecture show that hierarchical structuring, hyperlinking, and selective expansion reduce cognitive load by allowing users to regulate information intake and avoid unnecessary integration costs \cite{pirolli1999information, card2009information}. Our findings suggest a similar redesign opportunity: structuring responses as \textit{navigable information architectures} rather than monolithic text blocks. 

Concretely, system responses could separate core, in-scope content from optional elaborations that users selectively access-through expandable sections, in-line annotations, or branching structures that allow deeper engagement without \textit{imposing} breadth. Such interaction patterns mirror hyperlinked environments such as wikis and documentation systems, where users actively control traversal paths rather than consuming information linearly \cite{pirolli1999information}. By decoupling the availability of information from its imposition, these architectures preserve AI's compensatory role under cognitive load while giving users control over scope and pacing. 

Moreover, discrete navigation choices also create informative behavioral signals \cite[e.g.,][]{hu2008collaborative, joachims2002optimizing}. Selective expansion, link traversal, and optional deep dives can serve as lightweight preference elicitation mechanisms and reveal user's evolving knowledge state and task trajectory-providing structured feedback that sequential prompt-response interaction currently lacks \cite{corbett1994knowledge, park2021scalable}. In this sense, navigable architectures not only reduce imposed cognitive load but also create opportunities for adaptive alignment over time. 

Sequential dialogue currently forces a binary choice: either the system provides comprehensive responses (risking scope expansion and extraneous load) or it provides minimal ones (sacrificing the compensatory benefit). Structured disclosure dissolves this trade-off by allowing comprehensive information to coexist with focused presentation--the system generates depth, but the user decides when and whether to access it, more closely approximating how people browse the web and use reference materials \cite{pirolli2007information}. Importantly, the design goal is not minimalism but \textit{cognitive alignment}: ensuring that users can benefit from AI-generated content at moments of high demand without reinforcing the self-sustaining load patterns that are associated with lower quality outcomes. Optional elaborations--such as assumptions, worked examples, or edge cases--should deepen understanding of the current subtask rather than introduce new ones by default. In this way, structured disclosure supports a more iterative, top-down interaction in which users progressively refine their engagement with AI-generated content rather than processing it exhaustively on receipt.N

\subsection{Theoretical Contributions}
This work makes four primary theoretical contributions to research on cognitive load, proactive systems, and AI-assisted knowledge work. Together, these contributions reframe how cognitive load operates in human-LLM interaction and clarify why proactivity can both enable and undermine performance depending on how it is deployed.

First, we extend Cognitive Load Theory by conceptualizing extraneous cognitive load as a dynamic, temporally persistent property of interaction rather than a static feature of tasks or instructional materials. Classical formulations of CLT distinguish intrinsic load, which is inherent to task complexity, from extraneous load, which arises from suboptimal information presentation or instructional design. Our findings refine this distinction by showing that, in conversational AI settings, extraneous load is not simply imposed at isolated moments but exhibits strong within-speaker persistence, with modest and unidirectional cross-speaker influence (Table~9). Extraneous load is therefore not only a function of how information is presented at a given moment, but of how each speaker's patterns of organizing information evolve over the course of the interaction. Once either speaker's load trajectory becomes elevated, it tends to remain elevated through its own momentum, and the absence of reliable cross-speaker correction means that neither party serves as a natural check on the other.

Second, we reconceptualize proactivity in dialogue systems as a state-dependent phenomenon rather than a sequence of isolated system actions. Prior HCI and CSCW work often treats proactive interventions as discrete events--interruptions, suggestions, or recommendations--whose effects can be evaluated independently. Our results complicate this view by showing that the associations between proactive behaviors and performance vary with ongoing cognitive states. Unsolicited expansions of scope or elaborative responses delivered during elevated-load moments are not merely associated with momentary disruption; given the strong within-speaker persistence we document, organizational patterns established at any point in the conversation tend to carry forward into subsequent turns through autoregressive momentum. This reframing shifts theoretical attention away from \textit{when} systems should intervene to \textit{how interventions interact with evolving cognitive conditions}. Proactivity, in this view, should be evaluated not only by its immediate informational value, but by whether it is associated with patterns of communication that persist or dissipate over subsequent turns.

Third, we identify task-structure alignment as a central theoretical mediator of when AI assistance helps versus harms performance. Across our analysis, the most damaging forms of proactivity are not those that merely increase information volume, but those that violate users' internally maintained task representations–particularly through unsolicited task switching. Conversely, complexity introduced by users themselves, such as prompt-side element interactivity, is often productive and positively associated with output quality. This asymmetry suggests that the cognitive cost of AI assistance hinges on whether new information reinforces or disrupts the task schema users are actively constructing. By foregrounding task-structure alignment, our work offers a unifying theoretical lens that connects Cognitive Load Theory with research on mixed-initiative interactions: AI systems support performance when they deepen or elaborate within an existing task schema, but impose extraneous load when they force users to reorganize that schema mid-stream. 

Fourth, our findings contribute to the literature on in-context learning in multi-turn conversations by foregrounding its cognitive implications for users. In-context learning enables LLMs to adapt to novel tasks at inference time by conditioning on examples and dialogue history, without parameter updates \cite{brown2020language, hegde2025factors}. In multi-turn settings, this accumulated context supports flexible, context-sensitive responses. Recent work, however, suggests that as conversational histories lengthen, models can become anchored to earlier and potentially "stale" information, leading to drift and structural disorganization \cite{laban2025llms}. Our results complement this model-centric perspective by highlighting a parallel human-side tension. That is, users may exhibit \textit{contextual inertia} - a growing reluctance to reset a conversation because accumulated context carries both informational value and coordination costs. Although preserving context avoids the need to restate objectives, it also increases the likelihood that model responses reflect organizational patterns that increase extraneous load over time. As a result, users who seek to benefit from contextual continuity may instead expend additional cognitive effort interpreting and managing increasingly burdensome responses. 

Taken together, these contributions move beyond debates about whether AI assistance is beneficial in general and toward a more conditional, mechanistic account of \textit{why} and \textit{when} it is effective. Rather than treating cognitive load, proactivity, and AI usage in isolation, our results show that their joint dynamics-mediated through compensatory uptake pathways, moderated by expertise, and shaped by path-dependent load trajectories-define the cognitive envolope within which AI-generated content can meaningfully augment human work. 

\subsection{Construct Validity of Transcript-Based Load Indicators}
Our intrinsic and extraneous load indicators are grounded in Cognitive Load Theory and operationalized through computational proxies-semantic coherence, embedding similarity, procedural relatedness-rather than established cognitive load instruments such as NASA-TLX self-report \cite{hart1988development}, dual-task paradigms \cite{tomporowski2020cognitive}, or physiological measures \cite{rayner1998eye}. As detailed in Appendix~\ref{app:construct_validation} and \ref{app:discrim_validity}, we provide preliminary convergent and discriminant validity analyses, demonstrating moderate alignment with established readability and text-complexity metrics while remaining distinct from surface-level linguistic density measures. Nonetheless, these indicators should be interpreted as theory-grounded estimates rather than fully validated instruments of cognitive load. Future work should further triangulate transcript-based indicators with complementary signals (e.g., post-task workload ratings, secondary-task performance, physiological measures) to strengthen convergent validity and refine measurement sensitivity.

\section{Limitations and Future Work}
Several limitations of this study warrant acknowledgment. 
First, our study employs an observational design that analyzes naturally occurring conversational data within a standardized task setting. While this approach provides high ecological validity by capturing how professionals use AI in a realistic task, it is fundamentally correlational.
Consequently, while we can identify strong associations, we cannot make definitive causal claims. Although intrinsic load (ICL) is included in our models as a proxy for subtask difficulty, unobserved or imperfectly measured aspects of difficulty-such as within-subtask instance variation, participant-specific effective difficulty, or measurement error in ICL-may still confound the observed associations. For example, residual difficulty not captured by our ICL proxy could influence both user confusion and the model's propensity to introduce topic shifts.

Second, while our operationalizations of intrinsic and extraneous cognitive load are firmly grounded in Cognitive Load Theory \cite{sweller1994cognitive, sweller1994some}, they rely on computational proxies derived from conversational traces rather than traditional self-report or dual-task measures. This approach enabled fine-grained, large-scale analysis but should be validated against established cognitive load assessments in future work. 

Third, our study focused on a single, complex task-financial valuation, performed by a relatively small sample of finance professionals under time pressure. Although this setting was intentionally chosen to approximate cognitively demanding real-world work, the generalizability of our findings to other domains (e.g., software engineering, healthcare) or to less time-constrained contexts remains to be tested. 

Fourth, while we captured full transcripts of human-AI interaction, we cannot fully disentangle whether observed effects of extraneous load arose from model outputs themselves, from user interpretations and repair strategies, or from broader contextual factors such as prior knowledge and task familiarity. 

Fifth, data were collected using GPT-4o, which is less capable than contemporary models. However, the RLHF training paradigm and text-based dialogue structure remain broadly similar across current systems, and recent evidence suggests that the dynamics we identify may intensify rather than diminish with increased model capability \cite{ranganathan2026ai, storey2026cognitive}.

Finally, we suggest that proactivity is a user-state dependent capability, but our data outlines quite a broad suite of model behaviors that involved sharing unsolicited task-related information. Subsequent experimental studies can parse and manipulate these behaviors more precisely to help us refine the concept and related measures. 

Future research should therefore extend this work in three directions. First, by triangulating our transcript-based measures with complementary methods such as secondary-task performance, physiological signals, or subjective ratings, researchers can strengthen validity and refine measurement sensitivity. Second, examining additional domains and user populations will help establish boundary conditions for when proactivity benefits or burdens human cognition. Third, controlled interventions that systematically vary timing, content, and form of proactive behaviors can test the design strategies we propose—particularly those aimed at suppressing harmful task switching and managing cascades of extraneous load.

\section{Conclusion}
To examine how proactive AI behaviors shape real-world task performance, we developed a portable, transcript-based measurement framework that estimates cognitive load dynamics from human-AI conversations at scale. The framework—combining semantic coherence measurement, dependency tracking, and load propagation analysis—is grounded in Cognitive Load Theory and instantiated here in the domain of financial valuation. Applying it to new domains requires formalizing domain-specific subtask hierarchies, but the measurement approach and analytical pipeline are designed to transfer.

Applying this framework to financial professionals using ChatGPT revealed how proactive AI is associated with both benefits and burdens. While AI content was associated with improved performance, proactive behaviors co-occurred with elevated extraneous load that diminished these gains through asymmetric dynamics: model responses tended to perpetuate user confusion rather than correct it, sustaining high-load states across turns. By decomposing these effects, we identified a hierarchy of harmful behaviors—task switching > information sprawl > truncation—providing actionable design targets.

This work contributes methodologically by extending Cognitive Load Theory into automated conversational trace analysis, theoretically by reframing proactivity as a state-dependent capability rather than generic enrichment, and practically by identifying design priorities for more precise proactivity: suppressing task switching, constraining sprawl, and treating high extraneous-load trajectories as detectable events requiring repair.

\printbibliography

@article{dell2023navigating,
  title={Navigating the jagged technological frontier: Field experimental evidence of the effects of AI on knowledge worker productivity and quality},
  author={Dell'Acqua, Fabrizio and McFowland III, Edward and Mollick, Ethan R and Lifshitz-Assaf, Hila and Kellogg, Katherine and Rajendran, Saran and Krayer, Lisa and Candelon, Fran{\c{c}}ois and Lakhani, Karim R},
  journal={Harvard Business School Technology \& Operations Mgt. Unit Working Paper},
  number={24-013},
  year={2023}
}

@article{mayer2001learning,
  author    = {Mayer, Richard E. and Chandler, Paul},
  title     = {When Learning is Just a Click Away: Does Simple User Interaction Foster Deeper Understanding of Multimedia Messages?},
  journal   = {Journal of Educational Psychology},
  volume    = {93},
  number    = {2},
  pages     = {390--397},
  year      = {2001},
  doi       = {10.1037/0022-0663.93.2.390}
}

@article{kalyuga2003expertise,
  author    = {Kalyuga, Slava and Ayres, Paul and Chandler, Paul and Sweller, John},
  title     = {The Expertise Reversal Effect},
  journal   = {Educational Psychologist},
  volume    = {38},
  number    = {1},
  pages     = {23--31},
  year      = {2003},
  doi       = {10.1207/S15326985EP3801_4}
}

@article{vanmerrienboer2005cognitive,
  author    = {van Merri{\"e}nboer, Jeroen J. G. and Sweller, John},
  title     = {Cognitive Load Theory and Complex Learning: Recent Developments and Future Directions},
  journal   = {Educational Psychology Review},
  volume    = {17},
  number    = {2},
  pages     = {147--177},
  year      = {2005},
  doi       = {10.1007/s10648-005-3951-0}
}

@inproceedings{horvitz1999principles,
  author    = {Horvitz, Eric},
  title     = {Principles of Mixed-Initiative User Interfaces},
  booktitle = {Proceedings of the SIGCHI Conference on Human Factors in Computing Systems},
  pages     = {159--166},
  year      = {1999},
  publisher = {ACM},
  doi       = {10.1145/302979.303030}
}

@book{pirolli2007information,
  author    = {Pirolli, Peter},
  title     = {Information Foraging Theory: Adaptive Interaction with Information},
  publisher = {Oxford University Press},
  year      = {2007},
  doi       = {10.1093/acprof:oso/9780195173321.001.0001}
}

@article{nielsen2006progressive,
  author    = {Nielsen, Jakob},
  title     = {Progressive Disclosure},
  journal   = {Nielsen Norman Group},
  year      = {2006},
  url       = {https://www.nngroup.com/articles/progressive-disclosure/}
}

@article{ranganathan2026ai,
  author    = {Ranganathan, Aruna and Ye, Xingqi Maggie},
  title     = {{AI} Doesn't Reduce Work---It Intensifies It},
  journal   = {Harvard Business Review},
  year      = {2026},
  month     = feb,
  url       = {https://hbr.org/2026/02/ai-doesnt-reduce-work-it-intensifies-it},
  note      = {Accessed: 2026-02-15}
}

@misc{storey2026cognitive,
  author    = {Storey, Margaret-Anne},
  title     = {How Generative and Agentic {AI} Shift Concern from Technical Debt to Cognitive Debt},
  howpublished = {Blog post},
  year      = {2026},
  month     = feb,
  url       = {https://margaretstorey.com/blog/2026/02/09/cognitive-debt/},
  note      = {Accessed: 2026-02-15}
}

@article{garvert2023hippocampal,
  title={Hippocampal spatio-predictive cognitive maps adaptively guide reward generalization},
  author={Garvert, Mona M and Saanum, Tankred and Schulz, Eric and Schuck, Nicolas W and Doeller, Christian F},
  journal={Nature Neuroscience},
  volume={26},
  number={4},
  pages={615--626},
  year={2023},
  publisher={Nature Publishing Group US New York}
}

@article{wickens2002multiple,
  title={Multiple resources and performance prediction},
  author={Wickens, Christopher D},
  journal={Theoretical issues in ergonomics science},
  volume={3},
  number={2},
  pages={159--177},
  year={2002},
  publisher={Taylor \& Francis}
}

@article{bokeria2021map,
  title={Map-like representations of an abstract conceptual space in the human brain},
  author={Bokeria, Levan and Henson, Richard N and Mok, Robert M},
  journal={Frontiers in Human Neuroscience},
  volume={15},
  pages={620056},
  year={2021},
  publisher={Frontiers Media SA}
}

@article{peer2021structuring,
  title={Structuring knowledge with cognitive maps and cognitive graphs},
  author={Peer, Michael and Brunec, Iva K and Newcombe, Nora S and Epstein, Russell A},
  journal={Trends in cognitive sciences},
  volume={25},
  number={1},
  pages={37--54},
  year={2021},
  publisher={Elsevier}
}

@article{lim2025proactivity,
  title={Proactivity of chatbots, task types and user’s characteristics when interacting with artificial intelligence (AI) chatbots},
  author={Lim, Jongmoon and Hwang, Wonil},
  journal={International Journal of Human--Computer Interaction},
  pages={1--19},
  year={2025},
  publisher={Taylor \& Francis}
}

@inproceedings{brenna2025toward,
  title={Toward Proactive Dialogic AI Agents},
  author={Brenna, Sofia and Jezek, Elisabetta and Magnini, Bernardo and others},
  booktitle={Joint Proceedings of the ACM IUI Workshops 2025},
  year={2025}
}

@article{hu2023enhancing,
  title={Enhancing large language model induced task-oriented dialogue systems through look-forward motivated goals},
  author={Hu, Zhiyuan and Feng, Yue and Deng, Yang and Li, Zekun and Ng, See-Kiong and Luu, Anh Tuan and Hooi, Bryan},
  journal={arXiv preprint arXiv:2309.08949},
  year={2023}
}

@article{kraus2020diy,
    author = {Kraus, Matthias and Schiller, Marvin and Behnke, Gregor and Bercher, Pascal and Dorna, Michael and Dambier, Michael and Glimm, Birte and Biundo, Susanne and Minker, Wolfgang},
    title = {"Was that successful?" On Integrating Proactive Meta-Dialogue in a DIY-Assistant using Multimodal Cues},
    journal = {Interncational Conference on Multimodal Interaction (ICMI '20')},
    year={2020}
}

@article{esmaeili2021current,
  title={A current view on dual-task paradigms and their limitations to capture cognitive load},
  author={Esmaeili Bijarsari, Shirin},
  journal={Frontiers in Psychology},
  volume={12},
  pages={648586},
  year={2021},
  publisher={Frontiers Media SA}
}

@book{taylor1911scientific,
    author = {Taylor, Frederick},
    title = {The Principles of Scientific Management},
    publisher = {Harper \& Brothers},
    year = {1911}
}

@inproceedings{kazemitabaar2024improving,
  title={Improving steering and verification in AI-assisted data analysis with interactive task decomposition},
  author={Kazemitabaar, Majeed and Williams, Jack and Drosos, Ian and Grossman, Tovi and Henley, Austin Zachary and Negreanu, Carina and Sarkar, Advait},
  booktitle={Proceedings of the 37th Annual ACM Symposium on User Interface Software and Technology},
  pages={1--19},
  year={2024}
}

@book{ayaz2018neuroergonomics,
  title={Neuroergonomics: The brain at work and in everyday life},
  author={Ayaz, Hasan and Dehais, Fr{\'e}d{\'e}ric},
  year={2018},
  publisher={Academic Press}
}

@article{wickens1976effects,
  title={The effects of divided attention on information processing in manual tracking.},
  author={Wickens, Christopher D},
  journal={Journal of Experimental Psychology: Human Perception and Performance},
  volume={2},
  number={1},
  pages={1},
  year={1976},
  publisher={American Psychological Association}
}

@article{hirshfield2024toward,
  title={Toward workload-based adaptive automation: The utility of fNIRS for measuring load in multiple resources in the brain},
  author={Hirshfield, Leanne M and Wickens, Christopher and Doherty, Emily and Spencer, Cara and Williams, Tom and Hayne, Lucas},
  journal={International Journal of Human--Computer Interaction},
  volume={40},
  number={22},
  pages={7404--7430},
  year={2024},
  publisher={Taylor \& Francis}
}

@inproceedings{wickens2024multiple,
  title={The multiple resource theory and model. Some misconceptions in data interpretations},
  author={Wickens, Christopher D},
  booktitle={Proceedings of the Human Factors and Ergonomics Society Annual Meeting},
  volume={68},
  number={1},
  pages={713--717},
  year={2024},
  organization={SAGE Publications Sage CA: Los Angeles, CA}
}

@article{norman1975data,
  title={On data-limited and resource-limited processes},
  author={Norman, Donald A and Bobrow, Daniel G},
  journal={Cognitive psychology},
  volume={7},
  number={1},
  pages={44--64},
  year={1975},
  publisher={Elsevier}
}

@article{wang2025tdag,
  title={Tdag: A multi-agent framework based on dynamic task decomposition and agent generation},
  author={Wang, Yaoxiang and Wu, Zhiyong and Yao, Junfeng and Su, Jinsong},
  journal={Neural Networks},
  volume={185},
  pages={107200},
  year={2025},
  publisher={Elsevier}
}

@article{khot2022decomposed,
  title={Decomposed prompting: A modular approach for solving complex tasks},
  author={Khot, Tushar and Trivedi, Harsh and Finlayson, Matthew and Fu, Yao and Richardson, Kyle and Clark, Peter and Sabharwal, Ashish},
  journal={arXiv preprint arXiv:2210.02406},
  year={2022}
}

@article{knisely2020cognitive,
  title={A cognitive decomposition to empirically study human performance in control room environments},
  author={Knisely, Benjamin M and Joyner, Janell S and Rutkowski, Anthony M and Wong, Matthew and Barksdale, Samuel and Hotham, Hayden and Kharod, Kush and Vaughn-Cooke, Monifa},
  journal={International Journal of Human-Computer Studies},
  volume={141},
  pages={102438},
  year={2020},
  publisher={Elsevier}
}

@article{coffey2013task,
  title={Task decomposition: a framework for comparing diverse training models in human brain plasticity studies},
  author={Coffey, Emily BJ and Herholz, Sibylle C},
  journal={Frontiers in human neuroscience},
  volume={7},
  pages={640},
  year={2013},
  publisher={Frontiers Media SA}
}

@article{brunken2002cognitive,
    author = {Brünken, Roland and Steinbacher, Susan and Plass, Jan L. and Leutner, Detlev},
    title = {Assessment of Cognitive Load in Multimedia Learning Using Dual-Task Methodology},
    journal = {Experimental Psychology},
    year = {2002} 
}

@article{berube2024proactive,
    author = {Bérube, Caterina and Nißen, Marcia and Vinay, Rasita and Geiger, Alexa and Budig, Tobias and Bhandari, Aashish and colleagues},
    title = {Proactive behavior in voice assistants: A systematic review and conceptual model},
    journal = {Computers in Human Behavior Reports},
    year = {2024} 
}

@article{babaei2025should,
  title={Should we use the NASA-TLX in HCI? A review of theoretical and methodological issues around Mental Workload Measurement},
  author={Babaei, Ebrahim and Dingler, Tilman and Tag, Benjamin and Velloso, Eduardo},
  journal={International Journal of Human-Computer Studies},
  pages={103515},
  year={2025},
  publisher={Elsevier}
}

@incollection{hart1988development,
  title={Development of NASA-TLX (Task Load Index): Results of empirical and theoretical research},
  author={Hart, Sandra G and Staveland, Lowell E},
  booktitle={Advances in psychology},
  volume={52},
  pages={139--183},
  year={1988},
  publisher={Elsevier}
}

@article{ahire2025designing,
    author = {Ahire, Shashank and Rohs, Michael},
    title = {Designing Proactive Voice Interfaces: Key Factors for Office Settings},
    journal = {Joint Proceedings of the ACM IUI Workshops 2025},
    year = {2025} 
}

@article{hazan2024influence,
  title={The Influence of Manipulating and Accentuating Task-Irrelevant Information on Learning Efficiency: Insights for Cognitive Load Theory},
  author={Hazan-Liran, Batel and Miller, Paul},
  journal={Journal of Cognition},
  volume={7},
  number={1},
  pages={36},
  year={2024}
}

@article{wilson2002truthfulness,
  title={Truthfulness and relevance},
  author={Wilson, Deirdre and Sperber, Dan},
  journal={Mind},
  volume={111},
  number={443},
  pages={583--632},
  year={2002},
  publisher={Mind Association}
}

@article{mikk2008sentence,
    author = {Mikk, Jaan},
    title = {Sentence length for revealing the cognitive load reversal effect in text comprehension},
    journal = {Educational Studies},
    year = {2008}
}

@article{kalyuga1999attention,
    author = {Kalyuga, Slava and Chandler, Paul and Swller, John},
    title = {Managing Split-Attention and Redundancy in Multimedia Instruction},
    journal = {Applied Cognitive Psychology},
    year = {1999}
}

@article{wiberg2021time,
    author = {Wiberg, Mikael and Stolterman, Erik},
    title = {Time and Temporality in HCI Research},
    journal = {Interacting with Computers},
    year = {2021}
}

@article{boehm-davis2009interruption,
    author = {Boehm-Davis, Deborah and Remington, Roger},
    title = {Reducing the disruptive effects of interruption: A cognitive framework for analysing the costs and benefits of intervention strategies},
    journal = {Accident Analysis \& Prevention},
    year = {2009}
}

@article{masotina2024relevance,
    author = {Masotina, Mariavittoria and Musi, Elena and Yates, Simeon},
    title = {Relevance theory for mapping cognitive biases in fact-checking; an argumentative approach},
    journal = {Frontiers in Pscyhology},
    year = {2024} 
}

@article{park2015cognitive,
    author = {Park, Babette and Flowerday, Terri and Brunken, Roland},
    title = {Cognitive and Affective Effects of Seductive Details in Multimedia Learning},
    journal = {Computers in Human Behavior},
    year = {2015}
}

@article{sperber1995relevance,
    author = {Sperber, Dan and Cara, Francesco and Girotto, Vittorio},
    title = {Relevance theory explains the selection task},
    journal = {Cognition},
    year = {1995}
}

@article{tomporowski2020cognitive,
    author = {Tomporowski, Phillip D and Qazi, Ahmed S.},
    title = {Cognitive-Motor Dual Task Interference Effects on Declarative Memory: A Theory-Based Review},
    journal = {Frontiers in Psychology},
    year = {2920}
}

@book{just198reading,
    author = {Just, M.A. and Carpenter, P.A.},
    title = {The psychology of reading and language comprehension},
    publisher = {Allyn \& Bacon},
    year = {1987}
}

@article{rayner1998eye,
    author = {Rayner, Keith},
    title = {Eye movements in reading and information processing: 20 years of research},
    journal = {Psychological Bulletin},
    year = {1998}
}

@article{lespiau2024reasoning,
    author = {Lespiau, Florence and Tricot, André},
    title = {Reasoning More Efficiently with Primary Knowledge Despite Extraneous Cognitive Load},
    journal = {Evolutionary Psychology},
    year = {2024}
}

@article{pu2025assistance,
    author = {Pu, Kevin and Lazaro, Daniel and Arawjo, Ian and Xia, Haijun and Xiao, Zhang and Grossman, Tovi},
    title = {Assistance or Disruption? Exploring and Evaluating the Design and Trade-offs of Proactive AI Programming Support},
    journal = {CHI Conference on Human Factors in Computing Systems (CHI`25)},
    year = {2025}
}

@article{klepsch2020understanding,
    author = {Klepsch, Melina and Seufert, Tina},
    title = {Understanding the instructional design effects by differentiated measures of intrinsic, extraneous, and germane cognitive load},
    journal = {Instructional Science},
    year = {2020}
}

@article{kosch2023cognitive,
author = {Kosch, Thomas and Karolus, Jakob and Zagermann, Johannes and Reiterer, Harald and Schmidt, Albrecht and Woźniak, Paweł},
year = {2023},
month = {01},
pages = {},
title = {A Survey on Measuring Cognitive Workload in Human-Computer Interaction},
volume = {55},
journal = {ACM Computing Surveys},
doi = {10.1145/3582272}
}

@article{kienitz2023seductive,
    author = {Kienitz, Anna and Krebs, Marie-Christin and Eittel, Alexander},
    title = {Seductive details hamper learning even when they do not disrupt},
    journal = {Instruction Science},
    year = {2023}
}

@article{sigurd2004word,
    author = {Sigurd, Bengt and Eeg-Olofsson, Mats and van de Weijer, Joost},
    title = {Word Lenght, Sentence Length and Frequency--ZIPF Revisited},
    journal = {Studia Linguistics},
    year = {2004}
}

@article{luna2018cognitive,
    author = {Luna, Karlos and Albuguergue, Pedro B. and Martín-Luengo, Beatriz},
    title = {Cognitive load eliminates the effects of perceptual information on judgements of learning with sentences},
    journal = {Memory \& Cognitive},
    year = {2018}
}

@article{balaraman2021dialogue,
    author = {Balaraman, Vevake and Sheikhalishahi, Seyedmostafa and Magnini, Bernardo},
    title = {Recent Neural Methods on Dialogue State Tracking for Task-Oriented Dialogue Systems: A Survey},
    journal = {Proceedings of the 22nd Annual Meeting of the Special Interest Group on Discourse and Dialogue},
    year = {2021}
}

@article{xu2024chain,
    author = {Xu, Lin and Peng, Ningxin and Zhou, Daquan and Ng, See-Kiong and Fu, Jinlan},
    title = {Chain of Thought Explanation for Dialogue State Tracking},
    journal = {arXiv preprint arXiv::2403.04656},
    year = {2024}
}

@article{wickens2008multiple,
  title={Multiple resources and mental workload},
  author={Wickens, Christopher D},
  journal={Human factors},
  volume={50},
  number={3},
  pages={449--455},
  year={2008},
  publisher={SAGE Publications Sage CA: Los Angeles, CA}
}

@article{ding2025enhancing,
  title={Enhancing domain-specific knowledge graph reasoning via metapath-based large model prompt learning},
  author={Ding, Ruidong and Zhou, Bin},
  journal={Electronics},
  volume={14},
  number={5},
  pages={1012},
  year={2025},
  publisher={MDPI}
}

@inproceedings{altmann2004task,
  title={Task interruption: Resumption lag and the role of cues},
  author={Altmann, Erik M and Trafton, J Gregory},
  booktitle={Proceedings of the Annual Meeting of the Cognitive Science Society},
  volume={26},
  number={26},
  year={2004}
}

@article{abu2021domain,
  title={Domain-specific knowledge graphs: A survey},
  author={Abu-Salih, Bilal},
  journal={Journal of Network and Computer Applications},
  volume={185},
  pages={103076},
  year={2021},
  publisher={Elsevier}
}

@article{dziri2019evaluating,
  title={Evaluating coherence in dialogue systems using entailment},
  author={Dziri, Nouha and Kamalloo, Ehsan and Mathewson, Kory W and Zaiane, Osmar},
  journal={arXiv preprint arXiv:1904.03371},
  year={2019}
}

@article{ye2021towards,
  title={Towards quantifiable dialogue coherence evaluation},
  author={Ye, Zheng and Lu, Liucun and Huang, Lishan and Lin, Liang and Liang, Xiaodan},
  journal={arXiv preprint arXiv:2106.00507},
  year={2021}
}

@article{teutloff2025winners,
    author = {Teutloff, Ole and Einsiedler, Johanna and Kassi, Otto and Braesemann, Fabian and Mishkin, Pamela and del Rio-Chanona, R. Maria},
    title = {Winners and losers of generative AI: Early Evidence of Shifts in Freelancer Demand},
    journal = {Journal of Economic Behavior \& Organization} ,
    year = {2025}
}

@book{marshall1995schemas,
    author = {Marshall, Sandra P.},
    title = {Schemas in Problem Solving},
    publisher = {Cambridge University Press},
    year = {1995}
}

@article{vasconcelos2023explanations,
  title={Explanations can reduce overreliance on ai systems during decision-making},
  author={Vasconcelos, Helena and J{\"o}rke, Matthew and Grunde-McLaughlin, Madeleine and Gerstenberg, Tobias and Bernstein, Michael S and Krishna, Ranjay},
  journal={Proceedings of the ACM on Human-Computer Interaction},
  volume={7},
  number={CSCW1},
  pages={1--38},
  year={2023},
  publisher={ACM New York, NY, USA}
}

@article{kraus2021trust,
    author = {Kraus, Matthias and Wagner, Nicolas and Callejas, Zoraida and Minker, Wolfgang},
    title = {The Role of Trust in Proactive Conversational Assistants},
    journal = {IEEE Access},
    year = {2021}
}

@article{feng2023dialogue,
    author = {Feng, Yujie and Lu, Zexin and Liu, Bo and Zhan, Liming and Wu, Xiao-Ming},
    title = {Towards LLM-driven Dialogue State Tracking},
    journal = {Proceedings of the 2023 Conference on Empirical Methods in Natural Language Processing},
    year = {2023}
}

@inproceedings{kraus2020effects,
  title={Effects of proactive dialogue strategies on human-computer trust},
  author={Kraus, Matthias and Wagner, Nicolas and Minker, Wolfgang},
  booktitle={Proceedings of the 28th ACM conference on user modeling, adaptation and personalization},
  pages={107--116},
  year={2020}
}

@article{buccinca2021trust,
  title={To trust or to think: cognitive forcing functions can reduce overreliance on AI in AI-assisted decision-making},
  author={Bu{\c{c}}inca, Zana and Malaya, Maja Barbara and Gajos, Krzysztof Z},
  journal={Proceedings of the ACM on Human-computer Interaction},
  volume={5},
  number={CSCW1},
  pages={1--21},
  year={2021},
  publisher={ACM New York, NY, USA}
}

@inproceedings{kraus2023improving,
  title={Improving proactive dialog agents using socially-aware reinforcement learning},
  author={Kraus, Matthias and Wagner, Nicolas and Riekenbrauck, Ron and Minker, Wolfgang},
  booktitle={Proceedings of the 31st ACM Conference on User Modeling, Adaptation and Personalization},
  pages={146--155},
  year={2023}
}

@article{wylie2000task,
  title={Task switching and the measurement of “switch costs”},
  author={Wylie, Glenn and Allport, Alan},
  journal={Psychological research},
  volume={63},
  number={3},
  pages={212--233},
  year={2000},
  publisher={Springer}
}

@article{eloundou2023gpts,
  title={Gpts are gpts: An early look at the labor market impact potential of large language models},
  author={Eloundou, Tyna and Manning, Sam and Mishkin, Pamela and Rock, Daniel},
  journal={arXiv preprint arXiv:2303.10130},
  year={2023}
}

@article{correa2023humans,
  title={Humans decompose tasks by trading off utility and computational cost},
  author={Correa, Carlos G and Ho, Mark K and Callaway, Frederick and Daw, Nathaniel D and Griffiths, Thomas L},
  journal={PLoS computational biology},
  volume={19},
  number={6},
  pages={e1011087},
  year={2023},
  publisher={Public Library of Science San Francisco, CA USA}
}

@book{anderson_1983,
    author = {Anderson, J.R.,},
    title = {The Architecture of Cognition},
    publisher = {Harvard University Press},
    year = {1983}
}

@inproceedings{iqbal2007disruption,
  title={Disruption and recovery of computing tasks: field study, analysis, and directions},
  author={Iqbal, Shamsi T and Horvitz, Eric},
  booktitle={Proceedings of the SIGCHI conference on Human factors in computing systems},
  pages={677--686},
  year={2007}
}

@article{ullman2012structural,
  title={Structural equation modeling},
  author={Ullman, Jodie B and Bentler, Peter M},
  journal={Handbook of psychology, second edition},
  volume={2},
  year={2012},
  publisher={Wiley Online Library}
}

@incollection{card2009information,
  title={Information visualization},
  author={Card, Stuart},
  booktitle={Human-computer interaction},
  pages={199--234},
  year={2009},
  publisher={CRC press}
}

@inproceedings{joachims2002optimizing,
  title={Optimizing search engines using clickthrough data},
  author={Joachims, Thorsten},
  booktitle={Proceedings of the eighth ACM SIGKDD international conference on Knowledge discovery and data mining},
  pages={133--142},
  year={2002}
}

@inproceedings{hu2008collaborative,
  title={Collaborative filtering for implicit feedback datasets},
  author={Hu, Yifan and Koren, Yehuda and Volinsky, Chris},
  booktitle={2008 Eighth IEEE international conference on data mining},
  pages={263--272},
  year={2008},
  organization={Ieee}
}

@inproceedings{park2021scalable,
  title={A scalable framework for learning from implicit user feedback to improve natural language understanding in large-scale conversational AI systems},
  author={Park, Sunghyun and Li, Han and Patel, Ameen and Mudgal, Sidharth and Lee, Sungjin and Kim, Young-Bum and Matsoukas, Spyros and Sarikaya, Ruhi},
  booktitle={Proceedings of the 2021 Conference on Empirical Methods in Natural Language Processing},
  pages={6054--6063},
  year={2021}
}

@article{hegde2025factors,
  title={Factors affecting the in-context learning abilities of LLMs for dialogue state tracking},
  author={Hegde, Pradyoth and Kesiraju, Santosh and {\v{S}}vec, J{\'a}n and Sedl{\'a}{\v{c}}ek, {\v{S}}imon and Yusuf, Bolaji and Plchot, Old{\v{r}}ich and {\v{C}}ernock{\`y}, Jan and others},
  journal={arXiv preprint arXiv:2506.08753},
  year={2025}
}

@article{corbett1994knowledge,
  title={Knowledge tracing: Modeling the acquisition of procedural knowledge},
  author={Corbett, Albert T and Anderson, John R},
  journal={User modeling and user-adapted interaction},
  volume={4},
  number={4},
  pages={253--278},
  year={1994},
  publisher={Springer}
}

@article{pirolli1999information,
  title={Information foraging.},
  author={Pirolli, Peter and Card, Stuart},
  journal={Psychological review},
  volume={106},
  number={4},
  pages={643},
  year={1999},
  publisher={American Psychological Association}
}

@inproceedings{zargham2022understanding,
  title={Understanding circumstances for desirable proactive behaviour of voice assistants: The proactivity dilemma},
  author={Zargham, Nima and Reicherts, Leon and Bonfert, Michael and Voelkel, Sarah Theres and Schoening, Johannes and Malaka, Rainer and Rogers, Yvonne},
  booktitle={Proceedings of the 4th conference on conversational user interfaces},
  pages={1--14},
  year={2022}
}

@inproceedings{rahmani2024clarifying,
  title={Clarifying the path to user satisfaction: An investigation into clarification usefulness},
  author={Rahmani, Hossein A and Wang, Xi and Aliannejadi, Mohammad and Naghiaei, Mohammadmehdi and Yilmaz, Emine},
  booktitle={Findings of the Association for Computational Linguistics: EACL 2024},
  pages={1266--1277},
  year={2024}
}

@article{kosmyna2025your,
  title={Your brain on ChatGPT: Accumulation of cognitive debt when using an AI assistant for essay writing task},
  author={Kosmyna, Nataliya and Hauptmann, Eugene and Yuan, Ye Tong and Situ, Jessica and Liao, Xian-Hao and Beresnitzky, Ashly Vivian and Braunstein, Iris and Maes, Pattie},
  journal={arXiv preprint arXiv:2506.08872},
  volume={4},
  year={2025}
}

@article{satterthwaite1946approximate,
  title={An approximate distribution of estimates of variance components},
  author={Satterthwaite, Franklin E},
  journal={Biometrics bulletin},
  volume={2},
  number={6},
  pages={110--114},
  year={1946},
  publisher={JSTOR}
}

@article{bingham2023data,
  title={From data management to actionable findings: A five-phase process of qualitative data analysis},
  author={Bingham, Andrea J},
  journal={International journal of qualitative methods},
  volume={22},
  pages={16094069231183620},
  year={2023},
  publisher={SAGE Publications Sage CA: Los Angeles, CA}
}

@article{meteyard2020best,
  title={Best practice guidance for linear mixed-effects models in psychological science},
  author={Meteyard, Lotte and Davies, Robert AI},
  journal={Journal of Memory and Language},
  volume={112},
  pages={104092},
  year={2020},
  publisher={Elsevier}
}

@techreport{anthroindex2026,
    author = {Appel, Ruth and Massenkoff, Maxim and McCrory, Peter and McCain, Miles and Heller, Ryan and Neylon, Tyler and Tamkin, Alex},
    title = {The Anthropic Economic Index report: Economic Primitives},
    institution = {Anthropic},
    year = {2026}
}

@inproceedings{mark2008cost,
  title={The cost of interrupted work: more speed and stress},
  author={Mark, Gloria and Gudith, Daniela and Klocke, Ulrich},
  booktitle={Proceedings of the SIGCHI conference on Human Factors in Computing Systems},
  pages={107--110},
  year={2008}
}

@article{roda2006attention,
  title={Attention aware systems: Theories, applications, and research agenda},
  author={Roda, Claudia and Thomas, Julie},
  journal={Computers in Human Behavior},
  volume={22},
  number={4},
  pages={557--587},
  year={2006},
  publisher={Elsevier}
}

@inbook{krabbe2017validity,
    author = {Krabbe, Paul F.M.},
    title = {The Measurement of Health and Health Status} ,
    publisher = {Academic Press},
    year = {2017},
    chapter = {Chapter 7 - Validity}
}

@techreport{kincaid1975derivation,
  title={Derivation of new readability formulas (automated readability index, fog count and flesch reading ease formula) for navy enlisted personnel},
  author={Kincaid, J Peter and Fishburne Jr, Robert P and Rogers, Richard L and Chissom, Brad S},
  institution={Naval Technical Training Command},
  year={1975}
}

@article{flesch1948readability,
    author = {Flesch, R},
    title = {A new readability yardstick},
    journal = {Journal of Applied Psychology},
    year = {1948}
}

@techreport{ohayre1966gobb,
    author = {O'Hayre, John},
    title = {Gobbledygook Has Gotta Go},
    institution = {U.S. Fish and Wildlife Service},
    year = {1966}
}

@book{gunning1952fog,
    author = {Gunning, Robert},
    title = {The Technique of Clear Writing},
    publisher = {McGraw-Hill},
    year = {1952}
}

@article{dalechall1948readability,
    author = {Dale, E and Chall, J.},
    title = {A Formula for Predicting Readability},
    journal = {Educational Research Bulletin},
    year = {1948}
}

@article{mc1969smog,
  title={SMOG grading-a new readability formula},
  author={Mc Laughlin, G Harry},
  journal={Journal of reading},
  volume={12},
  number={8},
  pages={639--646},
  year={1969},
  publisher={JSTOR}
}

@inproceedings{xiao2025streaming,
  title={Streaming, fast and slow: Cognitive load-aware streaming for efficient llm serving},
  author={Xiao, Chang and Yang, Zixiaofan},
  booktitle={Proceedings of the 38th Annual ACM Symposium on User Interface Software and Technology},
  pages={1--13},
  year={2025}
}

@book{clark1996using,
  title={Using language},
  author={Clark, Herbert H},
  year={1996},
  publisher={Cambridge university press}
}

@article{coleman1975readability,
    author = {Coleman, Meri and Liau, T.L.},
    title = {A computer readability formula designed for machine scoring },
    journal = {Journal of Applied Psychology},
    year = {1975}
}

@techreport{senter1967automated,
  title={Automated readability index},
  author={Senter, Richard J and Smith, Edgar A},
  year={1967}, 
  institution={Aerospace Medical Research Laboratories}
}

@article{pickering2004toward,
  title={Toward a mechanistic psychology of dialogue},
  author={Pickering, Martin J and Garrod, Simon},
  journal={Behavioral and brain sciences},
  volume={27},
  number={2},
  pages={169--190},
  year={2004},
  publisher={Cambridge University Press}
}

@article{sweller1998cognitive,
  title={Cognitive architecture and instructional design},
  author={Sweller, John and Van Merrienboer, Jeroen JG and Paas, Fred GWC},
  journal={Educational psychology review},
  volume={10},
  pages={251--296},
  year={1998},
  publisher={Springer}
}

@article{sweller1994some,
  title={Why some material is difficult to learn},
  author={Sweller, John and Chandler, Paul},
  journal={Cognition and instruction},
  volume={12},
  number={3},
  pages={185--233},
  year={1994},
  publisher={Taylor \& Francis}
}

@article{sweller2010element,
  title={Element interactivity and intrinsic, extraneous, and germane cognitive load},
  author={Sweller, John},
  journal={Educational psychology review},
  volume={22},
  pages={123--138},
  year={2010},
  publisher={Springer}
}

@article{sweller1994cognitive,
  title={Cognitive load theory, learning difficulty, and instructional design},
  author={Sweller, John},
  journal={Learning and instruction},
  volume={4},
  number={4},
  pages={295--312},
  year={1994},
  publisher={Elsevier}
}

@article{oberauer2019working,
  title={Working memory and attention--A conceptual analysis and review},
  author={Oberauer, Klaus},
  journal={Journal of cognition},
  volume={2},
  number={1},
  pages={36},
  year={2019}
}

@article{cowan2012models,
  title={Models of verbal working memory capacity: What does it take to make them work?},
  author={Cowan, Nelson and Rouder, Jeffrey N and Blume, Christopher L and Saults, J Scott},
  journal={Psychological review},
  volume={119},
  number={3},
  pages={480},
  year={2012},
  publisher={American Psychological Association}
}

@article{wiles2024genai,
  title={Genai as an exoskeleton: Experimental evidence on knowledge workers using genai on new skills},
  author={Wiles, Emma and Krayer, Lisa and Abbadi, Mohamed and Awasthi, Urvi and Kennedy, Ryan and Mishkin, Pamela and Sack, Daniel and Candelon, Fran{\c{c}}ois},
  journal={Available at SSRN 4944588},
  year={2024}
}

@article{huang2020grade,
  title={GRADE: Automatic graph-enhanced coherence metric for evaluating open-domain dialogue systems},
  author={Huang, Lishan and Ye, Zheng and Qin, Jinghui and Lin, Liang and Liang, Xiaodan},
  journal={arXiv preprint arXiv:2010.03994},
  year={2020}
}

@article{zhang2021dynaeval,
  title={DynaEval: Unifying turn and dialogue level evaluation},
  author={Zhang, Chen and Chen, Yiming and D'Haro, Luis Fernando and Zhang, Yan and Friedrichs, Thomas and Lee, Grandee and Li, Haizhou},
  journal={arXiv preprint arXiv:2106.01112},
  year={2021}
}

@article{brown2020language,
  title={Language models are few-shot learners},
  author={Brown, Tom and Mann, Benjamin and Ryder, Nick and Subbiah, Melanie and Kaplan, Jared D and Dhariwal, Prafulla and Neelakantan, Arvind and Shyam, Pranav and Sastry, Girish and Askell, Amanda and others},
  journal={Advances in neural information processing systems},
  volume={33},
  pages={1877--1901},
  year={2020}
}

@techreport{brynjolfsson2023generative,
  title={Generative AI at work},
  author={Brynjolfsson, Erik and Li, Danielle and Raymond, Lindsey R},
  year={2023},
  institution={National Bureau of Economic Research}
}

@article{deng2023survey,
  title={A survey on proactive dialogue systems: Problems, methods, and prospects},
  author={Deng, Yang and Lei, Wenqiang and Lam, Wai and Chua, Tat-Seng},
  journal={arXiv preprint arXiv:2305.02750},
  year={2023}
}

@article{hosseini2020simple,
  title={A simple language model for task-oriented dialogue},
  author={Hosseini-Asl, Ehsan and McCann, Bryan and Wu, Chien-Sheng and Yavuz, Semih and Socher, Richard},
  journal={Advances in Neural Information Processing Systems},
  volume={33},
  pages={20179--20191},
  year={2020}
}

@inproceedings{dong2025protod,
  title={ProTOD: Proactive Task-oriented Dialogue System Based on Large Language Model},
  author={Dong, Wenjie and Chen, Sirong and Yang, Yan},
  booktitle={Proceedings of the 31st International Conference on Computational Linguistics},
  pages={9147--9164},
  year={2025}
}

@article{rogers1995costs,
  title={Costs of a predictible switch between simple cognitive tasks.},
  author={Rogers, Robert D and Monsell, Stephen},
  journal={Journal of experimental psychology: General},
  volume={124},
  number={2},
  pages={207},
  year={1995},
  publisher={American Psychological Association}
}

@article{cowan2001magical,
  title={The magical number 4 in short-term memory: A reconsideration of mental storage capacity},
  author={Cowan, Nelson},
  journal={Behavioral and brain sciences},
  volume={24},
  number={1},
  pages={87--114},
  year={2001},
  publisher={Cambridge University Press}
}

@article{chandler1992split,
  title={The split-attention effect as a factor in the design of instruction},
  author={Chandler, Paul and Sweller, John},
  journal={British Journal of Educational Psychology},
  volume={62},
  number={2},
  pages={233--246},
  year={1992},
  publisher={Wiley Online Library}
}

@inproceedings{deng2023rethinking,
  title={Rethinking Conversational Agents in the Era of LLMs: Proactivity, Non-collaborativity, and Beyond},
  author={Deng, Yang and Lei, Wenqiang and Huang, Minlie and Chua, Tat-Seng},
  booktitle={Proceedings of the Annual International ACM SIGIR Conference on Research and Development in Information Retrieval in the Asia Pacific Region},
  pages={298--301},
  year={2023}
}

@article{desmond2024exploring,
  title={Exploring prompt engineering practices in the enterprise},
  author={Desmond, Michael and Brachman, Michelle},
  journal={arXiv preprint arXiv:2403.08950},
  year={2024}
}

@inproceedings{zamfirescu2023johnny,
  title={Why Johnny can’t prompt: how non-AI experts try (and fail) to design LLM prompts},
  author={Zamfirescu-Pereira, J Diego and Wong, Richmond Y and Hartmann, Bjoern and Yang, Qian},
  booktitle={Proceedings of the 2023 CHI conference on human factors in computing systems},
  pages={1--21},
  year={2023}
}

@article{mcvee2005schema,
    author = {McVee, Mary B. and Dunsmore, Kailonnie and Gavelek, James R.},
    title = {Schema Theory Revisited},
    journal = {Review of Educational Research} ,
    year = {2005}
}

@article{herlihy2024overcoming,
  title={On overcoming miscalibrated conversational priors in llm-based chatbots},
  author={Herlihy, Christine and Neville, Jennifer and Schnabel, Tobias and Swaminathan, Adith},
  journal={arXiv preprint arXiv:2406.01633},
  year={2024}
}

@article{kopf2023openassistant,
  title={Openassistant conversations-democratizing large language model alignment},
  author={K{\"o}pf, Andreas and Kilcher, Yannic and Von R{\"u}tte, Dimitri and Anagnostidis, Sotiris and Tam, Zhi Rui and Stevens, Keith and Barhoum, Abdullah and Nguyen, Duc and Stanley, Oliver and Nagyfi, Rich{\'a}rd and others},
  journal={Advances in neural information processing systems},
  volume={36},
  pages={47669--47681},
  year={2023}
}

@article{christiano2017deep,
  title={Deep reinforcement learning from human preferences},
  author={Christiano, Paul F and Leike, Jan and Brown, Tom and Martic, Miljan and Legg, Shane and Amodei, Dario},
  journal={Advances in neural information processing systems},
  volume={30},
  year={2017}
}

@article{stiennon2020learning,
  title={Learning to summarize with human feedback},
  author={Stiennon, Nisan and Ouyang, Long and Wu, Jeffrey and Ziegler, Daniel and Lowe, Ryan and Voss, Chelsea and Radford, Alec and Amodei, Dario and Christiano, Paul F},
  journal={Advances in neural information processing systems},
  volume={33},
  pages={3008--3021},
  year={2020}
}

@article{ouyang2022training,
  title={Training language models to follow instructions with human feedback},
  author={Ouyang, Long and Wu, Jeffrey and Jiang, Xu and Almeida, Diogo and Wainwright, Carroll and Mishkin, Pamela and Zhang, Chong and Agarwal, Sandhini and Slama, Katarina and Ray, Alex and others},
  journal={Advances in neural information processing systems},
  volume={35},
  pages={27730--27744},
  year={2022}
}

@article{jaeger2010redundancy,
  title={Redundancy and reduction: Speakers manage syntactic information density},
  author={Jaeger, T Florian},
  journal={Cognitive psychology},
  volume={61},
  number={1},
  pages={23--62},
  year={2010},
  publisher={Elsevier}
}

@article{clark1991grounding,
  title={Grounding in communication.},
  author={Clark, Herbert H and Brennan, Susan E},
  year={1991},
  journal={Perspectives on Socially Shared Cognition},
  publisher={American Psychological Association}
}

@article{piantadosi2012communicative,
  title={The communicative function of ambiguity in language},
  author={Piantadosi, Steven T and Tily, Harry and Gibson, Edward},
  journal={Cognition},
  volume={122},
  number={3},
  pages={280--291},
  year={2012},
  publisher={Elsevier}
}

@article{sarkar2025conversational,
  title={Conversational user-ai intervention: A study on prompt rewriting for improved llm response generation},
  author={Sarkar, Rupak and Sarrafzadeh, Bahareh and Chandrasekaran, Nirupama and Rangan, Nagu and Resnik, Philip and Yang, Longqi and Jauhar, Sujay Kumar},
  journal={arXiv preprint arXiv:2503.16789},
  year={2025}
}

@article{prasad2024towards,
  title={Towards adoption of generative AI in organizational settings},
  author={Prasad Agrawal, Kalyan},
  journal={Journal of Computer Information Systems},
  volume={64},
  number={5},
  pages={636--651},
  year={2024},
  publisher={Taylor \& Francis}
}

@article{bailey2006attention,
    author = {Bailey, Brian P. and Konstan, Joseph A.},
    title = {On the need for attention-aware systems: measuring effects of interruption on task performance, error-rate, and affective state},
    journal = {Computers in Human Behavior},
    year = {2006}
}

@article{edmunds2000information,
    author = {Edmunds, Angela and Morris, Anne},
    title = {The problem of information overload in business organizations: a review of the literature},
    journal = {International Journal of Information Management},
    year = {2000}
}

@article{altmann2002memory,
    author = {Altmann, Erik M and Trafton, J. Gregory},
    title = {Memory for goals: an activation-based model},
    journal = {Cognitive Science},
    year = {2002}
}

@article{brazzolotto2022interruptions,
    author = {Brazzolotto, P and Duran, G and Michael, G.A.},
    title = {How do we handle interruptions? Investigating the processes underlying the resumption of interrupted tasks},
    journal = {Psychologie Francaise},
    year = {2022}
}

@article{kiesel2010control,
  title={Control and interference in task switching—A review.},
  author={Kiesel, Andrea and Steinhauser, Marco and Wendt, Mike and Falkenstein, Michael and Jost, Kerstin and Philipp, Andrea M and Koch, Iring},
  journal={Psychological bulletin},
  volume={136},
  number={5},
  pages={849},
  year={2010},
  publisher={American Psychological Association}
}

@inproceedings{kuang2024enhancing,
  title={Enhancing UX evaluation through collaboration with conversational AI assistants: Effects of proactive dialogue and timing},
  author={Kuang, Emily and Li, Minghao and Fan, Mingming and Shinohara, Kristen},
  booktitle={Proceedings of the 2024 CHI Conference on Human Factors in Computing Systems},
  pages={1--16},
  year={2024}
}

@article{wang2023dialogue,
  title={Dialogue planning via brownian bridge stochastic process for goal-directed proactive dialogue},
  author={Wang, Jian and Lin, Dongding and Li, Wenjie},
  journal={arXiv preprint arXiv:2305.05290},
  year={2023}
}

@article{xiao2020tell,
  title={Tell me about yourself: Using an AI-powered chatbot to conduct conversational surveys with open-ended questions},
  author={Xiao, Ziang and Zhou, Michelle X and Liao, Q Vera and Mark, Gloria and Chi, Changyan and Chen, Wenxi and Yang, Huahai},
  journal={ACM Transactions on Computer-Human Interaction (TOCHI)},
  volume={27},
  number={3},
  pages={1--37},
  year={2020},
  publisher={ACM New York, NY, USA}
}

@article{altmann2007timecourse,
  title={Timecourse of recovery from task interruption: Data and a model},
  author={Altmann, Erik M and Trafton, J Gregory},
  journal={Psychonomic bulletin \& review},
  volume={14},
  number={6},
  pages={1079--1084},
  year={2007},
  publisher={Springer}
}

@article{altmann2014momentary,
  title={Momentary interruptions can derail the train of thought.},
  author={Altmann, Erik M and Trafton, J Gregory and Hambrick, David Z},
  journal={Journal of Experimental Psychology: General},
  volume={143},
  number={1},
  pages={215},
  year={2014},
  publisher={American Psychological Association}
}

@article{trafton2007task,
  title={Task interruptions},
  author={Trafton, J Gregory and Monk, Christopher A},
  journal={Reviews of human factors and ergonomics},
  volume={3},
  number={1},
  pages={111--126},
  year={2007},
  publisher={SAGE Publications Sage CA: Los Angeles, CA}
}

@article{monsell2003task,
  title={Task switching},
  author={Monsell, Stephen},
  journal={Trends in cognitive sciences},
  volume={7},
  number={3},
  pages={134--140},
  year={2003},
  publisher={Elsevier}
}

@article{laban2025llms,
  title={Llms get lost in multi-turn conversation},
  author={Laban, Philippe and Hayashi, Hiroaki and Zhou, Yingbo and Neville, Jennifer},
  journal={arXiv preprint arXiv:2505.06120},
  year={2025}
}

@article{gibson2024AI,
    author = {Gibson, Kate},
    title = {Benefits of AI in Business},
    journal = {Harvard Business School Online} ,
    year = {2024} 
}

@inproceedings{balaraman2020proactive,
  title={Proactive systems and influenceable users: Simulating proactivity in task-oriented dialogues},
  author={Balaraman, Vevake and Magnini, Bernardo},
  booktitle={Proceedings of the 24th Workshop on the Semantics and Pragmatics of Dialogue-Full Papers, Virually at Brandeis, Waltham, New Jersey, July. SEMDIAL},
  year={2020}
}

@inproceedings{speer2017conceptnet,
  title={Conceptnet 5.5: An open multilingual graph of general knowledge},
  author={Speer, Robyn and Chin, Joshua and Havasi, Catherine},
  booktitle={Proceedings of the AAAI conference on artificial intelligence},
  volume={31},
  number={1},
  year={2017}
}

@article{kalyuga2000learner,
    author = {Kalyuga, Slava and Chandler, Paul and Sweller, John},
    title = {Incorporating learner experience into the design of multimedia instruction},
    journal = {Journal of Educational Psychology},
    year = {2000}
}

@article{trypke2023redundancy,
    author = {Trypke, Melanie and Stebner, Ferdinand and Wirth, Joachim},
    title = {Two types of redundancy in multimedia learning: a literature review},
    journal = {Frontiers in Psychology},
    year = {2023} 
}

@inproceedings{lee2025impact,
  title={The Impact of Generative AI on Critical Thinking: Self-Reported Reductions in Cognitive Effort and Confidence Effects From a Survey of Knowledge Workers},
  author={Lee, Hao-Ping and Sarkar, Advait and Tankelevitch, Lev and Drosos, Ian and Rintel, Sean and Banks, Richard and Wilson, Nicholas},
  booktitle={Proceedings of the 2025 CHI Conference on Human Factors in Computing Systems},
  year={2025}
}

@book{damodaran2012investment,
  title={Investment Valuation: Tools and Techniques for Determining the Value of Any Asset},
  author={Damodaran, Aswath},
  year={2012},
  edition={3rd},
  publisher={John Wiley \& Sons},
  address={Hoboken, NJ}
}


\newpage

\appendix
\renewcommand{\thesection}{\Alph{section}}
\renewcommand{\thesubsection}{\Alph{section}.\arabic{subsection}}

\renewcommand{\thetable}{\Alph{section}\arabic{table}}
\renewcommand{\thefigure}{\Alph{section}\arabic{figure}}

\setcounter{table}{0}
\setcounter{figure}{0}

\section{Data and Sample Construction}\label{apsec:data-and-sample-construction}

\subsection{Descriptive Statistics}\label{app:descriptives}

\begin{table}[!htbp]
\centering
\scriptsize
\setlength{\tabcolsep}{4pt}
\begin{threeparttable}
\caption{Descriptive Statistics Across Analytic Samples (Pre-Standardization)}
\label{tab:descriptives_all}
\begin{tabular}{lrrrrrrrr}
\toprule
\textbf{Variable} & \textbf{N} & \textbf{Mean} & \textbf{SD} & \textbf{Min} & \textbf{25\%} & \textbf{50\%} & \textbf{75\%} & \textbf{Max} \\
\midrule

\multicolumn{9}{l}{\textbf{Panel A: Participant--Subtask Analytic Sample (N = 1344)}} \\
\midrule
Output Quality (DV) & 1344 & 2.662 & 1.742 & 0.000 & 1.000 & 3.000 & 4.000 & 5.000 \\
AI-Generated Content Usage (AIGCU) & 1344 & 0.084 & 1.413 & -1.217 & -0.911 & -0.459 & 0.548 & 4.868 \\
Professional Experience (Years) & 1344 & 9.002 & 6.456 & 2.000 & 4.000 & 8.000 & 12.000 & 30.000 \\
Prior Organizational Seniority$^{\dagger}$ & 1344 & 4.307 & 2.839 & 1.000 & 2.000 & 4.000 & 6.000 & 9.000 \\

\midrule
\multicolumn{9}{l}{\textbf{Panel B: PSC-Based Participant--Subtask Sample (N = 1178)}} \\
\midrule
Output Quality (DV) & 1178 & 2.724 & 1.741 & 0.000 & 1.000 & 3.000 & 4.000 & 5.000 \\
AI-Generated Content Usage (AIGCU) & 1178 & 0.115 & 1.426 & -1.217 & -0.887 & -0.411 & 0.550 & 4.868 \\
Intrinsic Cognitive Load (ICL) & 1178 & 11.852 & 0.762 & 7.222 & 11.499 & 11.852 & 12.402 & 13.622 \\
Bidirectional PSC (raw coherence)$^{*}$ & 1178 & 0.333 & 0.624 & -1.300 & -0.029 & 0.462 & 0.892 & 1.000 \\
Professional Experience (Years) & 1178 & 8.854 & 6.332 & 2.000 & 4.000 & 8.000 & 12.000 & 30.000 \\
Prior Organizational Seniority$^{\dagger}$ & 1178 & 4.295 & 2.839 & 1.000 & 2.000 & 4.000 & 6.000 & 9.000 \\

\midrule
\multicolumn{9}{l}{\textbf{Panel C: Utterance Analytic Sample (N = 2204)}} \\
\midrule
Extraneous Load (EL) & 2204 & -0.004 & 0.667 & -1.336 & -0.389 & -0.039 & 0.426 & 2.110 \\
Intrinsic Load (IL) & 2204 & 0.030 & 0.848 & -1.755 & -0.465 & 0.273 & 0.641 & 1.408 \\
Content Usage (CCU) & 2204 & 0.048 & 1.853 & -0.762 & -0.762 & -0.762 & -0.047 & 19.978 \\

\midrule
\multicolumn{9}{l}{\textbf{Panel D: Utterance Lag-Model Sample (N = 1940)}} \\
\midrule
Extraneous Load (EL) & 1940 & -0.009 & 0.658 & -1.336 & -0.389 & -0.042 & 0.417 & 2.110 \\
Intrinsic Load (IL) & 1940 & 0.070 & 0.819 & -1.755 & -0.318 & 0.308 & 0.641 & 1.408 \\
Content Usage (CCU) & 1940 & 0.037 & 1.823 & -0.762 & -0.762 & -0.762 & -0.053 & 18.598 \\
Utterance Length (chars) & 1940 & 1621.363 & 2431.761 & 3.000 & 85.750 & 567.000 & 2354.750 & 29869.000 \\
Sequence Position (seq) & 1940 & 74.007 & 76.050 & 3.000 & 21.000 & 45.000 & 99.000 & 359.000 \\
Relative Position (0--1) & 1940 & 0.516 & 0.277 & 0.011 & 0.279 & 0.513 & 0.750 & 1.000 \\

\bottomrule
\end{tabular}

\begin{tablenotes}[flushleft]
\footnotesize
\item \textit{Notes.} All variables are reported in raw (pre-standardized) units. Output Quality remains on its original ordinal grading scale. 
\item $^{*}$Bidirectional PSC is reported in its raw form, where higher values indicate greater contextual coherence. In subsequent analyses, PSC is standardized and inverted (multiplied by $-1$) to align directionally with the extraneous load construct (higher values indicating greater 
load). \\
\item[$\dagger$] See \S~8.4 in the Supplemental Materials for information on variable construction. 
\end{tablenotes}

\end{threeparttable}
\end{table}

\subsection{Construction of the Participant-Subtask Analytic Sample}

\begin{table}[ht]
\centering
\caption{Construction of the Participant--Subtask Analytic Sample}
\label{tab:sample_construction}
\begin{tabular}{lr}
\toprule
\textbf{Stage} & \textbf{Observations} \\
\midrule
Total potential participant--subtask pairs ($34 \times 101$) & 3,434 \\
\addlinespace[4pt]
\multicolumn{2}{l}{\textit{Grader-Based Attempt Filter}} \\
\quad Graded as attempted (completeness $\geq 1$) & 3,394 \\
\addlinespace[4pt]
\multicolumn{2}{l}{\textit{Transcript Processing}} \\
\quad Participant--subtask pairs mentioned in transcripts & 1,587 \\
\quad Pipeline-emitted observed panel (after graph construction \& aggregation) & 1,395 \\
\quad Attempted within observed panel & 1,366 \\
\addlinespace[4pt]
\multicolumn{2}{l}{\textit{Behavioral Metric Availability}} \\
\quad Exclude: survey-only intrinsic load regime & (22) \\
\quad Main analytic sample & 1,344 \\
\addlinespace[4pt]
\multicolumn{2}{l}{\textit{Bidirectional PSC Requirement}} \\
\quad Exclude: no PSC record available for merge & (166) \\
\quad PSC-based analytic sample & 1,178 \\
\bottomrule
\end{tabular}

\vspace{6pt}
\raggedright
\footnotesize
\textit{Notes.} The theoretical frame comprises all graded subtasks crossed with all participants. 
The pipeline's observed panel (1,395) is produced through transcript processing, graph 
construction, and aggregation---it is not a simple filter on the 1,587 transcript-mentioned 
pairs but reflects additional pipeline-internal constraints. ``Survey-only intrinsic load 
regime'' refers to observations where all six role-aggregate behavioral components (element 
interactivity, phase spread, and dependency debt for both prompt and response) are missing, 
causing the intrinsic load composite to collapse to survey-based measurement alone; these 
are excluded to prevent mixing informational bases during standardization. Bidirectional PSC 
requires sufficient repeated subtask mentions within a conversation to construct a stable 
contextual memory representation; subtasks lacking this support have no PSC record and are 
excluded from PSC-based models.
\end{table}

Table~\ref{tab:sample_construction} details construction of the participant-subtask analytic sample. The theoretical frame comprises $3,434$ participant-subtask pairs (34 participants $\times$ 101 graded subtasks). Restricting to subtasks graded as attempted (completeness$\geq$1) yields 3,394 observations. Separately, $1,587$ unique participant-subtask pairs are mentioned at least once across participants' transcripts. However, the pipeline's participant-subtask dataset is not a simple filter on this set; it is an observed panel produced through transcript processing, graph construction, and aggregation, yielding 1,395 participant-subtask pairs. Of these, 1,366 are graded as attempted. We next exclude 22 observations where the intrinsic load composite collapses to a survey-only measurement regime-that is, where all six role-aggregate behavioral components (element interactivity, phase spread, and dependency debt for both prompt and response) are missing. Because these cases represent a qualitatively distinct measurement regime, they are excluded to prevent mixing information bases during standardization, yielding a main analytic sample of $N=1,344$. Finally, models incorporating bidirectional PSC require a non-missing PSC record for the participant-subtask pair. For 166 of the 1,344 observations, no PSC record exists (the subtask lacked sufficient repeated mentions to construct a stable contextual memory representation), resulting in a PSC-based analytic sample of $N=1,178$.

\subsection{Construction of the Utterance-Level Analytic Sample}

\begin{table}[!htbp]
    \centering 
    \scriptsize
    \setlength{\tabcolsep}{4pt}
    \begin{threeparttable}
    \caption{Construction of the Utterance-Level Analytic Sample}
    \label{tab:analytic_sample_utt_level}
    \begin{tabular}{l r}
    \toprule
    \textbf{Stage} & \textbf{Utterances} \\
    \midrule
    Raw transcripts & 2352 \\
    Remove erroneous response & 2351 \\
    \midrule
    \multicolumn{2}{l}{\textit{Missing Current Extraneous Load}} \\
    Missing extraneous load  & 147 \\
    Remaining after current-load filter & 2204 \\
    \midrule
    \multicolumn{2}{l}{\textit{Additional Omissions in Dynamic Lag Models}} \\
    Missing lag predictors (L1/L2 intrinsic or extraneous load) & 321 \\
    Overlap: missing current load and missing lags & (57) \\
    \midrule
    \textbf{Final utterance analysis sample} & \textbf{1940} \\
    \bottomrule    
    \end{tabular}
    \begin{tablenotes}[flushleft]
    \footnotesize
    \item \textit{Notes.} Extraneous load is undefined when the unpooled subtask--concept alignment metric cannot be computed (e.g., no concept embeddings, no subtask embeddings, or degenerate embedding overlap). Dynamic fixed-effects models additionally require non-missing one- and two-utterance lags of intrinsic and extraneous load. Because lagged values are functions of prior load, missing current load values mechanically induce missing lag predictors in subsequent utterances. The final analytic sample reflects the union of current-load and lag-based omissions.
    \end{tablenotes}
    \end{threeparttable}
\end{table}

Table~\ref{tab:analytic_sample_utt_level} details construction of the utterance-level analytic sample. We begin with $2,352$ utterances ($1,176$ dyadic turns across 34 participants). One response was removed because the LLM returned an error message rather than a substantive reply. 

The extraneous load measure depends on the utterance-level clarity composite derived from the unpooled subtask-concept alignment metric. This metric is undefined when alignment cannot be computed (e.g., no subtask embeddings, no concept embeddings, or degenerate embedding overlap), resulting in 147 utterances with missing extraneous load and  remaining sample of $N=2,204$. 

Dynamic fixed-effects models further require complete data on current extraneous load and both one- and two-utterance lags of intrinsic and extraneous load. Because lagged values are functions of prior load, missing current load mechanically induces missing lag predictors in subsequent utterances. Accounting for missing lag predictors ($n=321$) and overlap with missing current load ($n=57$), the final utterance-level analytic sample comprises $N=1,940$ observations.

\section{Full Model Specifications}
\subsection{Linear Interactions: Load and Content Usage}\label{app:linear-interaction-models}
Table~\ref{tab:quality_interactions_linear} reports linear interaction models testing whether the association between AI-generated content usage (AIGCU) and output quality varies as a function of intrinsic or extraneous load. Specifications progressively introduce interaction terms and participant-level controls. 

\begin{table}[t]
\centering
\scriptsize
\setlength{\tabcolsep}{3pt}
\renewcommand{\arraystretch}{1.15}
\caption{Linear interaction models predicting output quality}
\label{tab:quality_interactions_linear}
\begin{tabular}{lcccccccc}
\toprule
 & \multicolumn{8}{c}{Dependent Variable: Output Quality} \\
\cmidrule(lr){2-9}
 & \multicolumn{2}{c}{AIGCU $\times$ EL--PSC}
 & \multicolumn{2}{c}{AIGCU $\times$ ICL}
 & \multicolumn{2}{c}{AIGCU $\times$ EL--PSC + ICL}
 & \multicolumn{2}{c}{Both Interactions} \\
\cmidrule(lr){2-3}\cmidrule(lr){4-5}\cmidrule(lr){6-7}\cmidrule(lr){8-9}
 & (1) No Ctrls & (2) + Ctrls
 & (3) No Ctrls & (4) + Ctrls
 & (5) No Ctrls & (6) + Ctrls
 & (7) No Ctrls & (8) + Ctrls \\
\midrule
\multicolumn{9}{l}{\textit{Main effects}} \\

AIGCU
 & 0.283*** & 0.281*** 
 & 0.310*** & 0.308*** 
 & 0.323*** & 0.322*** 
 & 0.334*** & 0.330*** \\
 & (0.069)  & (0.069)
 & (0.059)  & (0.059)
 & (0.070)  & (0.070)
 & (0.073)  & (0.073) \\

Extraneous (EL--PSC)
 & $-$0.435*** & $-$0.447*** 
 &        &        
 & $-$0.459*** & $-$0.471*** 
 & $-$0.461*** & $-$0.473*** \\
 & (0.081)     & (0.080)
 &        &        
 & (0.081)     & (0.080)
 & (0.081)     & (0.080) \\

Intrinsic (ICL)
 &        &        
 & $-$0.174*** & $-$0.174*** 
 & $-$0.144*** & $-$0.145*** 
 & $-$0.152*** & $-$0.152*** \\
 &        &        
 & (0.045)     & (0.045)
 & (0.043)     & (0.043)
 & (0.046)     & (0.046) \\

\addlinespace
\multicolumn{9}{l}{\textit{Interaction terms}} \\

AIGCU $\times$ EL--PSC
 & 0.076 & 0.082 
 &        &        
 & 0.051 & 0.056 
 & 0.046 & 0.052 \\
 & (0.074) & (0.074)
 &        &        
 & (0.074) & (0.074)
 & (0.075) & (0.075) \\

AIGCU $\times$ ICL
 &        &        
 & $-$0.012 & $-$0.009 
 &        &        
 & $-$0.029 & $-$0.025 \\
 &        &        
 & (0.058)  & (0.058)
 &        &        
 & (0.059)  & (0.059) \\

\addlinespace
\multicolumn{9}{l}{\textit{Controls}} \\

Professional Exp.\ (Years)
 &        & $-$0.522** 
 &        & $-$0.499* 
 &        & $-$0.528** 
 &        & $-$0.526** \\
 &        & (0.187)
 &        & (0.213)
 &        & (0.184)
 &        & (0.184) \\

Prior Org.\ Seniority
 &        & 0.370* 
 &        & 0.292 
 &        & 0.355* 
 &        & 0.353* \\
 &        & (0.190)
 &        & (0.222)
 &        & (0.187)
 &        & (0.187) \\

\addlinespace
Intercept
 & 2.776*** & 2.784*** 
 & 2.877*** & 2.882*** 
 & 2.786*** & 2.791*** 
 & 2.793*** & 2.797*** \\
 & (0.168)  & (0.151)
 & (0.183)  & (0.171)
 & (0.166)  & (0.149)
 & (0.167)  & (0.150) \\

\midrule
Observations   & 1,178 & 1,178 & 1,344 & 1,344 & 1,178 & 1,178 & 1,178 & 1,178 \\
Participants   & 33    & 33    & 33    & 33    & 33    & 33    & 33    & 33    \\

AIC
 & 4293.0 & 4290.2
 & 4901.3 & 4900.3
 & 4283.8 & 4280.7
 & 4285.6 & 4282.5 \\

Log-likelihood
 & $-$2140.5 & $-$2137.1
 & $-$2444.7 & $-$2442.1
 & $-$2134.9 & $-$2131.4
 & $-$2134.8 & $-$2131.3 \\

Estimation     & ML & ML & ML & ML & ML & ML & ML & ML \\
\bottomrule
\end{tabular}
\vspace{0.6em}
\footnotesize \\
\textit{Notes.} Linear mixed-effects models estimated via maximum likelihood with participant-level random intercepts. Fixed-effects inference uses Satterthwaite-approximated degrees of freedom. All continuous predictors are $z$-standardized. EL--PSC denotes the bidirectional prompt--response extraneous load composite. Across specifications, interaction terms between AIGCU and cognitive load (AIGCU $\times$ EL--PSC; AIGCU $\times$ ICL) are small and statistically non-significant. Inclusion of interaction terms does not materially improve model fit. Standard errors in parentheses. * $p \leq 0.10$, ** $p \leq 0.05$, *** $p \leq 0.01$.
\end{table}

\subsection{Expertise Heterogeneity}\label{app:expertise-hetero}
Table~\ref{tab:expertise_heterogeneity_full} reports cross-level interaction models testing whether the associations between cognitive load, AI-generated content usage (AIGCU), and output quality vary as a function of professional experience.

\begin{table}[t]
\centering
\footnotesize
\setlength{\tabcolsep}{3.5pt}
\renewcommand{\arraystretch}{1.15}
\caption{Expertise heterogeneity in the associations between cognitive load, AI content usage, and output quality.}
\label{tab:expertise_heterogeneity_full}
\begin{tabular}{lccccc}
\toprule
 & \multicolumn{5}{c}{Dependent Variable: Output Quality} \\
\cmidrule(lr){2-6}
 & (1) Exp $\times$ EL
 & (2) Exp $\times$ AIGCU
 & (3) Exp $\times$ ICL
 & (4) Joint Two-Way
 & (5) Three-Way \\
\midrule
\multicolumn{6}{l}{\textit{Main effects}} \\
AI-Generated Content Usage (AIGCU)
 & 0.332*** & 0.398*** & 0.350*** & 0.386*** & 0.375*** \\
 & (0.060)  & (0.062)  & (0.059)  & (0.062)  & (0.072)  \\
Intrinsic Cognitive Load (ICL)
 & $-$0.150*** & $-$0.153*** & $-$0.149*** & $-$0.156*** & $-$0.154*** \\
 & (0.043)     & (0.043)     & (0.043)     & (0.042)     & (0.043)     \\
Extraneous Load (EL--PSC)
 & $-$0.467*** & $-$0.468*** & $-$0.485*** & $-$0.438*** & $-$0.442*** \\
 & (0.078)     & (0.078)     & (0.078)     & (0.078)     & (0.081)     \\
\addlinespace
\multicolumn{6}{l}{\textit{Cross-level interactions}} \\
Experience $\times$ EL--PSC
 & 0.145** &        &        & 0.208*** & 0.208*** \\
 & (0.074) &        &        & (0.077)  & (0.078)  \\
Experience $\times$ AIGCU
 &        & $-$0.121** &        & $-$0.159*** & $-$0.180*** \\
 &        & (0.050)    &        & (0.051)     & (0.062)     \\
Experience $\times$ ICL
 &        &        & $-$0.004 &        &        \\
 &        &        & (0.043)  &        &        \\
AIGCU $\times$ EL--PSC
 &        &        &        &        & 0.026 \\
 &        &        &        &        & (0.075) \\
Experience $\times$ AIGCU $\times$ EL--PSC
 &        &        &        &        & 0.048 \\
 &        &        &        &        & (0.076) \\
\addlinespace
\multicolumn{6}{l}{\textit{Controls}} \\
Professional Experience (Years)
 & $-$0.504*** & $-$0.540*** & $-$0.525*** & $-$0.517*** & $-$0.539*** \\
 & (0.187)    & (0.181)    & (0.184)    & (0.183)    & (0.184)    \\
Prior Organizational Seniority
 & 0.343* & 0.348* & 0.353* & 0.331* & 0.329* \\
 & (0.190) & (0.184) & (0.187) & (0.186) & (0.184) \\
\addlinespace
Intercept
 & 2.805*** & 2.848*** & 2.814*** & 2.844*** & 2.829*** \\
 & (0.148)  & (0.144)  & (0.146)  & (0.145)  & (0.148)  \\
\midrule
Observations   & 1{,}178 & 1{,}178 & 1{,}178 & 1{,}178 & 1{,}178 \\
Participants   & 33    & 33    & 33    & 33    & 33    \\
AIC            & 4277.5 & 4275.3 & 4281.3 & 4270.0 & 4273.5 \\
Log-likelihood & $-$2129.7 & $-$2128.7 & $-$2131.6 & $-$2125.0 & $-$2124.7 \\
Estimation     & ML & ML & ML & ML & ML \\
\bottomrule
\end{tabular}
\vspace{0.6em}

\footnotesize
\textit{Notes.} Linear mixed-effects models estimated via maximum likelihood with participant-level random intercepts. Fixed-effects inference uses Satterthwaite-approximated degrees of freedom. All continuous predictors are $z$-standardized. Cross-level interaction terms test whether within-participant associations between load (or AIGCU) and output quality vary as a function of professional experience (between-participant moderator). Columns~(1)--(3) introduce two-way interactions separately; Column~(4) includes both significant two-way experience interactions jointly; Column~(5) adds the AIGCU $\times$ EL--PSC interaction and three-way term for completeness. * $p \leq 0.10$, ** $p \leq 0.05$, *** $p \leq 0.01$.
\end{table}

\begin{table}[H]
\centering
\footnotesize
\setlength{\tabcolsep}{5pt}
\renewcommand{\arraystretch}{1.15}
\caption{Split-sample comparison: core interaction model by professional experience group.}
\label{tab:expertise_split}
\begin{tabular}{lccc}
\toprule
 & \multicolumn{2}{c}{Output Quality} & \\
\cmidrule(lr){2-3}
 & (1) Low Experience & (2) High Experience & $\Delta$ \\
\midrule
\multicolumn{4}{l}{\textit{Main effects}} \\

AIGCU
 & 0.279** & 0.359** & $+0.080$ \\
 & (0.087)  & (0.117)  &          \\

ICL
 & $-$0.117* & $-$0.198* & $-$0.081 \\
 & (0.047)    & (0.081)    &          \\

EL--PSC
 & $-$0.607*** & $-$0.220 & $+0.387$ \\
 & (0.091)     & (0.147)  &          \\

\addlinespace
\multicolumn{4}{l}{\textit{Interaction}} \\

AIGCU $\times$ EL--PSC
 & 0.027 & 0.022 & $-0.005$ \\
 & (0.086) & (0.134) &        \\

\addlinespace
Intercept
 & 3.146*** & 2.510*** & \\
 & (0.206)  & (0.364)  & \\

\midrule
Observations   & 657  & 521  & \\
Participants   & 20   & 13   & \\
AIC            & 2222.9 & 2028.7 & \\
Log-likelihood & $-$1103.4 & $-$1006.4 & \\
\bottomrule
\end{tabular}
\vspace{0.6em}
\footnotesize \\
\textit{Notes.} The sample is split at the median of standardized professional experience. Both models are estimated via maximum likelihood with participant-level random intercepts. The $\Delta$ column reports the difference (High~$-$~Low) in point estimates; these are descriptive comparisons, not formal tests of coefficient equality. For less experienced professionals, extraneous load is strongly negatively associated with quality ($\beta = -0.607$, $p < 0.001$), roughly three times the magnitude observed for more experienced professionals ($\beta = -0.220$, $p = 0.135$). The AIGCU $\times$ EL--PSC interaction is null in both subsamples (Low: $\beta = 0.027$, $p = 0.753$; High: $\beta = 0.022$, $p = 0.870$), confirming that the expertise-dependent pattern operates through the direct load--quality association rather than through differential AIGCU buffering. AIGCU is positively associated with quality in both groups, with a descriptively larger coefficient for experienced professionals ($\beta = 0.359$ vs.\ $0.279$). Standard errors in parentheses. \textsuperscript{***}$p\leq0.01$, \textsuperscript{**}$p\leq0.05$, \textsuperscript{*}$p\leq0.1$.
\end{table}

\subsection{Moderated Mediation Models}
Table~\ref{tab:moderated_mediation_paths} reports the moderated mediation structural equation model testing whether professional experience moderates both the load $\rightarrow$ AIGCU pathway and the AIGCU $\rightarrow$ quality pathway.

\begin{table}[t]
\centering
\small
\setlength{\tabcolsep}{5pt}
\renewcommand{\arraystretch}{1.15}
\caption{Moderated mediation model: experience moderates the load--AIGCU--quality pathway.}
\label{tab:moderated_mediation_paths}
\begin{tabular}{lcc}
\toprule
 & (a) Mediator: AIGCU
 & (b) Outcome: Quality \\
\midrule
\multicolumn{3}{l}{\textit{Panel A: Main paths}} \\
ICL ($a_1$ / $d_{1}'$)
 & 0.126*** & $-$0.171*** \\
 & (0.020)  & (0.046)     \\
EL--PSC ($a_2$ / $d_{2}'$)
 & 0.602*** & $-$0.667*** \\
 & (0.035)  & (0.073)     \\
AIGCU ($b$)
 &          & 0.405***    \\
 &          & (0.060)     \\
\addlinespace
\multicolumn{3}{l}{\textit{Panel B: Experience moderation}} \\
Exp $\times$ ICL ($a_{1,\text{mod}}$ / $d_{1,\text{mod}}'$)
 & $-$0.008   & $-$0.021  \\
 & (0.025)    & (0.055)   \\
Exp $\times$ EL ($a_{2,\text{mod}}$ / $d_{2,\text{mod}}'$)
 & 0.294*** & 0.243*** \\
 & (0.045)  & (0.073)  \\
Exp $\times$ AIGCU ($b_{\text{mod}}$)
 &          & $-$0.213*** \\
 &          & (0.044)     \\
\addlinespace
\multicolumn{3}{l}{\textit{Panel C: Controls}} \\
Professional Experience
 & 0.088*** & $-$0.605*** \\
 & (0.024)  & (0.044)     \\
Seniority
 & 0.038* & 0.287*** \\
 & (0.022) & (0.048) \\
Log(Turns)
 & 0.340*** & 0.050    \\
 & (0.024)  & (0.055)  \\
\addlinespace
Intercept
 & $-$0.022 & 2.733*** \\
 & (0.022)  & (0.049)  \\
\midrule
Observations & 1,178 & 1,178 \\
Participants & 33 & 33 \\
\bottomrule
\end{tabular}
\vspace{0.6em}

\footnotesize
\textit{Notes.} Structural equation model estimated via maximum likelihood with percentile bootstrap standard errors (1{,}000 draws), clustered on participant. All continuous predictors are $z$-standardized. Column (a) models AIGCU as a function of intrinsic and extraneous load, their interactions with professional experience, and controls. Column (b) models output quality as a function of AIGCU, both load types, their experience interactions, and controls. Panel A reports main path coefficients; Panel B reports moderation terms; Panel C reports control variables. Standard errors in parentheses. \textsuperscript{***}$p\leq0.01$, \textsuperscript{**}$p\leq0.05$, \textsuperscript{*}$p\leq0.1$.
\end{table}

\subsection{Conditional Indirect Associations}
Table~\ref{tab:conditional_effects} reports conditional path coefficients, indirect effects, direct effects, and total effects evaluated at $\pm 1$ standard deviation of professional experience, based on the moderated mediation model.

\begin{table}[t]
\centering
\small
\setlength{\tabcolsep}{4pt}
\renewcommand{\arraystretch}{1.15}
\caption{Conditional indirect, direct, and total effects on quality at $\pm 1$ SD of professional experience.}
\label{tab:conditional_effects}
\begin{tabular}{lccc}
\toprule
 & Low Exp ($-1$ SD) & High Exp ($+1$ SD) & $\Delta$ \\
\midrule
\multicolumn{4}{l}{\textit{Panel A: Conditional a-paths (Load $\to$ AIGCU)}} \\
ICL $\to$ AIGCU
 & 0.133*** & 0.118*** &  \\
 & [0.086, 0.180] & [0.052, 0.200] &  \\
EL $\to$ AIGCU
 & 0.308*** & 0.896*** &  \\
 & [0.216, 0.403] & [0.766, 1.033] &  \\
\addlinespace
\multicolumn{4}{l}{\textit{Panel B: Conditional b-path (AIGCU $\to$ Quality)}} \\
AIGCU $\to$ Quality
 & 0.618*** & 0.192*** &  \\
 & [0.467, 0.784] & [0.070, 0.331] &  \\
\addlinespace
\multicolumn{4}{l}{\textit{Panel C: Conditional indirect effects (Load $\to$ AIGCU $\to$ Quality)}} \\
ICL $\to$ AIGCU $\to$ Quality
 & 0.082*** & 0.023** & $-$0.060*** \\
 & [0.051, 0.116] & [0.007, 0.048] & [$-$0.097, $-$0.018] \\
EL $\to$ AIGCU $\to$ Quality
 & 0.190*** & 0.172*** & $-$0.018 \\
 & [0.129, 0.259] & [0.061, 0.314] & [$-$0.136, 0.134] \\
\addlinespace
\multicolumn{4}{l}{\textit{Panel D: Conditional direct effects (Load $\to$ Quality, net of AIGCU)}} \\
ICL $\to$ Quality
 & $-$0.150*** & $-$0.192** &  \\
 & [$-$0.263, $-$0.033] & [$-$0.404, $-$0.070] &  \\
EL $\to$ Quality
 & $-$0.909*** & $-$0.424*** &  \\
 & [$-$1.067, $-$0.752] & [$-$0.678, $-$0.180] &  \\
\addlinespace
\multicolumn{4}{l}{\textit{Panel E: Conditional total effects (direct + indirect)}} \\
ICL $\to$ Quality
 & $-$0.067 & $-$0.169** &  \\
 & [$-$0.174, 0.043] & [$-$0.372, $-$0.053] &  \\
EL $\to$ Quality
 & $-$0.719*** & $-$0.252*** &  \\
 & [$-$0.864, $-$0.575] & [$-$0.434, $-$0.052] &  \\
\midrule
\multicolumn{4}{l}{AIC = 15,233.6 \quad BIC = 15,421.2 \quad SRMR = 0.050 \quad $\Delta$AIC vs.\ baseline = $-$94.3 \quad $\Delta$BIC = $-$53.7} \\
\bottomrule
\end{tabular}
\vspace{0.6em}

\footnotesize
\textit{Notes.} Conditional effects are derived from the moderated mediation model (Table~\ref{tab:moderated_mediation_paths}) by evaluating professional experience at $\pm 1$ SD from the mean. Indirect effects are computed as products of conditional $a$- and $b$-paths. Direct and total effects correspond to conditional $d'$ and $(a \times b + d')$ components. Brackets report percentile bootstrap 95\% confidence intervals (1{,}000 draws, clustered on participant). $\Delta$ denotes the difference between high- and low-experience estimates. Model fit statistics correspond to the full moderated mediation specification. \textsuperscript{***}$p\leq0.01$, \textsuperscript{**}$\leq0.05$, \textsuperscript{*}$p\leq0.1$.
\end{table}

\subsection{Dynamic Fixed-Effects Models of Extraneous Load}\label{app:temp-dynamics}
Tables~\ref{tab:rq2_full_models} reports the full model estimates underlying the dynamic analyses presented in the main text. Specifications examine autoregressive persistence, cross-speaker spillover, and the propagation of extreme extraneous load shocks. Table~\ref{tab:el_gap_decomposition} decomposes the attenuation of the unconditional prompt-response extraneous load gap by separately introducing AR(2) terms and utterance length controls. AR(2) persistence alone eliminates the speaker difference ($\beta=0.007$, $p=0.96$), confirming that the raw asymmetry reflects temporal accumulation.

\begin{table}[h]
\centering
\small
\setlength{\tabcolsep}{4pt}
\renewcommand{\arraystretch}{1.15}
\caption{Full model estimates underlying Table~\ref{tab:rq2_core_dynamics}: Dynamic fixed-effects models of extraneous cognitive load.}
\label{tab:rq2_full_models}
\begin{tabular}{lccc}
\toprule
 & \textbf{(1)} & \textbf{(2)} & \textbf{(3)} \\
 & Base AR & Role Interactions & IL-Partialed \\
\midrule
\multicolumn{4}{l}{\textit{Speaker role}} \\
$\Ind[\text{Prompt}]$ 
 & $-0.041$ & $-0.041$ & $-0.019$ \\
 & $(0.490)$ & $(0.502)$ & $(0.763)$ \\
\addlinespace[4pt]

\multicolumn{4}{l}{\textit{Cross-speaker dynamics (Lag 1)}} \\
$\text{EL}_{t-1}$ 
 & $0.061^{***}$ & $0.015$ & $0.034$ \\
 & $(0.0009)$ & $(0.668)$ & $(0.374)$ \\
$\text{EL}_{t-1} \times \Ind[\text{Prompt}_{t-1}]$ 
 &  & $0.074$ & $0.059$ \\
 &  & $(0.135)$ & $(0.264)$ \\
\addlinespace[4pt]

\multicolumn{4}{l}{\textit{Within-speaker persistence (Lag 2)}} \\
$\text{EL}_{t-2}$ 
 & $0.220^{***}$ & $0.170^{***}$ & $0.145^{***}$ \\
 & $(<0.001)$ & $(<0.001)$ & $(<0.001)$ \\
$\text{EL}_{t-2} \times \Ind[\text{Prompt}_{t-2}]$ 
 &  & $0.098$ & $0.110^{*}$ \\
 &  & $(0.106)$ & $(0.072)$ \\
\addlinespace[4pt]

\multicolumn{4}{l}{\textit{Intrinsic load controls}} \\
$\text{IL}_{t-1}$ 
 &  &  & $-0.149^{***}$ \\
 &  &  & $(<0.001)$ \\
$\text{IL}_{t-2}$ 
 &  &  & $0.043$ \\
 &  &  & $(0.170)$ \\
\addlinespace[4pt]

\multicolumn{4}{l}{\textit{Controls}} \\
Utterance length 
 & $0.309^{***}$ & $0.309^{***}$ & $0.295^{***}$ \\
 & $(<0.001)$ & $(<0.001)$ & $(<0.001)$ \\
Sequence position 
 & $0.001^{**}$ & $0.001^{**}$ & $0.001^{**}$ \\
 & $(0.032)$ & $(0.032)$ & $(0.021)$ \\
Relative position 
 & $-0.329^{**}$ & $-0.329^{**}$ & $-0.349^{***}$ \\
 & $(0.012)$ & $(0.012)$ & $(0.008)$ \\
\addlinespace[2pt]

Participant FE & Yes & Yes & Yes \\
AR(2) terms & Yes & Via interactions & Via interactions \\
\midrule
$N$ & 1{,}940 & 1{,}940 & 1{,}940 \\
\bottomrule
\end{tabular}
\vspace{0.6em}
\footnotesize \\
\textit{Notes.} Dependent variable: $z$-standardized extraneous load ($z\_el\_utt$). All models estimated via OLS with participant fixed effects and Driscoll--Kraay standard errors; $p$-values in parentheses. Column~(1) reports the baseline AR(2) coefficients as main effects. Columns~(2)--(3) interact lagged EL terms with the speaker role of the lagged utterance to distinguish same-speaker persistence from cross-speaker spillover. Linear combinations reported in Table~\ref{tab:rq2_core_dynamics} are derived from Column~(2). Column~(3) adds lagged intrinsic load to assess whether EL dynamics persist net of IL spillover. 
$\Ind[\text{Prompt}_{t-k}]$ equals 1 if the utterance at $t-k$ is a user prompt (0 if a model response). \textsuperscript{***}$p\leq0.01$, \textsuperscript{**}$p\leq0.05$, \textsuperscript{*}$p\leq0.1$.
\end{table}

\subsection{Cognitive Load and Content Usage Dynamics}\label{app:cog-load-content-usage-utt}

Table~\ref{tab:content_usage_full} reports the full fixed-effects specifications underlying the cognitive load and content usage dynamics presented in the main text. The models examine spike-triggered content usage responses, cross-speaker load transmission, and autoregressive persistence across consecutive turns.
\begin{table}[h]
\centering
\footnotesize
\setlength{\tabcolsep}{3pt}
\renewcommand{\arraystretch}{0.95}
\begin{threeparttable}
\caption{Full model estimates underlying Table~\ref{tab:content_usage_core}: Content usage and cognitive load dynamics.}
\label{tab:content_usage_full}
\begin{tabular}{lcccccc}
\toprule
 & \textbf{(1)} & \textbf{(2)} & \textbf{(3)} & \textbf{(4)} & \textbf{(5)} & \textbf{(6)} \\
 & Spikes + & Interaction & No Prompt & AIGCU$\rightarrow$ & EL $\rightarrow$ & AIGCU$\rightarrow$ \\
 & Prompt & & Load & Prompt EL & Resp AIGCU& AIGCU\\
\midrule
\multicolumn{7}{l}{\textit{Key predictors}} \\
IL spike
 & $0.214^{**}$ & $0.213^{**}$ & $0.279^{***}$ & & & \\
 & $(0.019)$ & $(0.020)$ & $(0.002)$ & & & \\
EL spike
 & $0.188^{**}$ & $0.190^{**}$ & $0.185^{*}$ & & & \\
 & $(0.028)$ & $(0.038)$ & $(0.065)$ & & & \\
Prompt IL (prev.)
 & $0.106^{***}$ & $0.107^{***}$ & & & & \\
 & $(0.003)$ & $(0.003)$ & & & & \\
Prompt EL (prev.)
 & $0.042$ & $0.041$ & & & & \\
 & $(0.244)$ & $(0.249)$ & & & & \\
EL spike $\times$ Prompt EL
 & & $0.011$ & & & & \\
 & & $(0.897)$ & & & & \\
Prev.\ CCU
 & & & & $0.032$ & & \\
 & & & & $(0.288)$ & & \\
Prev.\ EL
 & & & & $0.090^{**}$ & $0.079^{*}$ & \\
 & & & & $(0.042)$ & $(0.062)$ & \\
Prev.\ IL
 & & & & $0.015$ & $0.102^{***}$ & \\
 & & & & $(0.666)$ & $(0.001)$ & \\
\addlinespace[4pt]
\multicolumn{7}{l}{\textit{Autoregressive controls}} \\
$\text{CCU}_{t-1}$
 & $0.112^{***}$ & $0.112^{***}$ & $0.106^{***}$ & & & $0.115^{***}$ \\
 & $(0.006)$ & $(0.006)$ & $(0.005)$ & & & $(0.004)$ \\
$\text{CCU}_{t-2}$
 & $0.126^{*}$ & $0.126^{*}$ & $0.102^{*}$ & & & $0.109^{*}$ \\
 & $(0.056)$ & $(0.056)$ & $(0.090)$ & & & $(0.099)$ \\
$\text{EL}_{t-1}$
 & $0.009$ & $0.009$ & $0.015$ & & & $0.005$ \\
 & $(0.783)$ & $(0.779)$ & $(0.601)$ & & & $(0.877)$ \\
$\text{EL}_{t-2}$
 & $0.042$ & $0.042$ & $0.030$ & & & \\
 & $(0.279)$ & $(0.271)$ & $(0.484)$ & & & \\
$\text{IL}_{t-1}$
 & $0.017$ & $0.017$ & $0.045$ & & & $0.040$ \\
 & $(0.690)$ & $(0.697)$ & $(0.267)$ & & & $(0.302)$ \\
$\text{IL}_{t-2}$
 & $-0.017$ & $-0.017$ & $0.038$ & & & \\
 & $(0.685)$ & $(0.685)$ & $(0.272)$ & & & \\
\addlinespace[4pt]
\multicolumn{7}{l}{\textit{Controls}} \\
Utterance length
 & $0.449^{***}$ & $0.449^{***}$ & $0.394^{***}$ & $0.307^{**}$ & $0.521^{***}$ & $0.420^{***}$ \\
 & $(<0.001)$ & $(<0.001)$ & $(<0.001)$ & $(0.043)$ & $(<0.001)$ & $(<0.001)$ \\
Sequence position
 & $0.003$ & $0.003$ & $0.003^{*}$ & $0.001$ & $0.002^{**}$ & $0.003$ \\
 & $(0.175)$ & $(0.176)$ & $(0.098)$ & $(0.672)$ & $(0.016)$ & $(0.117)$ \\
Relative position
 & $0.189$ & $0.190$ & $0.053$ & $0.095$ & $-0.076$ & $0.025$ \\
 & $(0.373)$ & $(0.367)$ & $(0.777)$ & $(0.699)$ & $(0.664)$ & $(0.893)$ \\
\addlinespace[2pt]
Participant FE & Yes & Yes & Yes & Yes & Yes & Yes \\
Sample & Resp. & Resp. & Resp. & Prompts & Resp. & Resp. \\
$N$ & 971 & 971 & 1{,}068 & 1{,}008 & 1{,}053 & 1{,}079 \\
\bottomrule
\end{tabular}
\begin{tablenotes}[flushleft]
\footnotesize
\item \textit{Notes.} All models estimated via OLS with participant fixed effects and Driscoll--Kraay standard errors; $p$-values in parentheses. Columns~(1)--(3) use the response panel and correspond to Panel~A of Table~\ref{tab:content_usage_core} (DV: response-level $z$-standardized content usage, CCU). Column~(1) includes contemporaneous response load spikes (IL, EL) and preceding prompt load ($P(t)$); Column~(2) adds the EL spike $\times$ Prompt EL interaction; Column~(3) omits preceding prompt load terms. Columns~(4)--(6) correspond to Panel~B. Column~(4) estimates Link~A (DV: prompt-level $z$-standardized extraneous load at $P(t{+}1)$; predictors from the preceding response $R(t)$). Column~(5) estimates Link~B (DV: response-level AIGCUat $R(t{+}1)$; predictors from the preceding prompt $P(t{+}1)$). Column~(6) estimates the reduced form (DV: response-level AIGCUat $R(t{+}1)$; predictor: AIGCUat $R(t)$). Autoregressive terms in Columns~(1)--(3) and~(6) are response-panel lags; Column~(4) uses cross-speaker lags (response $\rightarrow$ prompt), and Column~(5) uses prompt $\rightarrow$ response lags. \textsuperscript{***}$p\leq0.01$, \textsuperscript{**}$p\leq0.05$, \textsuperscript{*}$p\leq0.1$.
\end{tablenotes}
\end{threeparttable}
\end{table}

\section{Supplementary Analyses}

\subsubsection{Prompt-Response Differences in Cognitive Load}\label{app:t-test}
Table~\ref{tab:rq2rq3_paired_ttests} reports paired t-tests comparing each participant's mean cognitive load in prompts versus responses. Both intrinsic and extraneous load are significantly lower in prompts than in responses ($\Delta=-0.322$ and $\Delta=-0.540$, respectively; both $p<0.001$). 

\begin{table}[t]
  \centering
  \begin{threeparttable}
  \caption{Prompt--Response Differences in Cognitive Load (Paired t-tests by Participant)}
  \label{tab:rq2rq3_paired_ttests}
  \begin{tabular}{l S[table-format=-1.3] S[table-format=1.4]}
    \toprule
    \textbf{Outcome} 
    & {\textbf{Mean Difference $\Delta$ (Prompt$-$Response)}} 
    & {\textbf{$p$-value}} \\
    \midrule
    Intrinsic Load  
      & {-0.322\textsuperscript{***}} 
      & 0.0000 \\
    Extraneous Load 
      & {-0.540\textsuperscript{***}} 
      & 0.0000 \\
    \bottomrule
  \end{tabular}
  \begin{tablenotes}[flushleft]
    \footnotesize
    \item \textit{Notes.} Paired t-tests are computed at the participant level, comparing each participant's mean cognitive load during prompts versus responses. The analysis includes 2{,}351 utterances from 34 participants. Negative values indicate lower cognitive load during prompts. Significance levels: \textsuperscript{***}$p\leq0.01$, \textsuperscript{**}$p\leq0.05$, \textsuperscript{*}$p\leq0.1$.
  \end{tablenotes}
  \end{threeparttable}
\end{table}

\subsection{Decomposing the Prompt-Response Extraneous Load Gap}
The raw speaker asymmetry in extraneous load could reflect a genuine structural difference between prompts and responses, or it could be an artifact of temporal accumulation and response verbosity. Table~\ref{tab:el_gap_decomposition} disentangles these explanations by progressing adding controls to a minimal fixed-effects specification. Two results stand out. First, including AR(2) lagged extraneous load terms alone reduces the prompt coefficient to near zero ($\beta=0.007$, $p=0.96$), indicating that the raw gap is almost entirely explained by autoregressive momentum-each utterance inherits load from its predecessors regardless of speaker role. Second, controlling for utterance length alone reverses the sign of the coefficient ($\beta=0.376$, $p=0.03$), suggesting that once verbosity is accounted for, prompts actually carry marginally \textit{higher} extraneous load per unit of text. The full model, which includes both AR(2) terms and length, yields a small positive but marginally significant prompt coefficient ($\beta=0.290$, $p=0.064$). Taken together, these results indicate that the unconditional prompt-response EL gap reflects temporal accumulation rather than merely role-based differences in the cognitive demands each speaker independently imposes. 
\begin{table}[ht]
\centering
\caption{Decomposing the prompt--response extraneous load gap: autoregressive dynamics versus utterance length.}
\label{tab:el_gap_decomposition}
\begin{tabular}{lcccc}
\toprule
& \multicolumn{4}{c}{Dependent Variable: Extraneous Load (z\_el\_utt)} \\
\cmidrule(lr){2-5}
& (C) Minimal & (B) + Length & (A) + AR(2) & Full \\
\midrule
\textit{Speaker role} \\
\quad 1[Prompt] & 0.097 & 0.376$^{**}$ & 0.007 & 0.290$^{*}$ \\
                & (0.600) & (0.032) & (0.964) & (0.064) \\
\addlinespace[4pt]
\textit{Autoregressive terms} \\
\quad EL$_{t-1}$ & --- & --- & \checkmark & \checkmark \\
\quad EL$_{t-2}$ & --- & --- & \checkmark & \checkmark \\
\addlinespace[4pt]
\textit{Controls} \\
\quad Utterance length & --- & \checkmark & --- & \checkmark \\
\quad Sequence position & \checkmark & \checkmark & \checkmark & \checkmark \\
\quad Relative position & \checkmark & \checkmark & \checkmark & \checkmark \\
\quad Turn number & \checkmark & \checkmark & \checkmark & \checkmark \\
\quad Participant FE & \checkmark & \checkmark & \checkmark & \checkmark \\
\addlinespace[4pt]
$\Delta$ from Minimal & --- & $+$0.279 & $-$0.089 & $+$0.193 \\
\addlinespace[2pt]
$N$ & 2{,}204 & 2{,}204 & 1{,}940 & 1{,}940 \\
\bottomrule
\end{tabular}

\vspace{6pt}
\raggedright
\footnotesize
\textit{Notes.} All models estimated via OLS with participant fixed effects and Driscoll--Kraay standard errors; $p$-values in parentheses. The dependent variable is $z$-standardized utterance-level extraneous load. Column~(C) includes only position controls and participant fixed effects. Column~(B) adds utterance length. Column~(A) adds AR(2) lagged extraneous load terms. The Full model includes both. The $\Delta$ row reports the change in the 1[Prompt] coefficient relative to the Minimal specification. Columns~(C) and~(B) share $N = 2{,}204$ (observations with non-missing extraneous load); Columns~(A) and Full share $N = 1{,}940$ (additionally requiring two non-missing lagged EL values per participant). AR(2) dynamics alone eliminate the speaker difference entirely ($\beta = 0.007$, $p = 0.96$), whereas utterance length alone reverses it ($\beta = 0.376$, $p = 0.03$). Because longer utterances naturally reference more subtasks and concepts, length partly reflects the informational density through which extraneous load operates; the reversal indicates that the raw speaker asymmetry reflects temporal accumulation rather than response verbosity. $^{*}p \leq 0.10$, $^{**}p \leq 0.05$, $^{***}p \leq 0.01$.
\end{table}

\section{Construct Validation}\label{app:construct_validation}

Herein, we have proposed metrics that attempt to measure the amount of cognitive load a user experiences in their interactions with a dialogue-based AI system. These metrics are constructed as a synthesis of the semantics encoded in the text utterances themselves--both in isolation and contextualized, the conceptual knowledge captured in the Finance Knowledge Graph, the procedural knowledge present in the task decomposition, and the behavioral dynamics (i.e., task switching, dependency debt) that inhere within participants' conversations with the AI system. It is precisely this synthesis and contextualized nature that--we believe--portends the utility of the metrics. However, this simultaneously deprives us of alternative constructs with which we may--with relative ease--validate our metrics. 

The metrics we employ to provide preliminary validation of the metrics we propose here fall short of being truly comparable on several dimensions. First, they do not explicitly claim to measure cognitive load itself, but rather correlates of cognitive load such as readability, sentence length, and word difficulty. Second, they are confined to the text of each utterance itself and therefore cannot reflect contextualized knowledge of the foregoing conversation, nor can they account for dynamics of unfolding conversation such as task switching or truncation. 

Of the metrics proposed here, we report correlations for two: our final composite measure of extraneous cognitive load and an antecedent to this composite that only incorporates semantic information of a single utterance. As a result, the latter (which we refer to as 'clutter') more faithfully reflects the inherent load posed by the text itself--irrespective of context and behavioral features. As the metrics we use for validation are largely focused on things such as syntactic complexity and readability, they only provide a basis for comparing to extraneous load which focuses on the presentation of information as opposed to the inherent complexity of a task itself. 

\begin{table}[htbp]
\centering
\small
\setlength{\tabcolsep}{6pt}
\renewcommand{\arraystretch}{1.15}
\caption{Spearman correlations between utterance-level metrics and readability / text characteristics.}
\label{tab:utt_readability_corr}
\begin{tabular}{lcccc}
\toprule
 & \multicolumn{2}{c}{\textbf{Extraneous Load Composite}} & \multicolumn{2}{c}{\textbf{Clutter}} \\
\cmidrule(lr){2-3} \cmidrule(lr){4-5}
\textbf{Metric} & $\rho$ & $n$ & $\rho$ & $n$ \\
\midrule
Flesch Reading Ease            & $-0.236^{***}$ & 2204 & $-0.385^{***}$ & 2209 \\
Flesch--Kincaid Grade          & $0.221^{***}$  & 2204 & $0.403^{***}$  & 2209 \\
SMOG Index                     & $0.244^{***}$  & 2204 & $0.436^{***}$  & 2209 \\
Difficult Words                & $0.390^{***}$  & 2204 & $0.634^{***}$  & 2209 \\
Average Sentence Length        & $0.138^{***}$  & 2204 & $0.316^{***}$  & 2209 \\
Polarity                       & $0.170^{***}$  & 2204 & $0.208^{***}$  & 2209 \\
Subjectivity                   & $0.171^{***}$  & 2204 & $0.273^{***}$  & 2209 \\
Gunning Fog Index              & $0.223^{***}$  & 2204 & $0.381^{***}$  & 2209 \\
Automated Readability Index    & $0.213^{***}$  & 2204 & $0.434^{***}$  & 2209 \\
Coleman--Liau Index            & $0.255^{***}$  & 2204 & $0.432^{***}$  & 2209 \\
Linsear Write Formula          & $0.182^{***}$  & 2204 & $0.363^{***}$  & 2209 \\
Dale--Chall Readability Score  & $0.184^{***}$  & 2204 & $0.348^{***}$  & 2209 \\
Readability Consensus          & $0.242^{***}$  & 2204 & $0.438^{***}$  & 2209 \\
McAlpine EFLAW Readability     & $0.105^{***}$  & 2204 & $0.261^{***}$  & 2209 \\
Reading Time                   & $0.373^{***}$  & 2204 & $0.617^{***}$  & 2209 \\
LLM Cognitive Load             & $0.297^{***}$  & 2204 & $0.430^{***}$  & 2209 \\
\bottomrule
\end{tabular}

\vspace{0.5em}
\footnotesize{\textit{Notes.} Entries are Spearman’s $\rho$. Pairwise complete observations used; $n$ may vary by metric. 
$^{***} p\leq0.01$, $^{**} p\leq0.05$, $^{*} p\leq0.1$.}
\end{table}

\subsubsection{Convergent Validity}
Convergent validity refers to how closely the new scale is related to other variables and other measures of the same construct \cite{krabbe2017validity}. To assess convergent validity we correlate our extraneous load composite with a battery of established text complexity measures to capture overlapping but distinct facets of processing difficulty. 

\subsubsection{Readability Metrics}

The most widely used family of readability formulas originates with the Flesch Reading Ease score \cite{flesch1948readability}, which estimates how difficult material is to read based on sentence length and syllable count; higher scores indicate easier text. This foundational measure was subsequently adapted into U.S. grade-level equivalents through the Flesch-Kincaid Grade Level formula \cite{kincaid1975derivation}, which expresses readability as the approximate U.S. grade level required to comprehend the text. Several parallel indices pursue the same objective through slightly different operationalizations: the Automated Readability Index \cite{senter1967automated} substitutes character counts for syllable counts, the Coleman-Liau Index \cite{coleman1975readability} relies on sentence and character length rather than syllable-based heuristics, and the Gunning Fog Index \cite{gunning1952fog} weights the proportion of complex (polysyllabic) words. Two additional formulas incorporate vocabulary-level information: the Dale-Chall Readability Formula \cite{dalechall1948readability} benchmarks word difficulty against a list of familiar words, and the SMOG Index \cite{mc1969smog} estimates years of education needed to comprehend a text based on polysyllabic word density. Finally, the Linsear Write metric \cite{ohayre1966gobb} distinguishes between easy and hard words using a simple two-category classification. 

\subsubsection{LLM-based Cognitive Load Estimation}
In their attempts to more efficiently facilitate streaming from LLM's \cite{xiao2025streaming}, the authors estimate Cognitive Load utilizing the Gunning-Fog Index \cite{gunning1952fog} and through LLM-based few-shot prompting. In the latter, the LLM is provided with a definition of Cognitive Load and prompted to append a cognitive load score using a special tag symbol after generating a text segment. We adapt this approach to out context by providing the LLM with context from prior utterances that serves to contextualize the utterance to inform the model's estimation of Cognitive Load. See 'LLM Cognitive Load' in Table~\ref{tab:utt_readability_corr}. 

\subsubsection{Interpretations}

Despite their different computation approaches, these measures share a common theoretical target: the surface-level linguistic complexity of text as experienced by a reader. They are therefore appropriate benchmarks for convergent validation. If our extraneous load composite captures processing costs associated with how information is organized and presented, it should correlate positively with indices of linguistic complexity (and negatively with reading ease). 

Table~\ref{tab:utt_readability_corr} reports Spearman correlations between the readability battery, the LLM-based cognitive load estimation and two measures: the full extraneous load composite and its clutter sub-component. All correlations are in the expected direction and statistically significant ($p<0.001$). The extraneous load composite correlates negatively with Flesch Reading Ease ($\rho=-0.236$) and positively with grade-level indices (Flesch-Kincaid $\rho=0.221$ ; SMOG $\rho=0.244$; Coleman-Liau $\rho=0.255$; Automated Readability Index $\rho=0.213$; Gunning Fog $\rho=0.223$). Vocabulary-based measures show a similar pattern (Dale-Chall $\rho=0.184$; Difficult Words $\rho=0.390$). The LLM-estimated cognitive load score exhibits a moderate positive correlation ($\rho=0.297$), indicating convergence across measurement paradigms--a formula-based readability approach and a model-based estimation approach both align with the composite in the expected direction. 

Two features of these correlations are worth noting. First, the magnitudes are moderate rather than strong, ranging from $\rho=0.105$ (McAlpine EFLAW) to $\rho=0.390$) for the full composite. This is the expected pattern for convergent validity with a related but non-identical construct. Readability formulas capture surface-level linguistic features--sentence length, syllable count, vocabulary difficulty--that contribute to processing costs but do not exhaust them. Our extraneous load composite additionally incorporates behavioral features at the interaction level such as task switching, information sprawl, and structural coherence, which operate above the sentence level and would not be captured by any readability formula. Moderate correlations therefore indicate meaningful overlap with established text-complexity measures while confirming that the composite captures additional variance beyond surface linguistics.

Second, the clutter subcomponent consistently shows stronger correlations than the full composite across all readability measures (e.g., Flesch-Kincaid: $\rho=0.403$ vs $0.221$; SMOG: $\rho=0.436$ vs. $0.244$; Difficult Words $\rho=0.634$ vs $0.390$). This pattern is expected and informative. Clutter is computed at the utterance level from textual features of individual turns, making it the component most directly comparable to readability indices that also operate the text-surface level. The full extraneous load composite aggregates clutter with other behavioral features--switching cost, information sprawl, truncation--that reflect organizational dynamics across the conversation rather than properties of any single utterance. The attenuation from clutter to the full composite confirms that higher-order behavioral components introduce variance that is orthogonal to surface-level text complexity, consistent with the composite capturing a broader construct than linguistic difficulty alone. 

The strongest individual correlations for the full composite with Difficult Words ($\rho=0.390$) and Reading Time ($\rho=0.373$). Both of these measures are sensitive to context density in ways that go beyond simple syntactic features: difficult word counts reflect vocabulary demands that increase processing effort, and reading time integrates other text characteristics into a single duration estimate. Their relatively stronger alignment with the extraneous load composite suggests that the composite is most convergent with readability measures that capture aggregate processing burden rather than any single surface feature.

\subsection{Concept Spread: Construction and Discriminant Validity}\label{app:discrim_validity}

To verify that our extraneous load (EL) composite captures processing costs associated with the organization and presentation of information rather than semantic density, we construct an alternative utterance-level measure-\textit{concept spread}-that captures the semantic diffuseness within an utterance's embedding space. For each utterance, we extract SBERT embeddings for \textit{(i)} all subtask-tagged text spans and \textit{(ii)} all FinanceKG concepts referenced in the utterance. We define concept spread as the trace of the covariance matrix of these vectors, which increases with greater dispersion across embedding dimensions. Because the trace mechanically increases with the number of vectors, we residualize concept spread on the log count of vectors in the utterance to isolate dispersion \textit{per unit context}. The resulting measure is $z$- standardized at the utterance level. 

Conceptually, concept spread captures \textit{how semantically scattered} an utterance is, whereas the EL composite captures how poorly organized information is relative to task structure (coherence, alignment, dispersion, and role-level dynamics). If EL were merely a proxy for semantic diffuseness, the two measures would exhibit similar speaker-role profiles, content-usage associations, and temporal signatures. We show that they do not. 

\begin{table}[h]
\centering
\small
\setlength{\tabcolsep}{5pt}
\renewcommand{\arraystretch}{1.15}
\caption{Discriminant validity: extraneous load composite vs.\ concept spread across four diagnostic tests.}
\label{tab:discriminant_validity}
\begin{tabular}{lcccc}
\toprule
 & \multicolumn{2}{c}{EL Composite} & \multicolumn{2}{c}{Concept Spread} \\
\cmidrule(lr){2-3}\cmidrule(lr){4-5}
 & $\beta$ / $\Delta$ & $p$ & $\beta$ / $\Delta$ & $p$ \\
\midrule
\multicolumn{5}{l}{\textbf{Panel A: Speaker-role profiles}} \\
\quad Paired $t$-test (Prompt $-$ Response)
 & $-0.540^{***}$ & $<0.001$ & $-0.037$ & $0.702$ \\
\quad FE + AR(2) + DK: is\_prompt
 & $-0.041$ & $0.490$ & $-0.115$ & $0.418$ \\
\addlinespace[6pt]
\multicolumn{5}{l}{\textbf{Panel B: Content usage associations (response-level spikes $\rightarrow$ CCU)}} \\
\quad Load spike $\rightarrow$ CCU
 & $+0.206^{**}$ & $0.018$ & $-0.263^{***}$ & $<0.001$ \\
\quad Preceding prompt load $\rightarrow$ CCU
 & $+0.039$ & $0.298$ & $+0.014$ & $0.692$ \\
\addlinespace[6pt]
\multicolumn{5}{l}{\textbf{Panel C: Shock propagation signatures (baseline-adjusted)}} \\
\quad Cross-speaker: From Prompt $\rightarrow$ current
 & $+0.089^{***}$ & $<0.001$ & $-0.015$ & $0.378$ \\
\quad Cross-speaker: From Response $\rightarrow$ current
 & $-0.011$ & $0.753$ & $-0.228^{***}$ & $0.004$ \\
\quad Within-speaker: Prompt $\rightarrow$ Prompt
 & $+0.231^{***}$ & $<0.001$ & $+0.184^{***}$ & $<0.001$ \\
\quad Within-speaker: Response $\rightarrow$ Response
 & $+0.162^{***}$ & $<0.001$ & $+0.220^{***}$ & $<0.001$ \\
\addlinespace[6pt]
\multicolumn{5}{l}{\textbf{Panel D: Tail spillover ($q = 0.90$, Prompt $\rightarrow$ Response)}} \\
\quad Continuous shock magnitude
 & $-0.068$ & $0.353$ & $+0.213^{***}$ & $0.002$ \\
\quad LPM: $\Pr(\text{response tail})$
 & $+0.016$ & $0.643$ & $+0.119^{***}$ & $0.003$ \\
\bottomrule
\end{tabular}
\vspace{0.6em}
\footnotesize \\
\textit{Notes.} Each row reports the same estimand using the EL composite (\texttt{z\_el\_utt}) versus concept spread (\texttt{z\_concept\_spread}). Panel~A compares prompt vs.\ response means (paired $t$-tests) and conditional differences in a participant fixed-effects model with AR(2) dynamics, utterance length, sequence position, and relative position (Driscoll--Kraay standard errors). Panel~B reports response-only models predicting standardized content usage (CCU) with both measures entered simultaneously. Panel~C reports baseline-adjusted autoregressive models with role-interacted lag structure. Panel~D evaluates spillover in the upper tail ($q=0.90$) using both continuous shocks and a linear probability model for tail membership. \textsuperscript{***}$p\leq0.01$, \textsuperscript{**}$p\leq0.05$, \textsuperscript{*}$p\leq0.1$.
\end{table}

Table~\ref{tab:discriminant_validity} reports the results of four diagnostic tests comparing the EL composite and concept spread. The measures diverge on every dimension tested. 

\textit{Speaker-role profiles} (Panel~A). The EL composite exhibits a large unconditional prompt-response difference ($\Delta=-0.540$, $p<0.001$) that collapses to near zero once autoregressive dynamics are controlled ($\beta=-0.041$, $p=0.490$), indicating that the raw speaker difference reflects temporal accumulation. Concept spread, by contrast, shows no speaker difference either unconditionally ($\Delta=-0.037$, $p=0.702$) or conditionally ($\beta=-0.115$, $p=0.418$). The two measures thus have qualitatively different relationships with speaker role. 

\textit{Content usage} (Panel~B). EL spikes are positively associated with content usage ($\beta=+0.206$, $p=0.018$), consistent with users incorporating more AI-generated content when organizational clutter is elevated. Concept spread spikes show the opposite pattern: a negative association with content usage ($\beta=-0.263$, $p<0.001$), indicating that semantically scattered responses co-occur with \textit{lower} content adoption. These opposite-sign associations confirm that the two constructs relate to content integration in qualitatively different ways. 

\textit{Shock propagation} (Panel-C). Panel C reports baseline-adjusted shock dynamics, which differ from the main dynamic FE results in Table~\ref{tab:rq2_core_dynamics}. Whereas Table~\ref{tab:rq2_core_dynamics} models persistence in the observed utterance-level EL process, Panel C first residualizes EL against speaker role, utterance length, sequence position, relative position, and turn number to isolate unexpected deviations ("shocks") from each participant's structural baseline. The dynamic model is then estimated on these residualized shock components. 

Under this stricter specification, prompt EL shocks propagate forward to the subsequent utterance ($\beta=+0.089$, $p<0.001$), whereas response EL shocks do not spill back to the user's next prompt ($\beta=-0.011$, $n.s.$). This pattern is substantively consistent with Table~\ref{tab:rq2_core_dynamics}, where the raw cross-speaker spillover from responses is small and statistically insignificant ($\beta=0.015$, $n.s.$). The difference in magnitude reflects the distinction between persistence in the overall EL process and propagation of baseline-adjuested deviations. Concept spread shocks exhibit the reverse pattern: response spread shocks are associated with \textit{lower} spread in the next utterance ($\beta=-0.228$, $p=0.004$), consistent with a contraction response, while prompt spread shocks show no forward propagation ($\beta=-0.015$, $n.s.$). The two measures thus have mirror-image dynamic signatures. 

\textit{Tail spillover} (Panel~D). At the 90th percentile, concept spread exhibits significant prompt-to-response tail contagion ($\beta=0.213$, $p=0.003$; $\Delta P =0.119$, $p=0.003$) that is entirely absent from the EL composite ($\beta=-0.068$, $n.s.; \Delta P = 0.016$, $n.s.$). Extreme semantic scatter in prompts is associated with elevated scatter in responses, but extreme organizational clutter does not propagate through the same threshold mechanism. 

Taken together, these tests confirm that the EL composite and concept spread capture distinct dimensions of extraneous processing demands. The EL composite reflects organizational clutter--coherence disruptions that are governed by within-speaker persistence and modest forward spillover. Concept spread reflects semantic diffuseness--a content-density-adjacent property with distinct speaker profiles, opposite content-usage associations, and a unique contraction dynamic. Including concept spread alongside the EL composite in the content usage models (Panel~B) does not attenuate the EL spike effect, further confirming that the EL composite is not a proxy for content volume.

\end{document}